\documentclass[12pt]{article}
\usepackage{enumitem}
\usepackage{mathpazo}
\usepackage[12pt]{moresize}
\usepackage[noblocks]{authblk}
\usepackage{hyperref}       % hyperlinks
\usepackage{url}            % simple URL typesetting
\usepackage{booktabs}       % professional-quality tables
\usepackage{amsfonts}       % blackboard math symbols
\usepackage{nicefrac}       % compact symbols for 1/2, etc.
\usepackage{fancyhdr}       % header
\usepackage{graphicx}       % graphics
\graphicspath{{media/}}     % organize your images and other figures under media/ folder
\usepackage{natbib}

\usepackage{balance} % for balancing columns on the final page

\usepackage{hyperref}       % hyperlinks
\usepackage{amsmath}
\usepackage{mathtools, amsthm}
\usepackage{adjustbox}
\usepackage{tabularx}
\usepackage{multirow}
\usepackage{tcolorbox}
\usepackage[ruled,linesnumbered,boxed]{algorithm2e}

\SetCommentSty{mycommfont}

\usepackage{savesym}
\usepackage{setspace} 
\usepackage{threeparttable}

\newtheorem{theorem}{Theorem}[section]
\numberwithin{theorem}{subsection}

\newtheorem{lemma}[theorem]{Lemma}
\newtheorem{proposition}[theorem]{Proposition}

\newtheorem{definition}[theorem]{Definition}
\newtheorem{hyp}{Assumption}

\newcommand{\order}[1]{}

\numberwithin{equation}{subsection}

\newcommand{\fedexp}{{\sc FedExp3}}

\newcommand{\gucb}{{\sc GUCB}}
\newcommand{\pfedexp}{{\sc FedOCO}}

\newcommand{\reals}{\mathrm{I\! R}}
\newcommand{\ind}{\mathrm{1\!\! I}}

\newcommand{\pp}{\pi}
\newcommand{\f}{w}

\newcommand{\1}{\mathbf{1}}

\renewcommand{\P}{\mathbb{P}}
\newcommand{\E}{\mathbb{E}}

\newcommand{\alg}{\mathcal{A}}
\newcommand{\gb}{\mathcal{G}}
\newcommand{\breg}{\textsf{R}}
\newcommand*{\rom}[1]{%
  \textup{\uppercase\expandafter{\romannumeral#1}}%
}

\savesymbol{iint}
\restoresymbol{TXF}{iint}
\DeclareGraphicsExtensions{.png, .pdf, .jpg, .mps, .ps}
\graphicspath{{./Figures/}}
\renewcommand{\cite}{\citet}

\makeindex

\begin{document}

\title{{\textbf{\large{The London School of Economics and Political Science}}\\ 
\bigskip\bigskip\bigskip\bigskip\bigskip\bigskip\bigskip\bigskip\bigskip\bigskip\bigskip  \emph{Regret-Minimization Algorithms for Multi-Agent Cooperative Learning Systems \bigskip\bigskip\bigskip\bigskip\bigskip\bigskip\bigskip\bigskip\bigskip\bigskip\bigskip\bigskip}}}

\author{\Large{Jialin Yi \bigskip\bigskip\bigskip\bigskip\bigskip\bigskip\bigskip\bigskip\bigskip\bigskip\bigskip\bigskip\bigskip\bigskip}}

\date{\normalsize{A thesis submitted to the Department of Statistics of the London School of Economics and Political Science for the degree of Doctor of Philosophy, London, February 2023.}}

\maketitle
\thispagestyle{empty}

\newpage \doublespacing

\section*{\textbf{\bigskip\bigskip\bigskip\bigskip\bigskip \Large{Declaration}}}

I certify that the thesis I have presented for examination for the PhD degree of the London School of Economics and Political Science is solely my own work other than where I have clearly indicated that it is the work of others (in which case the extent of any work carried out jointly by me and any other person is clearly identified in it).

The copyright of this thesis rests with the author. Quotation from it is permitted, provided that full acknowledgment is made. This thesis may not be reproduced without the prior written consent of the author.

I warrant that this authorization does not, to the best of my belief, infringe the rights of any third party.

I declare that my thesis consists of 22710 words.

\bigskip
\bigskip
\bigskip
\bigskip
\bigskip
\bigskip
\bigskip
\bigskip
\bigskip

\newpage

\section*{\textbf{\Large{Statement of co-authored work}}}

\bigskip\bigskip\bigskip\bigskip

I confirm that the work in Chapter~3 was jointly co-authored with Professor Milan Vojnovi{\'  c} and I contributed 80\% of this work (problem setting, algorithms, main proofs, and code implementations).
I confirm that the work in Chapter~4 was jointly co-authored with Professor Milan Vojnovi{\'  c} and I contributed 90\% of this work (problem setting, algorithms, proofs, and code implementations).
I confirm that the work in Chapter~5 was authored by me only.
I confirm that the work in Chapter~6 was jointly co-authored with Professor Milan Vojnovi{\'  c} and I contributed 
40\% of this work (algorithms, partial proofs, and code implementations).

\bigskip
\bigskip
\bigskip
\bigskip 
\bigskip
\bigskip
\bigskip
\bigskip
\bigskip

\newpage \onehalfspacing

\section*{\textbf{\Large{Acknowledgment}}}

\bigskip\bigskip\bigskip\bigskip

I would like to express my heartfelt gratitude to Milan Vojnovi{\'  c} and Clifford Lam, my advisors, who have provided me with invaluable guidance, support, and encouragement throughout my PhD journey. Their unwavering belief in my abilities and their relentless pursuit of excellence have been an inspiration to me.
Especially, I want to give my warmest thank to Milan Vojnovi{\'  c} for his advice and help on research directions, academic writing, presentation skills, and my career path; to Cl{\' e}ment Calauz{\` e}nes , Chuan Luo and Vianney Perchet, for their countless help on my research projects and job search.
The internship at Microsoft Research at Asia was a fantastic experience to me because of the guidance and help from Hang Dong and Chuan Luo.

\sloppy I would also like to extend my sincere thanks to Laurent Massouli{\'  e} and Chengchun Shi, for their insightful comments, constructive criticism, and valuable suggestions. Their invaluable contributions have greatly improved the quality of this thesis.

I am deeply grateful to Department of Statistics at London School of Economics and Political Science for providing me with the scholarship, resources and facilities necessary to complete this research.
I really appreciate the help from my colleagues in Department of Statistics as well.

I would like to dedicate this thesis to my parents, Li Huang and Xixiang Yi, for their ever-lasting love and support. Their encouragement and trust have been my driving force.

Finally, I want to extend my deepest appreciation to my forever friends, Chuan Cheng, Zezhun Chen, Junyi Liao, Jiaqi Lyu, Zhaoxi Li, Yaqian Wu and Yu Yi. 
Getting a PhD is a long journey during which I not only received precious memories but also lost important ones.
It were those friends whose accompany and care saved me during the hardest time in my PhD journey.

Thank you all.

\newpage

\section*{\textbf{\Large{Abstract}}}

\bigskip\bigskip\bigskip\bigskip\

A Multi-Agent Cooperative Learning (MACL) system is an artificial intelligence (AI) system where multiple learning agents work together to complete a common task.
Recent empirical success of MACL systems in various domains (e.g. traffic control, cloud computing, robotics) has sparked active research into the design and analysis of MACL systems for sequential decision making problems.
One important metric of the learning algorithm for decision making problems is its \emph{regret}, i.e. the difference between the highest achievable reward and the actual reward that the algorithm gains.
The design and development of a MACL system with low-regret learning algorithms can create huge economic values. 

\sloppy In this thesis, I analyze MACL systems for different sequential decision making problems.
Concretely, the Chapter~3 and 4 investigate the cooperative  multi-agent multi-armed bandit problems, with full-information or bandit feedback, in which multiple learning agents can exchange their information through a communication network and the agents can only observe the rewards of the actions they choose.
Chapter~5 considers the communication-regret trade-off for online convex optimization in the distributed setting.
Chapter~6 discusses how to form high-productive teams for agents based on their unknown but fixed types using adaptive incremental matchings.

For the above problems, I present the regret lower bounds for feasible learning algorithms and provide the efficient algorithms to achieve this bound. The regret bounds I present in Chapter~3, 4 and 5 quantify how the regret depends on the connectivity of the communication network and the communication delay, thus giving useful guidance on design of the communication protocol in MACL systems.

\newpage

%\section*{\textbf{\Large{Contents}}}

\tableofcontents

\newpage

%\section*{\textbf{\Large{List of Tables}}}
\listoftables

\newpage

%\section*{\textbf{\Large{List of Figures}}}
\listoffigures

% \newpage
% \section*{\textbf{\Large{Notations}}}

\newpage \onehalfspacing

 %%%%%%%%%%%%%%%%%%%%%%%%%%%%%%%%%%%%%

\newpage\onehalfspacing

\setcounter{section}{1}
\setcounter{subsection}{0}

\section*{\textbf{\Large{Chapter~1}}}
\addcontentsline{toc}{section}{Chapter~1. Introduction}

\bigskip
\bigskip

\textbf{\Large{Introduction}}

\bigskip
\bigskip
\bigskip

\noindent{\sloppy A Multi-Agent Cooperative Learning (MACL) system is an artificial intelligence (AI) system where multiple learning agents work together to complete a common task. These agents communicate with and learn from each other, improving their performance over time. The cooperative nature of these systems is beneficial for solving complex problems, as the agents can divide tasks, share information and thus accelerate their learning processes.
}

Recently, MACL systems have demonstrated successful applications in various domains, including traffic control \citep{chu2019multi}, cloud computing \citep{mao2022mean}, robotics \citep{wuhuman} and so on. 
This empirical success has sparked active research into the design and analysis of MACL systems for sequential decision making problems.
Despite significant advancements in the field, there remain challenges to overcome, even for some fundamental sequential decision making problems: the Multi-Armed Bandit (MAB) problem \citep{auer2002finite, auer2002nonstochastic} and the Online Convex Optimization (OCO) problem \citep{hazan2016introduction}.
In particular, when learning agents are connected by a communication network,
\begin{itemize}
    \item the \emph{optimal} distributed regret-minimization algorithm, 
    \item the effect of the communication network's \emph{connectivity} on the regret of an algorithm,
    \item the effect of \emph{delays} in communication on the regret of an algorithm,
    \item and the trade-off between the regret and the \emph{communication complexity}
\end{itemize}
are not yet fully understood. Addressing these questions from a theoretical perspective is important and challenging, and this thesis presents a set of efforts towards this aim.

The study begins with a canonical setting in non-stochastic MAB problems which involve selecting the most rewarding action from a fixed set of actions over a fixed time horizon.
In the single-agent non-stochastic MAB problem, near-optimal (Exp3) and optimal (Tsallis-INF) algorithms have been proposed by \cite{auer2002nonstochastic} and \cite{zimmert2021tsallis}, respectively, as is introduced in Chapter~2.
The natural extension to the cooperative multi-agent setting involves each agent individually selecting an action, followed by exchanging information among them. 
Based on rewards received by agents, cooperative multi-agent MAB problems can be divided into two settings: the homogeneous reward setting and the heterogeneous reward setting.

The homogeneous reward setting involves a scenario where multiple agents receive the same reward for choosing the same action at the same time step, which is commonly observed in the distributed training of reinforcement learning algorithms \citep{babaeizadehreinforcement}.
In this setting, all agents face the same MAB problem, and the exploration phase of the learning algorithm is executed in parallel by each agent. The challenge lies in devising efficient cooperative algorithms under communication delays.

In contrast to the homogeneous reward setting, the heterogeneous reward setting, motivated by federated learning systems  \citep{kairouz2021advances}, involves a scenario where multiple agents may choose the same action at the same time step, but receive \emph{different} rewards. The objective for each agent is to determine the globally optimal action in hindsight based on the collective experience of all agents, which requires communication among agents. To achieve this goal, agents must reach a consensus on the globally optimal action by exchanging their private information. The challenge in this setting lies in the design of an appropriate mechanism for exchanging private information, which allows for cooperative learning while preserving the privacy of each individual agent.

Despite borrowing algorithms and analysis from the single-agent sequential decision making, the multi-agent cooperative learning introduces unique challenges not present in the single-agent case. One such challenge is the trade-off between the regret of a cooperative online learning algorithm and the communication complexity incurred by that algorithm \citep{agarwal2022multi, wan2020projection}.

Except for continuous optimization problems, efficiently finding a maximum value matching for a given set of nodes based on observations of the previously proposed matching over a sequence of rounds is another fundamental sequential learning problem. This problem arises in many applications including team formation, online social networks, online gaming, online labor, and other online platforms \citep{KKL18, notes}. The challenge lies in the design of optimal regret-minimization algorithms to address this problem.

The study of MACL systems encompasses a broad area of research, which is beyond the scope of this thesis. This thesis focuses on the theory of non-stochastic multi-agent MAB problems, federated OCO problems and the sequential learning of a maximum value matching, while acknowledging the existence of other significant topics such as stochastic multi-agent MAB problems, which are not discussed in this thesis.

\subsection{The structure of this thesis}

The preliminary knowledge related to non-stochastic MAB problems is presented in Chapter~2. Subsequently, Chapter~3 and 4 delve deeper into cooperative non-stochastic multi-agent MAB problems, each focusing on a specific setting. Chapter~3 covers the homogeneous reward setting, and Chapter~4 examines the heterogeneous reward setting.
Chapter~5 is dedicated to the analysis of the trade-off between the regret and the communication complexity in federated online convex optimization problems. 
Chapter~6 investigates how to find a maximum value matching in a sequential manner under the minimum change constraint. 
Finally, the thesis concludes in Chapter~7 with a summary of the findings, and suggestions for future research directions.

\subsection{Contributions}

The contributions of this thesis can be summarized as follows:

\paragraph{Cooperative non-stochastic multi-armed bandits (Chapter~3)}
    We consider a non-stochastic multi-agent multi-armed bandit problem with agents collaborating via a communication network with delays.
    We show a lower bound for individual regret of all agents.
    We show that with suitable regularizers and communication protocols, a collaborative multi-agent \emph{follow-the-regularized-leader} (FTRL) algorithm has an individual regret upper bound that matches the lower bound up to a constant factor when the number of arms is large enough relative to degrees of agents in the communication graph. We also show that an FTRL algorithm with a suitable regularizer is regret optimal with respect to the scaling with the edge-delay parameter. 
    We present numerical experiments validating our theoretical results and demonstrate cases when our algorithms outperform previously proposed algorithms. 
This chapter is based on joint work with Milan Vojnovi{\'  c} \citep{yi2022regret}.

\paragraph{Doubly adversarial federated bandits (Chapter~4)}
\sloppy We study a new non-stochastic federated multi-armed bandit problem with multiple agents collaborating via a communication network. The losses of the arms are assigned by an oblivious adversary that specifies the loss of each arm not only for each time step but also for each agent, which we call ``doubly adversarial". In this setting, different agents may choose the same arm in the same time step but observe different feedback. The goal of each agent is to find a globally best arm in hindsight that has the lowest cumulative loss averaged over all agents, which necessities the communication among agents.
We provide regret lower bounds for any federated bandit algorithm under different settings, when agents have access to full-information feedback, or the bandit feedback.
The regret lower bound for the full-information setting also applies to distributed online linear optimization problems, and, to the best of our knowledge, is the first lower bound result for this problem.
For the bandit feedback setting, we propose a near-optimal federated bandit algorithm called \fedexp.
Our algorithm gives a positive answer to an open question proposed in \cite{cesa2016delay}: {\fedexp} can guarantee a sub-linear regret without exchanging sequences of selected arm identities or loss sequences among agents.
We also provide numerical evaluations of our algorithm to validate our theoretical results and demonstrate its effectiveness on synthetic and real-world datasets.
This chapter is based on joint work with Milan Vojnovi{\'  c} \citep{yi2023doubly}.

\paragraph{Communication-regret trade-off in federated online learning (Chapter~5)}
The increasing deployment of large-scale federated learning systems has led to a growing interest in research on distributed online convex optimization algorithms \citep{wangni2018gradient, reisizadeh2020fedpaq}.
However, a significant challenge in practical implementation of these algorithms is the high communication complexity which is measured in the cumulative number of messages exchanged between agents.
It remains an open question whether it is possible for a distributed online convex optimization algorithm, over $T$ time steps, to concurrently achieve $o(T)$ cumulative regret and $o(\sqrt{T})$ communication complexity.
In this chapter, we address this challenge by introducing a communication-efficient stochastic algorithm, {\pfedexp}.

\paragraph{Greedy Bayes incremental matching strategies (Chapter~6)}
We consider the problem of finding a maximum value matching in a given set of nodes based on observed evaluations of node pairs, where the value of a node pair is according to a function of hidden node types---we refer to as a value function. This is considered in a sequential learning setting where in each round, the learner selects a matching of nodes and observers the value of each matched pair of nodes. Importantly, the sequence of matchings must be incremental, meaning that in each round, the matching differs from the matching in the previous round for at most four pairs, i.e. a new matching is derived from the previous matching by re-matching nodes of any two previously matched pairs with each other. This problem is motivated by situations in which matchings need to be explored by making small incremental alternations of existing matches over time. We consider the efficiency of learning with respect to regret minimization for greedy Bayes incremental matching algorithms, which in each round select a matching that has the maximum expected reward in the next round based on the posterior distribution of node types, conditional on data observed up to the current round. Our focus is on the case of binary node types and understanding the efficiency of learning for different value functions, including all symmetric Boolean functions. We identify greedy Bayes incremental matching policies for different value functions and characterize how the regret depends on the population size and distribution of node types.  
This chapter is based on joint work with Milan Vojnovi{\'  c}.

\subsection{Numerical results}

All experiments in this thesis are run on a desktop with AMD Ryzen 5 2600 Six-Core Processor and 16GB memory.
Each experiment took less than 6 hours to finish. 

The code is written in Python~3.9 and uses Numpy package \citep{harris2020array} and NetworkX package \citep{hagberg2008exploring} for numerical calculations and graph operations.
The Numpy package and the NetworkX package are distributed with the BSD license.
%%%%%%%%%%%%%%%%%%%%%%%%%%%%%%%%%%%%%

\newpage\onehalfspacing

\setcounter{section}{2}
\setcounter{subsection}{0}

\section*{\textbf{\Large{Chapter~2}}}
\addcontentsline{toc}{section}{Chapter~2. Background}

\bigskip
\bigskip

\textbf{\Large{Background}}

\bigskip
\bigskip
\bigskip

\subsection{Multi-armed bandit (MAB) problems}

The multi-armed bandit problem stems from the concept of a ``bandit" in gambling. It's a term used to describe a one-armed bandit machine, also known as a slot machine, where a player pulls a lever to obtain a \emph{reward}. 
The multi-armed bandit problem models this situation as one where there are multiple levers, (\emph{arms} or \emph{actions}) to choose from and the player wants to maximize his or her rewards over a series of plays by balancing exploration (trying out different arms to learn their rewards) with exploitation (using the knowledge of the best arm to maximize rewards). The problem has been formalized in the field of reinforcement learning and decision theory and has applications in various fields such as online advertising, clinical trials and recommender systems \citep{li2010contextual, lattimore_szepesvari_2020}.

Depending on the assumption about how the observed rewards are generated, MAB problems can be classified into two main types: 
\begin{itemize}
    \item stochastic bandits: the reward from each arm at any given time is an independent and identically distributed sample from the underlying distribution,
    \item adversarial bandits: the reward from each arm at any given time is determined by an adversary.
\end{itemize}

In this thesis, the focus is on adversarial bandits, which are commonly conceptualized as repeated two-player zero-sum games between an agent and an adversary, each with distinct objectives and strategies. Given its strong association with online linear optimization, the adversarial bandit community often refers to the terms ``arms" as ``actions" and the inverse of ``rewards" as ``losses". Consequently, the objective of reward maximization is transformed into loss minimization. This convention is adopted in this thesis.

\subsubsection{Concepts}
We use $\ell(i)$ to denote the $i$-th coordinate of the vector $\ell$.
And $\mathcal{S}^\ast$ be the set of all sequences whose entries are from the set $\mathcal{S}$, i.e. $\mathcal{S}^\ast = \{(s_1, s_2, \dots): s_i \in \mathcal{S} \text{ for all } i\}$. 

\paragraph{Environment}
Let $\mathcal{A}$ be a set of actions with $|\mathcal{A}| = K$ and $T$ be the number of repeated rounds, i.e. horizon.
A $K$-armed adversarial bandit environment (or instance) is an arbitrary sequence of loss vectors $\{\ell_t\}_{t=1}^T$, where $\ell_t\in[0, 1]^K$.
At each time step $t$, the agent chooses a distribution $p_t$ over the actions.
Then an action $a_t \in \mathcal{A}$ is sampled from $p_t$ and the agent receive a loss of $\ell_t(a_t)$.

\paragraph{Policy}
The agent in the adversarial bandit environment is allowed to make the decision based on the historical information $\mathcal{F}_t = \{a_1, \ell_1(a_1), a_2, \ell_2(a_2), \dots, a_{t-1}, \ell_{t-1}(a_{t-1})\} \in (\mathcal{A}\times [0, 1])^\ast$.
Let $\mathcal{P}_{K-1}$ be the set of probability distributions on the $K$ actions, i.e. a $K-1$-simplex $\{p\in [0, 1]^K: \sum_{i=1}^K p_i = 1\}$.
A bandit algorithm, also called \emph{policy}, is a function
$$
\pi: (\mathcal{A}\times [0, 1])^\ast \to \mathcal{P}_{K-1} 
$$
mapping the historical information to a probability distribution over the actions.

\paragraph{Regret}

The performance of a policy $\pi$ is measured by its \emph{regret}, defined as the difference of the expected cumulative loss incurred by following the policy and the expected cumulative loss of a best \emph{fixed} arm in hindsight, i.e.
\begin{equation*}
R_T(\pi, \{\ell_t\}) = \mathbb{E} \left[ \sum_{t=1}^T \ell_t(a_t) - \min_{i\in \mathcal{A}}\left\{ \sum_{t=1}^T \ell_t(i)\right\} \right]
\end{equation*}
where the expectation is taken over chosen actions under algorithm $\pi$ on instance $L$. 
We will abbreviate $R_T(\pi, \{\ell_t\})$ as $R_T$ when the algorithm $\pi$ and adversarial bandit environment $\{\ell_t\}$ have no ambiguity in the context.

\subsubsection{Regret minimization}

The objective of the designer of a bandit algorithm is to identify a policy $\pi$ that minimizes regret $R_T$ for the agent in any bandit environment $\{\ell_t\}$ selected by the adversary. This requires the development of a policy that effectively balances \emph{exploration} and \emph{exploitation} in a manner that results in a small regret for the agent over time.
On the one hand, exploration is necessary to gather information about different actions and estimate their expected losses. On the other hand, exploitation is necessary to minimize the loss by selecting the best-performing action so far. The difficulty lies in finding the optimal trade-off between exploration and exploitation in an uncertain and dynamic environment.

\paragraph{Minimax lower bound}

From the optimization perspective, the design of a bandit algorithm with best worst-case regret can be formulated as a minimax optimization problem, i.e.,
\begin{align*}
\min_{\pi\in \Pi} \max_{\{\ell_t\in[0,1]^{K}\}} & R_T(\pi, \{\ell_t\}). \
\end{align*}
where $\Pi$ is the set of all policies. 

We present a lower bound on $\max_{\{\ell_t\in[0,1]^{K}\}} R_T(\pi, \{\ell_t\})$ which applies to all policy $\pi\in \Pi$, which is so-called \emph{minimax lower bound}.

\begin{theorem}[\cite{lattimore_szepesvari_2020}]
\label{chapter-2:thm:high-prob}
Let $c, C>0$ be sufficiently small/large universal constants and $K \geq 2, T \geq 1$ and $\delta \in(0,1)$ be such that $T \geq C K \log (1 /(2 \delta))$. Then there exists a loss sequence $\ell_t \in[0,1]^{K}$ such that
$$
\mathbb{P}\left[ \sum_{t=1}^T \ell_t(a_t) - \min_{i\in \mathcal{A}}\left\{ \sum_{t=1}^T \ell_t(i)\right\} \geq c \sqrt{KT \log \left(\frac{1}{2 \delta}\right)}\right] \geq \delta.
$$
\end{theorem}
Theorem~\ref{chapter-2:thm:high-prob} implies a lower bound on $R_T$:
\begin{align*}
    R_T 
    &= \mathbb{E} \left[\sum_{t=1}^T \ell_t(a_t) - \min_{i\in \mathcal{A}}\left\{ \sum_{t=1}^T \ell_t(i)\right\} \right] \\
    &= \int_{0}^\infty \mathbb{P}\left[ \sum_{t=1}^T \ell_t(a_t) - \min_{i\in \mathcal{A}}\left\{ \sum_{t=1}^T \ell_t(i)\right\} \geq x\right] dx \\
    &\geq \frac{1}{2} \int_{0}^\infty \exp\left(-\frac{x^2}{c^2 KT}\right) dx \\
    &= \Omega\left( \sqrt{KT} \right).
\end{align*}
Hence, for any policy $\pi\in\Pi$, there exists a bandit environment such that the regret of the policy is $\Omega\left( \sqrt{KT} \right).$

\paragraph{Regret minimization algorithms}

The minimax lower bound in Theorem~\ref{chapter-2:thm:high-prob} provides a benchmark for evaluating regret-minimization algorithms and can be used to determine whether an algorithm is optimal or sub-optimal. Optimal algorithms achieve the $O(\sqrt{KT})$ order in minimax lower bound, while sub-optimal algorithms have a higher regret with an order of $O(\sqrt{KT\log K})$ .
We present a sub-optimal algorithm (Exp3) and an optimal algorithm (Tsallis-INF) with their regret upper bounds.

\begin{algorithm}[h]
\caption{Exp3 \citep{auer2002nonstochastic}}
\label{chapter2:algo:Exp3}
\SetKwInOut{Init}{Initialization}
\SetKwInOut{Input}{Input}
\Input{$\gamma \in (0, 1]$.} 
\Init{$w_1(i)=1$ for $i=1, \ldots, K$.}
\For{each time step $t=1,2,\dots, T$}{
    Set $$p_t(i) = (1-\gamma) \frac{w_t(i)}{\sum_{j=1}^K w_t(j)} + \gamma_t/K$$ for $i=1, \ldots, K$\;
    Draw $a_t$ randomly accordingly to the probabilities $p_t(1), \ldots, p_t(K)$.\;
    Receive reward $\ell_t(a_t)$\;
    For $j=1,2,\dots,K$, set
    $$
    \begin{aligned}
    \hat{\ell}_t(j) & =\left\{\begin{array}{cl}
    \ell_t(j) / p_t(j) & \text { if } j=a_t, \\
    0 & \text { otherwise }
    \end{array}\right. \\
    w_{t+1}(j) & =w_t(j) \exp \left(\gamma \hat{\ell}_t(j) / K\right);
    \end{aligned}
    $$
}
\end{algorithm}

\begin{theorem}[\cite{auer2002nonstochastic}]
    For any $K>0$ and $T>0$, suppose the agent follows Exp3 algorithm as described in Algorithm~\ref{chapter2:algo:Exp3}, with 
    $$
    \gamma=\min \left\{1, \sqrt{\frac{K \ln K}{(e-1) T}}\right\},
    $$
    the regret of the agent 
    $$
    R_T \leq 2.63 \sqrt{KT\log K}.
    $$
\end{theorem}

\begin{algorithm}[h]
\caption{Tsallis-INF \citep{zimmert2021tsallis}}
\label{chapter2:algo:tsallis-inf}
\SetKwInOut{Init}{Initialization}
\SetKwInOut{Input}{Input}
\Input{a sequence of learning rates $\{\eta_t \in (0, 1]\}$.} 
\Init{$w_1(i)=1$ for $i=1, \ldots, K$.}
\For{each time step $t=1,2,\dots, T$}{
    Set $$p_t(i) = (1-\gamma) \frac{w_t(i)}{\sum_{j=1}^K w_t(j)} + \gamma_t/K$$ for $i=1, \ldots, K$\;
    Draw $a_t$ randomly accordingly to the probabilities $p_t(1), \ldots, p_t(K)$.\;
    Receive reward $\ell_t(a_t)$\;
    For $j=1,2,\dots,K$, set
    $$
    \begin{aligned}
    \hat{\ell}_t(j) & =\left\{\begin{array}{cl}
    \ell_t(j) / p_t(j) & \text { if } j=a_t, \\
    0 & \text { otherwise }
    \end{array}\right. \\
    w_t & =\underset{w \in \Delta^{K-1}}{\arg \max }\left\langle w,-\sum_{s=1}^t\hat{\ell}_t\right\rangle+\frac{4}{\eta_t} \sum_i \sqrt{w_i} .;
    \end{aligned}
    $$
}
\end{algorithm}

\begin{theorem}[\cite{zimmert2021tsallis}]
    For any $K>0$ and $T>0$, suppose the agent follows Tsallis-INF algorithm as described in Algorithm~\ref{chapter2:algo:tsallis-inf}, with 
    $$
    \eta_t=2\sqrt{\frac{1}{t}},
    $$
    the regret of the agent 
    $$
    R_T \leq 4 \sqrt{KT} + 1.
    $$
\end{theorem}

\subsection{Extensions to MAB problems}
Recently, the basic MAB problem, which involves the exploration and exploitation of multiple arms with unknown rewards, has been extended to accommodate more complex scenarios. 
These modifications provide more nuanced models of real-world systems and offer new insights into the behavior of the MAB problem under different conditions \citep{lattimore_szepesvari_2020}.

The extensions of most relevance to this thesis include the multi-agent setting \citep{cesa2019delay, bar2019individual, dubey2020differentially, zhu2021federated, shi2021federated, reda2022near}, the delayed feedback setting \citep{joulani2013online, zimmert2020optimal} and the combinatorial bandit setting \citep{comb, CTPL15, KZAC15, combgenreward, combsumfull, combstochdom}.

In this section, we provide an overview of previous studies that are relevant to our research, and elucidate unique contributions of this thesis in the context of existing literature.

\subsubsection{Multi-agent MAB problems}

\paragraph{Multi-agent cooperative adversarial bandits}

\cite{cesa2016delay, bar2019individual} studied the adversarial case where agents receive the same loss for the same action chosen at the same time step, whose algorithms require the agents to exchange their raw data with neighbors.
In Chapter~3, we give a regret lower bound for any learning algorithm in which each agent can only communicate with their neighbors. We present a center-based algorithm whose regret upper bound matches the lower bound when the number of arms is large enough. We present an algorithm that has a regret upper bound with $\sqrt{d}$ dependence on the delay $d$ per edge, which is optimal. All our regret bounds are parametrized with the delay parameter $d$, which is unlike to \citet{cesa2019delay} and \citet{bar2019individual} which considered only the special when $d=1$. In what follows we summarise our results in more details.

In these works, agents that choose the same action at the same time step receive the same reward or loss value and agents aggregate messages received from their neighbors. Our work in Chapter~4 relaxes this assumption by allowing agents to receive different losses even for the same action in a time step. Besides, we propose a new algorithm that uses a different aggregation of messages than in the aforementioned papers, which is based on distributed dual averaging method in \cite{nesterov2009primal, xiao2009dual, duchi2011dual}.

% In prior research, agents were required to continuously exchange updated information with their neighbor, resulting in a linear increase in communication costs with each time step. In Chapter~5, we introduce a novel algorithm that achieves both sub-linear expected communication costs and cumulative individual regret simultaneously.

\paragraph{Federated bandits}

Solving bandit problems in the federated learning setting has gained attention in recent years. \cite{dubey2020differentially} and \cite{huang2021federated}  considered the linear contextual bandit problem and extended the LinUCB algorithm \citep{li2010contextual} to the federated learning setting.
\cite{zhu2021federated} and \cite{shi2021federated}
studied a federated multi-armed bandit problem where the losses observed by different agents are i.i.d. samples from some common unknown distribution.
\cite{reda2022near} considered the problem of identifying a globally best arm for multi-armed bandit problems in a centralized federated learning setting. All these works focus on the stochastic setting, i.e. the reward or loss of an arm is sampled from some unknown but fixed distribution. Our work in Chapter~4 considers the non-stochastic setting, i.e. losses are chosen by an oblivious adversary, which is a more appropriate assumption for non-stationary environments.

\subsubsection{Delayed feedback}

Multi-armed bandits with delayed feedback have been studied extensively in the single-agent setting \citep{joulani2013online, thune2019nonstochastic, flaspohler2021online}. Specifically,
\citet{zimmert2020optimal} considered a setting in which the agent has no prior knowledge about the delays and showed an optimal regret of $O(\sqrt{KT}+\sqrt{d\log(K)T})$ where $d$ is the average delay over $T$ time steps. We present in Chapter~3, in the multi-agent setting, a distributed learning algorithm whose regret upper bound can also achieve this optimal $\sqrt{d}$ dependence on the delay per edge $d$.

\subsubsection{Combinatorial bandits}

For combinatorial bandits, in each round, the decision maker selects an arm that corresponds to a feasible subset of base arms. For example, given a set of $N$ base arms, an arm is an arbitrary subset of base arms of cardinality $K$. Under full bandit feedback, the reward of the selected arm is observed, which is according to an underlying function of the  corresponding base arms' values. For example, this may be sum of values \citep{comb,CTPL15,combsumfull}, or some non-linear function of values \citep{combgenreward,combstochdom}. Under semi-bandit feedback, individual values of all selected base arms are observed. In our work in Chapter~5, we may think of node pairs to be base arms, matchings to be arms, and we have semi-bandit feedback. Alternatively, we may regard nodes to be base arms, matchings to be arms, and we have full-bandit feedback for each match in a matching. Note that requiring for sequence of matching selections to be according to an incremental matching sequence is not studied in the line of work on combinatorial bandits.

\subsection{Differential equation approximation of discrete stochastic processes}

Here we provide background results on differential equation approximation for discrete random processes by \cite{Wormald95}, which we use in the proof of Theorem~\ref{thm:max}.

Consider a discrete random process on probability space $\Omega$, denoted by $X_0, X_1, \ldots$, where each $X_t$ takes values in some set $S$. We use ${\mathcal F}_t$ to denote $(X_0, X_1, \ldots, X_t)$, the history of the process up to time $t$. For a function $y$ of history ${\mathcal F}_t$, random variable $y({\mathcal F}_t)$ is denoted by $Y_t$. 

We consider a sequence $\Omega_n$, $n = 1,2, \ldots$ of random processes. For a random variable $X$, we say $X = o(f(n))$ always if 
$$
\max\{x: \P[X=x]\neq 0\} = o(f(n)).
$$
An event is said to hold \emph{almost surely} if its probability in $\Omega_n$ is $1-o(1)$. Let $S_n^+$ denote the set of all functions $f_n (x_0, x_1,\ldots,x_t)$ where each $x_i \in S_n$, for $t\geq 0$. 

A function $f:\reals^d\rightarrow \reals$ is said to satisfy a Lipschitz condition on $D\subseteq \reals^d$ if there exists a constant $L > 0$ such that $|f(x)-f(y)|\leq L ||x-y||_1$ for all $x$ and $y$ in $D$.

Recall the definition of the extension of the solution to a differential equations 
$$
\frac{d}{d t} x(t) = f(t, x)
$$
where $f$ and $f_x$ are continuous functions.
\begin{definition}[extension of solutions]
    Given two solutions $u: J \subset \mathbb{R} \rightarrow \mathbb{R}^{d-1}$ and $v: I \subset \mathbb{R} \rightarrow \mathbb{R}^{d-1}, u$ is called an extension of $v$ if $I \subset J$ and
    $$
u(x)=v(x) \quad x \in I.
$$
\end{definition}

Let $d$ be a fixed positive integer. For $i \in [d] = \{1,\ldots, d\}$, let
$$
y^{(i)} : \cup_n S_n^+\rightarrow \reals
$$
and $f_i: \reals^{d+1}\rightarrow \reals$ be such that for some constant $C$ and all $i\in [d]$,
$$
|y^{(i)}({\mathcal F}_t)| \leq Cn \hbox{ for all } {\mathcal F}_t \in S_n^+ \hbox{ for all } n.
$$
Assume that for some sequence $m(n)$, the following conditions hold:
\begin{enumerate}
\item[(C1)] there is a constant $C'$ such that for all $t< m(n)$ and all $i\in [d]$, it holds always,
$$
|Y^{(i)}_{t+1}-Y^{(i)}_t|\leq C'.
$$
\item[(C2)] for all $i\in [d]$ and uniformly over $t < m(n)$, it holds always,
$$
\E[Y^{(i)}_{t+1}-Y^{(i)}_t \mid {\mathcal F}_t] = f_i\left(\frac{t}{n}, \frac{Y_t^{(1)}}{n}, \ldots, \frac{Y^{(d)}_t}{n}\right) + o(1).
$$
\item[(C3)] for all $i\in [d]$, $f_i$ is continuous and satisfies a Lipschitz condition on $D$ where $D$ is some bounded connected open set containing the intersection of $\{(t,y^{(1)}, \ldots, y^{(d)}): t\geq 0\}$ with some neighborhood of $\{(0, y^{(1)}, \ldots, y^{(d)}): \P[\cap_{i\in [d]}\{Y^{(i)}_0 = y^{(i)}n\}]\neq 0 \hbox{ for some } n\}$.
\end{enumerate}

\begin{theorem}[\cite{Wormald95}] Assume that conditions (C1)-(C3) hold. Then, 
\begin{enumerate}
\item For $(0,\hat{x}^{(1)}, \ldots, \hat{x}^{(d)})\in D$, the system of differential equations
$$
\frac{d}{dt}x^v_t = f_i(t,x_1(t), \ldots, x_d(t)), \hbox{ for } i\in [d]
$$
has a unique solution in $D$ for all $x_i:\reals\rightarrow \reals$ such that
$$
x_i(0) = \hat{x}^{(i)} \hbox{ for } i\in [d]
$$
and which extends to points arbitrarily close to the boundary of $D$.
\item The following holds almost surely
$$
Y^{(i)}_t = n x_i(t/n) + o(n)
$$
uniformly for $0 \leq t \leq \min\{\sigma(n) n, m(n)\}$, for all $i\in [d]$, where $x^v_t$ is the solution of the differential equation with $\hat{x}^{(i)}(0) = Y^{(i)}_0 / n$ and $\sigma(n)$ is the supremum of $t$ for which the solution of the differential equation can be extended.
\end{enumerate}
\label{thm:wormald}
\end{theorem}

%%%%%%%%%%%%%%%%%%%%%%%%%%%%%%%%%%%%%

\newpage\onehalfspacing

\setcounter{section}{3}
\setcounter{subsection}{0}

\section*{\textbf{\Large{Chapter~3}}}

\addcontentsline{toc}{section}{Chapter~3. Cooperative Non-stochastic Multi-Armed Bandits}

\bigskip
\bigskip

\textbf{\Large{Cooperative Non-stochastic Multi-Armed Bandits}}

\bigskip
\bigskip
\bigskip

\noindent{In this chapter, we consider the multi-agent version of a multi-armed bandit problem which is one of the most fundamental decision making problems under uncertainty.
In this problem, a learning agent needs to consider the exploration-exploitation trade-off, i.e. balancing the exploration of various actions in order to learn how much rewarding they are and selecting high-rewarding actions. In the multi-agent version of this problem, multiple agents collaborate with each other trying to maximize their individual cumulative rewards, and the challenge is to design efficient cooperative algorithms under communication constraints, e.g. the topology of the network and communication delays.
}
\subsection{Introduction}\label{sec::chapter-aamas-intro}

We consider the non-stochastic (adversarial) multi-armed bandit problem in a cooperative multi-agent setting, with $K\geq 2$ arms and $N\geq 1$ agents. In each time step, each agent selects an arm and then observes the incurred loss corresponding to its selected arm. The losses of arms are according to an arbitrary loss sequence, which is commonly referred to as the non-stochastic or adversarial setting. Each agent observes only the loss of the arm this agent selected in each time step. The agents are allowed to cooperate by exchanging messages, which is constrained by a communication graph $G$ such that any two agents can exchange a message directly between themselves only if they are neighbors in graph $G$. Each exchange of a message over an edge has delay of $d$ time steps. The goal of each agent is to minimize its cumulative loss over a time horizon of $T$ time steps. We study the objective of minimizing the individual regret of agents, i.e. the difference between the expected cumulative loss incurred by an agent and the cumulative loss of the best arm in hindsight. We also study the average regret of all agents.

The multi-agent multi-armed bandit problem formulation that we study captures many systems that use a network of learning agents. For example, in peer-to-peer recommender systems, the agents are users and the arms are products that can be recommended to users \citep{baraglia2013peer}. The delay corresponds to the time it takes for a message to be transmitted between users. Note that in this application scenario, the number of products (i.e. arms) may be much larger than the number of users (i.e. agents).

\sloppy The collaborative multi-agent multi-armed bandit problem was studied, e.g., in \citet{cesa2019delay} and \citet{bar2019individual}, where each edge has unit delay. Our setting is more general in allowing for arbitrary delay $d$ per edge.  \citet{cesa2019delay} showed that when each agent selects arms according to a cooperative Exp3 algorithm (Exp3-Coop), the average regret is $O(\sqrt{(\alpha(G)/N +1/K)\log(K)KT})$ for large enough $T$, where $\alpha(G)$ is the independence number of graph $G$. \citet{bar2019individual} have shown that individual regret of each agent $v$ is $O(\sqrt{(1/|\mathcal{N}(v)|+1/K)\log(K)KT})$ when $T\geq K^2 \log(K)$, where $\mathcal{N}(v)$ is the set of neighbors of agent $v$ and itself in graph $G$. This regret bound is shown to hold for an algorithm where some agents, referred to as center agents, select arms using the Exp3-Coop policy and other agents copy the actions of center agents. These bounds reveal the effect of collaboration on the learning and what graph properties effect the efficiency of learning. However, some fundamental questions still remain. For example, to the best of our knowledge, it is unknown from the previous literature what is the lower bound for this problem. Moreover, it is unknown whether better algorithms can be designed whose regret matches a lower bound under certain conditions. 

In this work, we give a regret lower bound for any learning algorithm in which each agent can only communicate with their neighbors. We present a center-based algorithm whose regret upper bound matches the lower bound when the number of arms is large enough. We present an algorithm that has a regret upper bound with $\sqrt{d}$ dependence on the delay per edge, which is optimal. All our regret bounds are parametrized with the delay parameter $d$, which is unlike to \citet{cesa2019delay} and \citet{bar2019individual} which considered only the special when $d=1$.

\subsubsection{Related work}

The multi-armed bandit problem in a multi-agent setting, where agents collaborate with each other subject to some communication constraints, has received considerable attention in recent years.
\citet{awerbuch2008competitive} introduced the cooperative non-stochastic multi-armed bandit problem setting where communication is through a public channel (corresponding to a complete graph) and some agents may be dishonest.
\citet{kar2011bandit} considered a special collaboration network in which only one agent can observe the loss of the selected arm in each time step. 
 \citet{szorenyi2013gossip} discussed two specific P2P networks in which at each time step, each agent can send messages to only two other agents.
\citet{cesa2020cooperative} studied an online learning problem where only a subset of agents play in each time step. They showed that an optimal average regret bound for this problem is $\Theta(\sqrt{\alpha(G)T})$ when the set of agents that play in each time step is chosen randomly, while $\Omega(T)$ bound holds when the set of agents can be chosen arbitrarily in each time step.
\citet{kolla2018collaborative}, \citet{landgren2016distributed} and \citet{martinez2019decentralized} considered a setting in which communication is constrained by a communication graph such that any two agents can communicate \emph{instantly} if there is an edge connecting them.

The communication model considered in this chapter was introduced by \citet{cesa2019delay}. Here, agents communicate via messages sent over edges of a fixed connected graph and sending a message over an edge incurs a delay of value $d$. \citet{cesa2019delay} considered the case when $d=1$ whereas in this chapter, we consider $d\geq 1$.

They proposed an algorithm, referred to as Exp3-Coop, in which each agent constructs loss estimators for each arm using an importance-weighted estimator.
The Exp3-Coop algorithm has an upper bound of $O(\sqrt{(\alpha(G)/N + 1/K)\log(K)KT} + \log(T))$ on the average regret. 

\citet{bar2019individual} combines the idea of center-based communication from \citet{kolla2018collaborative} with the Exp3-Coop algorithm, showing that the center-based Exp3 algorithm has a regret upper bound of 
$O(\sqrt{(1/|\mathcal{N}(v)|+1/K)\log(K)KT})$ for each individual agent when $d=1$. We show that a better regret bound can be guaranteed with respect to the scaling with the number of arms $K$.

Multi-armed bandits with delayed feedback have been studied extensively in the single-agent setting \citep{joulani2013online, thune2019nonstochastic, flaspohler2021online}. Specifically,
\citet{zimmert2020optimal} considered a setting in which the agent has no prior knowledge about the delays and showed an optimal regret of $O(\sqrt{KT}+\sqrt{d\log(K)T})$ where $d$ is the average delay over $T$ time steps. We present, in the multi-agent setting, a distributed learning algorithm whose regret upper bound can also achieve this optimal $\sqrt{d}$ dependence on the delay per edge $d$.

We use the \emph{Tsallis entropy} family of regularizers proposed by \citet{tsallis1988possible}. \citet{zimmert2021tsallis} have shown that an online mirror descent algorithm with a Tsallis entropy regularizer achieves optimal regret for the single-agent bandit problem. We show a distributed learning algorithm for the multi-agent bandit setting, which uses a Tsallis entropy regularizer.

\subsubsection{Organization of this chapter} 

Section~\ref{sec::chapter-aamas-problem-formulation} provides problem formulation.
Section~\ref{sec::chapter-aamas-algorithms-and-regret-upper-bounds} presents our two algorithms and their regret bounds. In Section~\ref{sec::chapter-aamas-regret-lower-bound}, we present a lower bound on individual regret of each agent.  Section~\ref{sec::chapter-aamas-numerical-experiements} contains numerical results. Finally, proofs of our results are available in the Section~\ref{sec::chapter-aamas-Section}.

\subsection{Problem formulation}
\label{sec::chapter-aamas-problem-formulation}

We consider a multi-armed bandit problem with a finite set  $\mathcal{A}=\{1,\ldots, K\}$ of actions (arms) played by $N$ agents. The agents can communicate through a communication network $G = (\mathcal{V}, \mathcal{E})$
where $\mathcal{V}$ is the set of $N$ agents and $\mathcal{E}$ is the set of edges such that $(u,v)\in \mathcal{E}$ if, and only if, agent $u$ can send/receive messages to/from agent $v$. We denote the neighbors of the agent $v$ and itself by the set $\mathcal{N}(v) = \{u\in \mathcal{V}: (u, v)\in \mathcal{E}\}\cup\{v\}$.
Sending a message over edge $e\in \mathcal{E}$ incurs a delay of value $d_e \geq 0$ time steps. We consider the \emph{homogeneous} setting under which $d_e = d\geq 1$ for every edge $e\in E$.
Note that the delayed communication network model in \citet{cesa2019delay} and \citet{bar2019individual} is restricted to the special case $d=1$. 

%\mv{*** This suggests that \citet{cesa2020cooperative} only considers the case $d=1$ while earlier discussion in the related work section suggests that they allow for arbitrary $d$. ***.}

% \jy{***unit delay per edge***}

At each time step $t = 1,2, \ldots, T$, each agent $v\in \mathcal{V}$ chooses an action $I_t(v)\in \mathcal{A}$ according to distribution $p_t^v$ over $\mathcal{A}$ and then observes the loss value, $\ell_t(I_t(v)) \in [0,1]$. Notice that the loss does not depend on the agent, but only on the time step and the chosen action. Hence, if two agents choose the same action at the same time step, they incur the same loss. We consider the \emph{non-stochastic setting} where the losses are determined by an oblivious adversary, meaning that the losses do not depend on the agent's realized actions. 

At the end of each time step $t$, each agent $v\in \mathcal{V}$ sends a message $S_t(v)$ of size $b_t(v)$ information bits to all its neighbors and after this, each agent $v\in \mathcal{V}$ has messages $\cup_{u\in \mathcal{N}(v)}\{S_{s}(u): s+d = t\}$.

We assume that at each time step $t$, each agent $v$ can send to each of its neighbors a message $S_t(v)=\left\langle v, t, I_{t}(v), \ell_{t}\left(I_{t}(v)\right), p_{t}^{v}\right\rangle$, i.e. the agent id, the time step, the chosen arm id, the instant loss received and the instant action distribution. We denote with $b_t(v)$ the number of information bits to encode $S_t(v)$.
The total communication cost in each time step is $\sum_{v\in \mathcal{V}}\sum_{u:(u,v)\in \mathcal{E}}b_t(u)$ information bits.

The \emph{individual regret} of each agent $v$ is defined as the difference between its expected accumulated loss and the loss of the best action in hindsight, i.e. 
$$
R_{T}^v=\mathbb{E}\left[\sum_{t=1}^{T} \ell_{t}\left(I_{t}(v)\right)\right]-\min _{i \in \mathcal{A}} \sum_{t=1}^{T} \ell_{t}(i).
$$
The \emph{average regret} of $N$ agents is defined as 
$$
R_{T}=\frac{1}{N} \sum_{v \in \mathcal{V}} R_{T}^v.
$$

% \paragraph{Additional notation}
% We define $\mathcal{P}_{K-1}$ to be the $K-1$ simplex.

Let $\alpha(G)$ be the size of a maximal independent set of graph $G$, where the maximal independent set is the largest subset of nodes such that no two nodes in this set are connected by an edge.

\subsection{Algorithms and regret upper bounds}
\label{sec::chapter-aamas-algorithms-and-regret-upper-bounds}

In this section, we propose two collaborative multi-agent bandit algorithms, the center-based cooperative follow-the-regularized-leader (CFTRL) algorithm and the decentralized cooperative follow-the-regularized-leader (DFTRL) algorithm. The first algorithm has optimal regret up to a constant factor when the number of arms is large enough. The second algorith has optimal depence on the delay parameter $d$.

\subsubsection{A center-based cooperative algorithm}

We consider an algorithm where some agents, referred to as \emph{centers}, run a FTRL algorithm, and each other agent copies the action selection distribution from its nearest center. The strategy based on using center agents was proposed in \citet{bar2019individual}, where agents played the Exp3 strategy instead. These centre agents collaboratively update their strategies by exchanging messages with other agents, and each non-center agent copies the strategy of its nearest center agent. The center agents are selected such that they have a sufficiently large degree, which can be shown to reduce individual regret of center agents. Moreover, the center agents are selected such that each non-center agent is within a small distance to a center agent. 

Let $\mathcal{C} \subseteq \mathcal{V}$ be the set of centers. The set of agents $\mathcal{V}$ is partitioned into disjoint components $\mathcal{V}_c$, $c\in \mathcal{C}$.
Each non-center agent $v$ belongs to a unique component. For each agent $v$, let $C(v)$ denote its center agent,  $c=C(v)$ if and only if $v\in \mathcal{V}_{c}$.  
Let $d(v)$ be the distance between a non-center agent $v$ and its center $C(v)$. The set of centers $\mathcal{C}$ and the partitioning $\{\mathcal{V}_c: c\in \mathcal{C}\}$ are computed according to Algorithms~3 and 4 in \citet{bar2019individual}. 

Let $\mathcal{J}_t(v) = \{I_t(v^\prime): v^\prime\in \mathcal{N}(v)\}$ be the set of actions chosen by agent $v$ or its neighbors at time step $t$. 
Each center agent $c\in C$ runs a FTRL algorithm with the collaborative importance-weighted loss estimators observable up to time step $t$, for each action $i$,
$$
\hat{L}^{c,obs}_t(i) = \sum_{s=1}^{t-1}\hat{\ell}^{c,obs}_s(i)
$$
and
$$
\hat{\ell}_{t}^{c,obs}(i)= 
\left\{\begin{array}{cl}
     \frac{\ell_{t-d}(i)}{q_{t-d}^c(i)} \mathbb{I}\left\{i\in\mathcal{J}_{t-d}(c)\right\}  & \text { if } t> d \\
     0 & \text { otherwise }
    \end{array}\right.
$$
where 
$$
q_t^c(i) = 1-\prod_{v \in \mathcal{N}(c)}\left(1-p_t^{v}(i)\right)
$$
is the \emph{neighborhood-aggregated importance weight}.

In each time step, the center agents update their action selection distributions according to the FTRL algorithm, i.e.
$$
p_t^c=\operatorname{argmin}_{p\in \mathcal{P}_{K-1}} \left\{ \left\langle p, \hat{L}_{t}^{c, obs}\right\rangle + F_t(p)\right\}.
$$
where $F_t(p)$ is the \emph{Tsallis entropy regularizer}
\citep{zimmert2021tsallis} with the learning rate $\eta(c)$, 
\begin{equation}
    \label{eq::tsallis-entropy}
    F_t(p) = -2\sum_{i=1}^K\sqrt{p_i} / \eta(c).
\end{equation}

Each non-center agent $v\in \mathcal{V}\backslash C$ selects actions according to the uniform distribution until time step $t>d(v)d$.
Then the non-center agent copies the action selection distribution from its center, i.e.
$p^v_{t} = p^{C(v)}_{t-d(v)d}.$

The details of the CFTRL algorithm are described in Algorithm \ref{algo:CFTRL}.

\begin{algorithm}[t]
\caption{Center-based cooperative FTRL (CFTRL)}\label{algo:CFTRL}
\SetKwInOut{Init}{Initialization}
\SetKwInOut{Input}{Input}
\Input{Tsallis regularizer Eq.~(\ref{eq::tsallis-entropy}), learning rate $\eta(c)$ and the delay $d$.} 
\Init{$\hat{L}_1^{c, obs}(i) = 0$ for all $i\in \mathcal{A}$ and $c\in C$, $p_1^v(i) = 1/K$ for all $i\in \mathcal{A}$ and $v\not\in C$.}
\For{each time step $t=1,2,\dots, T$}{

    Each $c\in \mathcal{C}$ updates $p_t^c= \operatorname{argmin}_{p\in \mathcal{P}_{K-1}}\{ \langle p, \hat{L}_{t}^{c, obs}\rangle + F_t(p)\} 
    $\;
    Each $c\in \mathcal{C}$ chooses $I_t(c)=i$ with probability $p_t^c(i)$ and receives the loss $\ell_t(I_t(c))$\;
    Each $c\in \mathcal{C}$ sends the message $S_t(c) = \left\langle c, t,  I_t(c), \ell_t(I_t(c)), p_t^c\right\rangle$ to all their neighbors\;
    Each $c\in \mathcal{C}$ receives messages $\{S_{t-d}(v): v\in \mathcal{N}(c)\}$ and computes $\hat{L}_{t+1}^{c, obs}$\;
    Each $v\in \mathcal{V}\setminus \mathcal{C}$ updates $p_{t}^v = p_{t-d(v)d}^{C(v)}$ when $t>d(v)d$ and $p_{t}^v = p_{1}^{v}$ otherwise\;
    Each $v\in V\setminus \mathcal{C}$ chooses $I_t(v)=i$ with probability $p_t^v(i)$ and receives $\ell_t(I_t(v))$\;
    Each $v\in \mathcal{V}\setminus \mathcal{C}$ sends $S_t(v) = \left\langle v, t,   I_t(v), \ell_t(I_t(v)), p_t^v\right\rangle$ to all its neighbors.
}
\end{algorithm}

\paragraph{Individual regret upper bound}

We show an individual regret upper bound for Algorithm~\ref{algo:CFTRL} in the following theorem.

\begin{theorem}\label{thm::CFTRL}

Assume that $K\geq \max_{v\in \mathcal{V}}|\mathcal{N}(v)|$ and $T\geq 36(d+1)^2K\max_{c}|\mathcal{N}(c)|$, and agents follow the CFTRL algorithm with each center agent $c\in \mathcal{C}$ using the learning rate $\eta(c) = \sqrt{|\mathcal{N}(c)| / (3T)}$. Then, the individual regret of each agent $v\in \mathcal{V}$ is bounded as 
$$
R_T^v = O\left(\frac{1}{ \sqrt{|\mathcal{N}(v)|}} \sqrt{KT} + d\log(K)\sqrt{|\mathcal{N}(C(v))|T}\right).
$$
\end{theorem}
The proof of the theorem is provided in Section~\ref{sec::chapter-aamas-center-based}.
 
In the following, we provide a proof sketch. The proof relies on two key lemmas which are shown next.

\begin{lemma}
\label{lm:ftrl-coop:individual}
Assume that the delay of each edge is $d\geq 1$, then the individual regret of each center agent $v$ with the regularizer $F_t(p) = \sum_{i=1}^K f_t(p_i)$ satisfies
\begin{eqnarray*}
R_{T}^v & \leq & M + \frac{1}{2} \mathbb{E} \left[\sum_{t=1}^T  \sum_{i\in \mathcal{A}} \frac{1}{q_t^v(i) f_t^{\prime\prime}(p_t^v(i))}\right] \\
&& + d \cdot \mathbb{E} \left[\sum_{t=1}^T \sum_{i\in \mathcal{A}} \frac{1}{f_t^{\prime\prime}(p_t^v(i))}\right]
\end{eqnarray*}
where 
$$
M = \max_{x \in \mathcal{P}_{K-1}}-F_{1}(x)+\sum_{t=2}^{T} \max_{x \in \mathcal{P}_{K-1}}\left(F_{t-1}(x)-F_{t}(x)\right).
$$
\end{lemma}
\begin{lemma}
\label{lm:ftrl-tsallis:selection-prob}

For any $\delta>1$, assume that agent $v$ runs a FTRL algorithm
with Tsallis entropy (\ref{eq::tsallis-entropy}) and learning rate $\eta(v)\leq (1-1/\sqrt{\delta})/(\delta^{3d/2}\sqrt{K})$ and $K\geq 2$, then for all $t\geq 1$ and $i\in \mathcal{A}$
$$
(1-(1+\delta)\eta(v)\hat{\ell}_t^{v,obs}(i)) p_t^v(i) \leq p_{t+1}^v(i) \leq \delta p_{t}^v(i).
$$
\end{lemma}
The proofs of the two lemmas are provided in Section~\ref{sec::chapter-aamas-upper-bound-lemma-3} and \ref{sec::chapter-aamas-sandwich}. 
Similar property as in Lemma~\ref{lm:ftrl-tsallis:selection-prob} was known  to hold for the Exp3 algorithm by a result in \citet{cesa2019delay}. To the best of our knowledge, this property was previously not known to hold for the FTRL algorithm with Tsallis entropy.

Lemma~\ref{lm:ftrl-coop:individual} bounds the individual regret by the sum of a constant, the regret due to the \emph{instant} bandit feedback, and the regret due to the \emph{delayed} full-information feedback, which is $O(\sqrt{KT/|\mathcal{N}(c)|})$.
Since the action selection distributions of non-center agents are copied from their centers in the past rounds, Lemma~\ref{lm:ftrl-tsallis:selection-prob} bounds the difference between action selection distributions of non-center agent $v$ and its center $C(v)$ in the same rounds when $T>d(v)d$.
Consequently, the difference between the individual regret of a non-center agent $v$ and its center $C(v)$ is bounded by $O(d(v)d \eta(C(v)) T) = O(d\log(K) \sqrt{|\mathcal{N}(C(v))|T})$.

\subsubsection{A decentralized cooperative algorithm}

Theorem~\ref{thm::CFTRL} provides a bound for individual regrets, which increases linearly in the edge-delay parameter $d$. This can be problematic when the delay in the communication network is large.
We show that the effect of delays on regret can be reduced by using a decentralized follow-the-regularized-leader (DFTRL) algorithm.

In the DFTRL algorithm, each agent runs a FTRL algorithm 
 
with
a \emph{hybrid regularizer} $F_t(p)$ defined in \citet{zimmert2020optimal} as follows
\begin{equation}
\label{eq::hybrid-regularizer}
F_t(p) = \sum_{i=1}^{K} \left( \frac{- 2  \sqrt{p_{i}}}{\eta_t} + \frac{p_i \log(p_i)}{\zeta_t} \right)
\end{equation}
where $\eta_t$ and $\zeta_t$ are some non-increasing sequences.

As is shown in Theorem~\ref{thm:lower-2}, there is a regret lower bound that consists of two parts: the first part is the regret lower bound of the multi-armed bandit problem and the second part is the regret lower bound of the bandit problem with full-information but delayed feedback \citep{joulani2013online}. 
The hybrid regularizer combines the Tsallis entropy regularizer with an optimal regularizer in the full-information setting, the negative entropy regularizer.
The learning rates of the two regularizers can be tuned separately to minimize the regret from the two parts.

The details of the DFTRL are described in Algorithm~\ref{algo:DFTRL}.
\begin{algorithm}[t]
\caption{Decentralized cooperative FTRL (DFTRL)}\label{algo:DFTRL}
\SetKwInOut{Init}{Initialization}
\SetKwInOut{Input}{Input}
\Input{Hybrid regularizer Eq.~(\ref{eq::hybrid-regularizer}), learning rates $\eta_t$, $\zeta_t$, and delay $d$.}
\Init{$\hat{L}_1^{v, obs}(i) = 0$ for all $i\in \mathcal{A}$ and $v\in \mathcal{V}$.}
\For{each time step $t=1,2,\dots, T$}{
    Each $v\in \mathcal{V}$ updates $
    p_t^v=\operatorname{argmin}_{p\in \mathcal{P}_{K-1}}\{\langle p, \hat{L}_{t}^{v, obs}\rangle + F_t(p)\}
    $\;
    Each $v\in \mathcal{V}$ chooses $I_t(v)=i$ with probability $p_t^v(i)$ and receives the loss $\ell_t(I_t(v))$\;
    Each $v\in \mathcal{V}$ sends the message $S_t(v) = \left\langle v, t,  I_t(v), \ell_t(I_t(v)), p_t^v\right\rangle$ to all their neighbors\;
    Each $v\in \mathcal{V}$ receives messages $\{S_{t-d}(v^\prime): v^\prime\in \mathcal{N}(v)\}$ and computes $\hat{L}_{t+1}^{v, obs}$\;
}
\end{algorithm}

\paragraph{Average regret upper bound}

We show a bound on the average regret for the DFTRL algorithm in the following theorem. The sequences $\eta_t$ and $\zeta_t$ are assumed to be set as
$$
\eta_t = (1/(1-1/e)) (\alpha(G)/ N + 1/K)^{-1/4} \sqrt{2/T}
$$
and 
$$
\zeta_t = \sqrt{\log(K) / (dt)}.
$$

\begin{theorem}\label{thm::DFTRL} 

Assume that each agent follows the DFTRL algorithm

and the delay of each edge is $d\geq 1$, then the average regret over $N$ agents is bounded as

$$
R_{T} = O\left( \left(\frac{\alpha(G)}{N} + \frac{1}{K}\right)^{1/4}\sqrt{KT} +  \sqrt{d\log(K)T} \right).
$$
\end{theorem}

Proof of the theorem is provided in Section~\ref{sec::chapter-aamas-upper-bound-theorem-1}. 
We note that the average regret scales as $\sqrt{d}$ which is better than linear scaling of the CFTRL algorithm.

For the special case when $d=1$, as in  \citet{cesa2019delay},Theorem~\ref{thm::DFTRL}
shows that when the number of arms $K$ is large enough, then the DFTRL algorithm has an $O((\alpha(G)/N)^{1/4}\sqrt{KT})$ regret, which is better than $O((\alpha(G)/N)^{1/2}\sqrt{KT}\sqrt{\log(K)})$ of Exp3-Coop from \citet{cesa2019delay}. Specifically, this improvement holds when
$
K = \Omega(\exp\left(\sqrt{N/\alpha(G)}\right)).
$

In what follows we provide a proof sketch of Theorem~\ref{thm::DFTRL}.
First we present a key lemma whose proof is provided in Section~\ref{sec::chapter-aamas-upper-bound-lemma-2}.
\begin{lemma}
\label{lm:ftrl-coop-independence}
For every agent $v\in \mathcal{A}$ and any probability distribution $p^v$ over $\mathcal{A}$, it holds
$$
\sum_{i\in \mathcal{A}}\sum_{v \in \mathcal{V}} \frac{p^v(i)^{\frac{3}{2}}}{q^v(i)} \leq 
N\sqrt{\frac{1}{1-1/e}\left(\frac{\alpha(G)}{N}+\frac{1}{K}\right)K}
$$
where $q^v(i) = 1-\prod_{v^\prime \in \mathcal{N}(v)}(1-p^{v^\prime}(i))$.
\end{lemma}

Lemma~\ref{lm:ftrl-coop-independence} shows that the average regret from the FTRL algorithm with the hybrid regularizer with instant feedback is $O((\alpha(G)/N + 1/K)^{1/4}\sqrt{KT})$.
For the regularizer in Eq.~(\ref{eq::hybrid-regularizer}), the delay effect term is $O(\sqrt{dT\log(K)})$.
Lemma~\ref{lm:ftrl-coop:individual} bounds the average regret by the sum of two terms.

\subsection{Regret lower bounds}
\label{sec::chapter-aamas-regret-lower-bound}

We present lower bounds on individual regret $R_T^v$ for every agent $v\in \mathcal{V}$ and average regret $R_T$. 

\begin{theorem} \label{thm:lower-2} 
The worst-case individual regret of each agent $v\in \mathcal{V}$, $R_T^v$, is
$$
\Omega\left(\max\left\{ \min \left\{T, \frac{1}{\sqrt{|\mathcal{N}(v)|}}\sqrt{KT}\right\}, \sqrt{d \log(K)T}\right\} \right)
$$
and the worst-case average regret, $R_T$, is
$$
\Omega\left( \max\left\{ \min \left\{T, c_G\sqrt{KT}\right\}, \sqrt{d \log(K)T}\right\} \right)
$$
where $c_G = (1/N)\sum_{v\in \mathcal{V}}1/\sqrt{|\mathcal{N}(v)|}$.
\end{theorem}

The proof is provided in Section~\ref{sec::chapter-aamas-lower-bound}. 
The lower bounds contain two parts. The first part is derived from the lower bounds in \citet{shamir2014fundamental} for a class of online algorithms. The second part handles the effect of delays by showing that the individual regret of each agent cannot be smaller than the regret of a single agent with delayed full information.

We note that the individual regret of Algorithm~\ref{algo:CFTRL} is optimal with respect to scaling with the number of arms $K$ and the average regret of Algorithm~\ref{algo:DFTRL} is optimal with respect to scaling with delay $d$.

\subsection{Numerical experiments}
\label{sec::chapter-aamas-numerical-experiements}

In this section, we present results of numerical experiments whose goal is to compare performance of CFTRL and DFTRL algorithms with some state-of-the-art algorithms and demonstrate the tightness of our theoretical bounds. We consider the classic stochastic multi-armed bandit problem with agents communicating via different networks.

The stochastic multi-armed bandit problem is defined as follows: each arm $i$ is associated with a Bernoulli distribution with mean $\mu_i$ for $i=1,2,\dots, K$. The loss $\ell_t(i)$ from choosing arm $i$ at time step $t$ is sampled independently from the corresponding Bernoulli distribution.
In our experiments, we set $\mu_i = (1+8(i-1)/(K-1))/10$ so that $\mu_1, \dots, \mu_K$ is a linearly decreasing sequence.
Each problem instance is specified by a tuple $(K, G, d)$. The two baseline algorithms we choose are the center-based Exp3 algorithm in \citet{bar2019individual} and the Exp3-Coop algorithm in \citet{cesa2019delay} whose regret upper bounds are suboptimal as discussed in the introduction.

The numerical results are for four experiments whose goals are as follows:
\begin{itemize}
    \item the first experiment compares the performance of CFTRL, DFTRL and the baselines when the number of arms increases,
    \item the second experiment validates the effect of the graph degree in the regret upper bound on CFTRL,
    \item the third experiment validates the effect of the delay parameter on the regret upper bounds of CFTRL and DFTRL, and, finally,
    \item the fourth experiment compares CFTRL and the center-based Exp3 algorithm on some sparse random graphs.
\end{itemize}

In summary, our numerical results validate theoretical results and demonstrate that CFTRL and DFTRL can achieve significant performance gains over some previously proposed algorithms.
The code for producing our experimental results is available online in the GitHub repository: \href{https://anonymous.4open.science/r/On-Regret-Optimal-Cooperative-Non-stochastic-Multi-armed-Bandits-4C76/}{[link]}.

\begin{figure}[t]
    \centering
    \includegraphics[width=.4\textwidth]{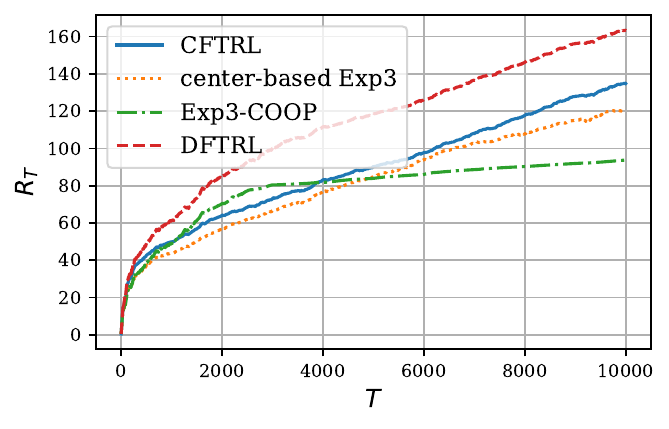}
    \includegraphics[width=.4\textwidth]{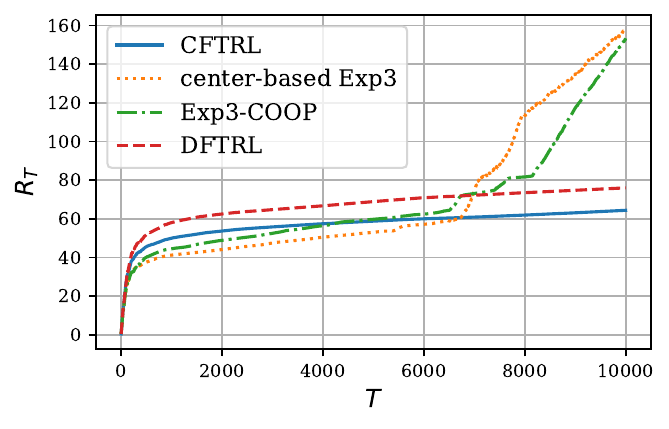}
    \includegraphics[width=.4\textwidth]{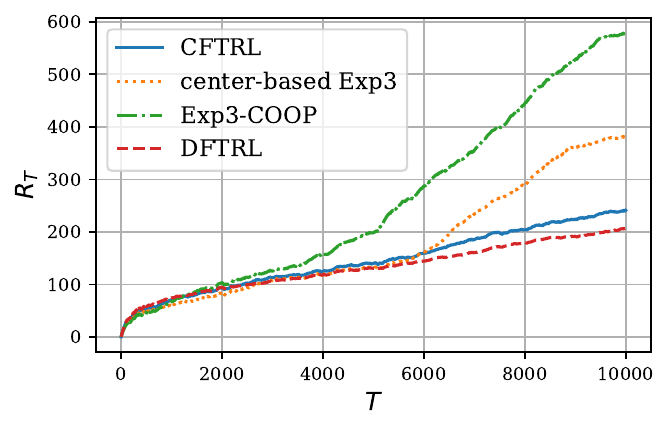}
    \caption{Average regret $R_T$ versus $T$ for different algorithms on a $2$-regular graph with $N=3$ agents and edge-delay $d=1$, and varied number of arms: (top) $K=20$ (middle) $K=30$, and (top) $K=40$. We used 10 independent simulation runs.
    }
    \label{fig:-aamas-regret}
\end{figure}

\subsubsection{The effect of the number of arms}

In the first experiment, we evaluate the performance of CFTRL and DFTRL against the baselines on a $r$-regular graph (all nodes have the same degree of $r$). Note that in a regular graph, each agent has equal probability to be a center agent. For a $r$-regular graph, CFTRL has an individual regret upper bound of $O(\sqrt{(1/r)}\sqrt{KT})$ and DFTRL has an average regret upper bound of $O(\sqrt[4]{1-r/N}\sqrt{KT})$ when the number of arms $K$ is large enough according to our analysis. 

We first demonstrate numerical results showing that CFTRL and DFTRL can achieve significant performance gains over the center-based Exp3 algorithm whose individual regret upper bound is $O(\sqrt{(1/r)\log(K)}\sqrt{KT})$ and the Exp3-Coop algorithm whose average regret upper bound is $O(\sqrt{(1-r/N)\log(K)}\sqrt{KT})$ when the number of arms $K$ is large enough.

Figure~\ref{fig:-aamas-regret} shows the regret $R_T$ versus $T$ for different number of arms, namely $20$, $30$ and $40$. The results demonstrate that CFTRL and DFTRL achieve better regret than Exp3-COOP and center-based Exp3 when the number of arms is large enough.

\subsubsection{The effect of graph degree on CFTRL}

\begin{figure}[t!]
    \centering
    \includegraphics[width=.4\textwidth]{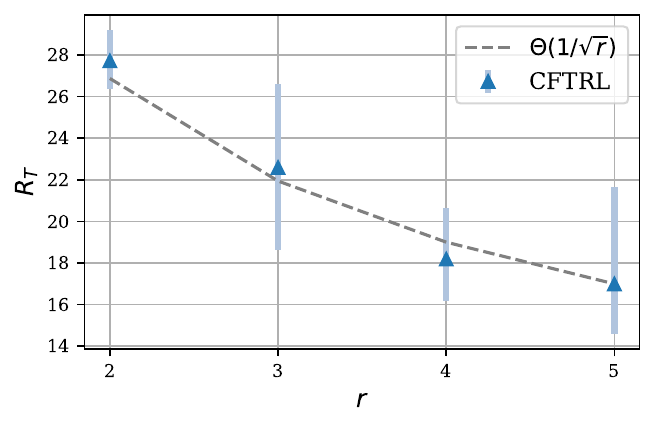}
    \caption{Average regret of CFTRL versus graph degree $r$, for a $r$-regular graph with $N=6$ nodes, $K = 10$ and $d = 1$.}
    \label{fig:-aamas-degree}
\end{figure}

In the second experiment, we validate the scaling of the graph degree in the regret upper bound of CFTRL.
On the $r$-regular graph, CFTRL has a regret that scales as $O(1/\sqrt{r})$ according to Theorem~\ref{thm::CFTRL}.
We run CFTRL on the problem instances with fixed number of arms $K$, delay $d$ and increasing node degree $r$.
The results in Figure~\ref{fig:-aamas-degree} shows that the averaged regret decreases as the graph degree increases and the rate of decrease is approximately $O(1/\sqrt{r})$.

\subsubsection{The effect of delay on CFTRL and DFTRL}

\begin{figure}[t!]
    \centering
    \includegraphics[width=.4\textwidth]{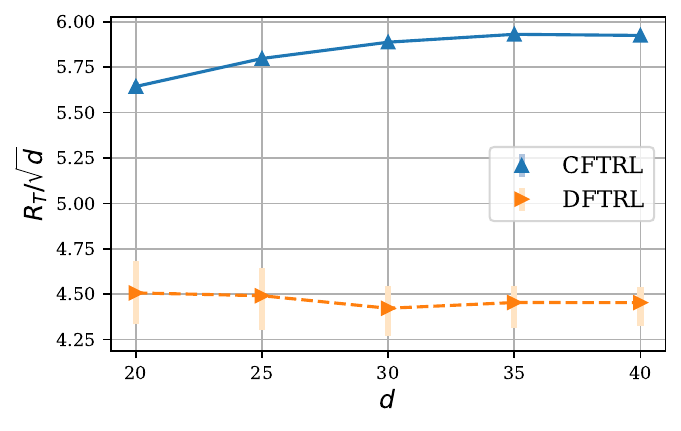}
    \caption{Average regret of CFTRL and DFTRL versus the edge-delay $d$ on a star regular graph with $N = 20$ and $K = 3$.
    }
    \label{fig:-aamas-delay}
\end{figure}

In the third experiment, we run CFTRL and DFTRL algorithms on a fixed star graph $G$ with a fixed number of arms $K$ and varied edge-delay $d$ . Figure~\ref{fig:-aamas-delay} shows that the normalized regret of CFTRL is $R_T / \sqrt{d} = O(\sqrt{d})$ while the normalized regret of DFTRL is $R_T / \sqrt{d} = O(1)$. Hence, when the delay $d$ is large enough, CFTRL has a linearly increasing regret with respect to $d$ which is in contrast to the sub-linear increasing regret of DFTRL.
This is consistent with our theoretical analysis, which states that CFTRL has a regret upper bound of $O(d)$ and DFTRL has a regret upper bound of $O(\sqrt{d})$. 

\subsubsection{The effect of graph sparsity}

\begin{figure}
    \centering
    \includegraphics[width=.4\textwidth]{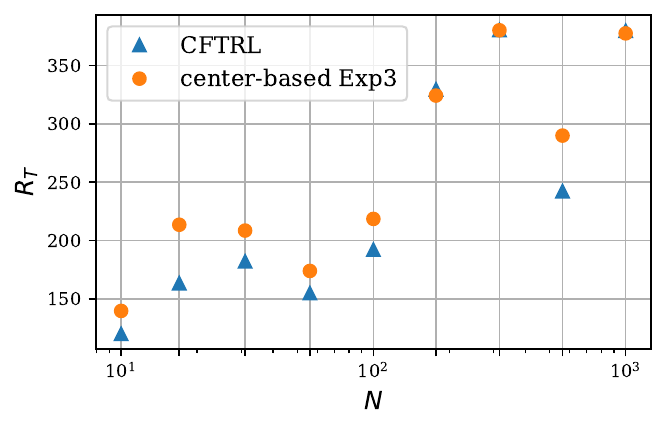}
    \caption{Average regret $R_T$ versus the number of nodes $N$ for sparse Erdős–Rényi random graphs, for CFTRL and center-based Exp3 algorithms.}
    \label{fig:-aamas-erdos}
\end{figure}

In the fourth experiment, we validate that our CFTRL algorithm can outperform the center-based Exp3 algorithm on some random graphs.
We consider Erdős–Rényi random graphs of $N$ nodes with probability of an edge equal to $2\log(N)/N$. This condition ensures that the graph is connected 
and 
$|\mathcal{N}(v)| = O(\log(N)))$ for all $v\in \mathcal{V}$, almost surely \citep[Corollary 8.2]{blum2020foundations}. 
This random graph allows us to evaluate performance of algorithms for a large sparse random graph. We fix $K$ and $d$ and vary the number of nodes $N$ and compare the performance of CFTRL and the center-based Exp3 algorithm on these graphs. 
We fix $T$ to $1000$ time steps.
According to our analysis, CFTRL has a lower individual regret bound than the center-based Exp3 algorithm 
when the number of arms is large enough relative to the number of agents.  
The results in Figure~\ref{fig:-aamas-erdos} indicate that CFTRL has at least as good performance as the center-based Exp3 algorithm when $K$ varies, and can have significantly better performance when the number of arms is large relative to the number of agents.

%%%%%%%%%%%%%%%%%%%%%%%%%%%%%%%%%%%%%%%%%%%%%%%%%%%%%%%%%%%%

\subsection{Proofs}
\label{sec::chapter-aamas-Section}

\subsubsection{An auxiliary lemma}

We show and prove the next lemma which is used for proving some other lemmas.

\begin{lemma}
\label{lm:conditional_exp}
For all $1\leq s \leq t$, $v\in \mathcal{V}$ and $i\in \mathcal{A}$, it holds
\begin{enumerate}
    \item $\mathbb{E}_{s}[\hat{\ell}_t^v(i)] = \ell_t(i)$
    \item $\mathbb{E}_t[\hat{\ell}_t^v(i)^2] = \ell_t(i)^2 / q_t^v(i)$, and 
    \item $\mathbb{E}_t[\hat{\ell}_t^v(i) \hat{L}_{t}^{v,miss}(i)] = \ell_t(i) \hat{L}_{t}^{v,miss}(i)$.
\end{enumerate}
\end{lemma}

\begin{proof}
Consider an arbitrary $v\in \mathcal{V}$. By definition, $p_t^v$ is determined by $\hat{L}_{t}^{v, obs}$ which depends on the realization of $I_1(v), I_2(v),\dots,I_{t-d-1}(v)$. Note that the following equations hold, for all $i\in \mathcal{A}$,
\begin{equation*}
\begin{aligned}
    \mathbb{E}_{t}[\hat{\ell}_t^v(i)] &= \frac{\ell_t(i)}{q_t^v(i)}\mathbb{E}_{t}\left[ \mathbb{I}\left\{i\in\mathcal{J}_t(v)\right\}\right] = \frac{\ell_t(i)}{q_t^v(i)}q_t^v(i) = \ell_t(i) \\
    \mathbb{E}_{t}[\hat{\ell}_t^v(i)^2] &= \frac{\ell_t(i)^2}{q_t^v(i)^2}\mathbb{E}_{t}\left[ \mathbb{I}\left\{i\in\mathcal{J}_t(v)\right\}\right] = \frac{\ell_t(i)^2}{q_t^v(i)^2}q_t^v(i) = \frac{\ell_t(i)^2}{q_t^v(i)}.  
\end{aligned}
\end{equation*}
By the tower law of conditional expectation, we have  $\mathbb{E}_{s}[\hat{\ell}_t^v(i)] = \ell_t(i)$ for all $s\leq t$. Note that $\hat{L}^{v,obs}_t(i) = \sum_{s=1}^{t-d-1}\hat{\ell}^{v, obs}_s(i)$, thus, $\hat{L}_{t}^{v,miss}(i) = \sum_{s=t-d}^{t-1}\hat{\ell}_s^v(i)$, which depends on the realization of $I_1(v), I_2(v),\dots,I_{t-1}(v)$ for all $v\in \mathcal{V}$. Hence, for all $i\in \mathcal{A}$, we have
\begin{equation*}
    \mathbb{E}_t[\hat{\ell}_t^v(i) \hat{L}_{t}^{v,miss}(i)] =  \mathbb{E}_t[\hat{\ell}_t^v(i)]\hat{L}_{t}^{v,miss}(i) = \ell_t(i)\hat{L}_{t}^{v,miss}(i).
\end{equation*}
\end{proof}

\subsubsection{Proof of Lemma~\ref{lm:ftrl-coop:individual}}
\label{sec::chapter-aamas-upper-bound-lemma-3}

Let $i^\ast = \arg\min_{i\in\mathcal{A}} \sum_{t=1}^T \ell_t(i)$ be the best action in hindsight. Consider an arbitrary agent $v\in \mathcal{V}$.
Following arguments in Theorem~3 from \cite{zimmert2020optimal}, we have

\begin{equation*}
\begin{aligned}
R_{T}^v
&=\mathbb{E}\left[\sum_{t=1}^{T} \ell_{t}(I_t(v))-\sum_{t=1}^{T}\ell_{t}(i^\ast)\right] \\
&=  \mathbb{E}\left[\sum_{t=1}^T\left\langle p_t^v, \hat{\ell}_{t}(v)\right\rangle-\hat{L}_{T+1}^v(i^\ast)\right]\\
&= \mathbb{E}\left[\alpha + \beta + \gamma \right] \\
\end{aligned}
\end{equation*}
where 
\begin{equation*}
\begin{aligned}
\alpha &= \sum_{t=1}^T\left[\bar{F}_{t}^{*}\left(-\hat{L}_{t}^{v,obs}-\hat{\ell}_t^v\right)-\bar{F}_{t}^{*}\left(-\hat{L}_{t}^{v,obs}\right)+\left\langle p_t^v, \hat{\ell}_t^v\right\rangle\right] \\
\beta &=  \sum_{t=1}^T\left[\bar{F}_{t}^{*}\left(-\hat{L}_{t}^{v,obs}\right)-\bar{F}_{t}^{*}\left(-\hat{L}_{t}^{v,obs}-\hat{\ell}_t^v\right) -\bar{F}_{t}^{*}\left(-\hat{L}_t^v\right) +\bar{F}_{t}^{*}\left(-\hat{L}_{t+1}^v\right)\right]\\
\gamma &= \sum_{t=1}^T\left[\bar{F}_{t}^{*}\left(-\hat{L}_t^v\right)-\bar{F}_{t}^{*}\left(-\hat{L}_{t+1}^v\right)\right]-\hat{L}_{T+1}^v(i^{*}).
\end{aligned}
\end{equation*}

Next, we bound $\alpha, \beta$ and $\gamma$, as follows.

First, for $\alpha$ we have
\begin{equation*}
    \begin{split}
        \mathbb{E}\left[\alpha\right] 
        & \leq \frac{1}{2} \mathbb{E}\left[\sum_{t=1}^T \sum_{i\in \mathcal{J}_t(v)} \frac{\hat{\ell}_t^v(i)^2}{f^{\prime\prime}_t(p_t^v(i))} \right] \\
        & 
\leq \frac{1}{2} \mathbb{E}\left[\sum_{t=1}^T \sum_{i\in \mathcal{A}} \frac{\hat{\ell}_t^v(i)^2}{f^{\prime\prime}_t(p_t^v(i))} \right] \\
& \leq \frac{1}{2} \mathbb{E}\left[\sum_{t=1}^T \sum_{i\in \mathcal{A}} \frac{1}{q_t^v(i)f^{\prime\prime}_t(p_t^v(i))}\right]
    \end{split}
\end{equation*}

% \begin{equation*}
% \begin{aligned}
% \mathbb{E}\left[\alpha\right] 
% &\leq \mathbb{E}\left[\sum_{t=1}^T \sum_{i\in \mathcal{J}_t(v)} D_{f^\ast}\left(f^\prime_t(p_t^v(i))-\hat{\ell}_t^v(i), f^\prime_t(p_t^v(i))\right)\right] \\
% &\leq \frac{1}{2} \mathbb{E}\left[\sum_{t=1}^T \sum_{i\in \mathcal{J}_t(v)} \frac{\hat{\ell}_t^v(i)^2}{f^{\prime\prime}_t(p_t^v(i))} \right] \\
% &\leq \frac{1}{2} \mathbb{E}\left[\sum_{t=1}^T \sum_{i\in \mathcal{A}} \frac{\hat{\ell}_t^v(i)^2}{f^{\prime\prime}_t(p_t^v(i))} \right] \\
% &\leq \frac{1}{2} \mathbb{E}\left[\sum_{t=1}^T \sum_{i\in \mathcal{A}} \frac{1}{q_t^v(i)f^{\prime\prime}_t(p_t^v(i))}\right]
% \end{aligned}
% \end{equation*}
where the first inequality comes from the proof in Lemma~1 in \cite{zimmert2020optimal} and the last inequality comes from Lemma \ref{lm:conditional_exp}.

Second, by Lemma~3 in \cite{zimmert2020optimal}, 
\begin{eqnarray*}
&&    \bar{F}_{t}^{*}\left(-\hat{L}_{t}^{v,obs}\right)-\bar{F}_{t}^{*}\left(-\hat{L}_{t}^{v,obs}-\hat{\ell}_t^v\right) -\bar{F}_{t}^{*}\left(-\hat{L}_t^v\right) +\bar{F}_{t}^{*}\left(-\hat{L}_{t+1}^v\right)\\
&\leq & \sum_{i\in \mathcal{J}_t(v)}
\int_{0}^{1} \hat{\ell}_t^v(i) f_{t}^{\prime \prime}\left(p_{t}^v(i)\right)^{-1} \hat{L}_{t}^{v,miss}(i) d x \\
&=& \sum_{i\in \mathcal{J}_t(v)} \hat{\ell}_t^v(i) f_{t}^{\prime \prime}\left(p_{t}^v(i)\right)^{-1} \hat{L}_{t}^{v,miss}(i).
\end{eqnarray*}
Hence, for $\beta$ we have 
\begin{equation*}
    \begin{split}
        \mathbb{E}\left[\beta\right]  
\leq \mathbb{E}\left[ \sum_{t=1}^T \sum_{i\in \mathcal{A}} \frac{\ell_t(i)\hat{L}_{t}^{v,miss}(i)}{f^{\prime\prime}_t(p_t^v(i))}  \right] 
 & \leq \mathbb{E}\left[ \sum_{t=1}^T \sum_{i\in \mathcal{A}} \frac{\sum_{s=1}^{d}\hat{\ell}_{t-s}^v(i)}{f^{\prime\prime}_t(p_t^v(i))} \right] \\
 & \leq d \sum_{t=1}^T \mathbb{E}\left[ \sum_{i\in \mathcal{A}} \frac{1}{f^{\prime\prime}_t(p_t^v(i))}\right]
    \end{split}
\end{equation*}
where the first and third inequalities come from Lemma \ref{lm:conditional_exp}.

Finally, by Lemma 2 in \cite{zimmert2020optimal}, we have

\begin{equation*}
\begin{aligned}
\gamma &\leq \max_{x \in \mathcal{P}_{K-1}}-F_{1}(x)+\sum_{t=2}^{T} \max_{x \in \mathcal{P}_{K-1}}\left(F_{t-1}(x)-F_{t}(x)\right). 
\end{aligned}
\end{equation*}

\subsubsection{Proof of Lemma~\ref{lm:ftrl-tsallis:selection-prob}}
\label{sec::chapter-aamas-sandwich}

Consider an arbitrary agent $v\in \mathcal{V}$. The arm selection distribution $p^v_t$ is the solution of the following convex optimization problem:
$$
\begin{array}{rl}
\hbox{minimize} & \eta(v) \sum_{s=1}^{t-1} \left<x,\hat{\ell}_{t}^{v, obs}\right> + \left(\sum_{i=1}^K -2\sqrt{x_i}\right)  \\
\hbox{over}     & x\in \mathbf{R}^K\\
\hbox{subject to} & x_i \geq 0, \hbox{ for all } i\in \mathcal{A}\\
& \sum_{i=1}^K x_i = 1.
\end{array}
$$

The distribution $p_t^v$ satisfies
\begin{equation}
p_t^v(i) = \frac{1}{\left(\eta(v) \sum_{s=1}^{t-1}\hat{\ell}_{t}^{v, obs}(i) + \lambda_t^v\right)^2}, \hbox{ for } i\in \mathcal{A}
\label{equ:pt}
\end{equation}
where $\lambda_t^v\geq 0$ is the Lagrange multiplier associated with the constraint $\sum_{i=1}^K p_t^v(i) = 1$, which is the solution of the following fixed-point equation
\begin{equation}
\sum_{i=1}^K \frac{1}{\left(\eta(v)\sum_{s=1}^{t-1}\hat{\ell}_{t}^{v, obs}(i) + \lambda_t^v\right)^2} = 1.
\label{equ:lambdat}
\end{equation}

From (\ref{equ:lambdat}) it follows that $\lambda_t^v$ is a decreasing sequence in $t$. To see this, note
$$
\sum_{i=1}^K \frac{1}{\left(\eta(v)\sum_{s=1}^{t-1}\hat{\ell}_{t}^{v, obs}(i) + \lambda_t^v + (\eta(v)\hat{\ell}_t^{v,obs}(i)+\lambda_{t+1}^v-\lambda_t^v)\right)^2} = 1
$$
If $\lambda_{t+1}^v - \lambda_t^v > 0$, then the left-hand side in the last equation is less than $1$ and, hence, the equation cannot hold. Since $\hat{\ell}_t^{v,obs}(i) > 0$ if, and only if, $I_t(v) = i$, it follows that $\hat{\ell}_t^{v,obs}(I_t(v)) + \lambda_{t+1}^v-\lambda_t^v \geq 0$ and $\lambda_{t+1}^v\leq \lambda_t^v$. 

From (\ref{equ:pt}), for all $i\in \mathcal{A}$,
$$
p_{t+1}^v(i) = \frac{p_t^v(i)}{\left(1+\sqrt{p_t^v(i)}(\eta(v)\hat{\ell}_t^{v,obs}(i)+\lambda_{t+1}^v-\lambda_t^v)\right)^2}.
$$
From this and the fact $\hat{\ell}_t^{v,obs}(I_t(v)) + \lambda_{t+1}^v-\lambda_t^v \geq 0$, it follows 
$$
p_{t+1}^v(I_t(v))\leq p_t^v(I_t(v)).
$$

Let $\gamma = \lambda_{t+1}^v-\lambda_t^v$. It holds
$$
\sum_{i\in \mathcal{A}\setminus \{I_t(v)\}} p_t^v(i)\frac{1}{\left(1+\sqrt{p_t(i)}\gamma\right)^2} + p_t^v(I_t(v)) \frac{1}{\left(1+\sqrt{p_t^v(I_t(v))}(\eta\hat{\ell}_t^{v,obs}(I_t(v))+\gamma)\right)^2}=1.
$$

By Jensen's inequality,
$$
\frac{1}{\left(1 + \sum_{i\in \mathcal{A}\setminus \{I_t(v)\}}p_t^v(i)\sqrt{p_t^v(i)}\gamma + p_t^v(I_t(v))\sqrt{p_t^v(I_t(v))}(\eta \hat{\ell}_t^{v,obs}(I_t(v))+\gamma)\right)^2}\leq 1.
$$
This is equivalent to
$$
\sum_{i\in \mathcal{A}\setminus \{I_t(v)\}}p_t^v(i)\sqrt{p_t^v(i)}\gamma + p_t^v(I_t(v))\sqrt{p_t^v(I_t(v))}(\eta(v) \hat{\ell}_t^{v,obs}(I_t(v))+\gamma)\geq 0
$$
from which it follows
% $$
% \sqrt{p_t^v(i)}\gamma \geq - \eta(v) \frac{\sqrt{p_t^v(I_t(v))p_t^v(i)}}{\sum_{j=1}^K p_t^v(j)^{3/2}}\ell_{t}(I_t(v)), \hbox{ for all } i\in \mathcal{A}.
% $$
% \jy{***The above inequality only holds when $\hat{\ell}_t^{v,obs}(I_t(v)) = \ell_{t}(I_t(v)) / p_t^v(I_t(v))$, however in our case $\hat{\ell}_t^{v,obs}(I_t(v)) = \ell_{t-d}(I_t(v)) / q_{t-d}^v(I_t(v))$. 
% So the original proof of the right hand side of the inequalities in the lemma is incorrect from this step.
% I provide a new proof in red fonts below. The new proof is by an induction argument.***}
$$
\gamma \geq -\eta(v)\frac{p^v_t\left(I_t(v)\right)^{3/2}}{\sum_{i\in\mathcal{A}}p_t^v(i)^{3/2}}\hat{\ell}_t^{v,obs}(I_t(v))
\geq -\eta(v)\frac{p^v_t\left(I_t(v)\right)^{3/2}}{q^v_{t-d}(I_t(v))\sum_{i\in\mathcal{A}}p_t^v(i)^{3/2}}
$$
where the second inequality comes from $\ell_{t-d}(I_t(v))\mathbb{I}\{I_t(v)\in\mathcal{J}_{t-d}(v)\}\leq 1$.

% \mv{*** An induction argument would require proving the \emph{base step} and the \emph{induction step}. May want to clearly indicate (a) what is the induction over ($t$), (b) what is the base step ($t = d$ or whatever), and (c) what is the induction step and where exactly it starts in the proof  ($t$ to $t+1$ for some condition on $t$). The induction based proofs are usually written such that (a), (b) and (c) are clearly stated. ***} 
Here we show that for any $\delta>1$, there exists a learning rate $\eta(v)>0$ such that 
$$
p_{t}^v(i) \leq \delta p_{t-1}^v(i) \quad \text{for all } i\in\mathcal{A}
$$
for all $t=2,\dots,T$ by induction over $t$.

\textit{Base step:}
it is trivial to see that the above inequalities hold for $t=2, \dots, d+1$
by noticing that
$$
\hat{\ell}^{v,obs}_t(i) = 0 \quad \text{for all } 1\leq t\leq d
$$
from which, combined with (\ref{equ:pt}) and (\ref{equ:lambdat}), it follows that 
$$
p^v_1(i) = \cdots = p^v_{d}(i) = p^v_{d+1}(i) \quad \text{for all } i\in \mathcal{A}.
$$
Hence, for $\delta>1$,
$$
p_{t}^v(i) = p_{t-1}^v(i) \leq \delta p_{t-1}^v(i) \quad \text{for all } i\in\mathcal{A}
$$
holds for $t=2,\dots,d+1$.

\textit{Induction step:} suppose that $p_{s}^v(i)\leq \delta p_{s-1}^v(i)$ holds for all $i\in\mathcal{A} $ and for all $s=2,\dots,t$ where $1+d\leq t \leq T-1$,  
% \mv{*** is there a condition on $t$ here? ***} 
then we have
$$
p^v_{t}(i)\leq \delta p^v_{t-1}(i) \leq \cdots \leq \delta^d p^v_{t-d}(i).
$$
Specifically, we have $p^v_{t}(I_t(v))\leq \delta^d p^v_{t-d}(I_t(v))$.

From this, it follows
$$
\frac{p^v_t\left(I_t(v)\right)^{3/2}}{q^v_{t-d}(I_t(v))\sum_{i\in\mathcal{A}}p_t^v(i)^{3/2}} \leq \delta^{3d/2}\frac{p^v_{t-d}\left(I_t(v)\right)^{3/2}}{q^v_{t-d}(I_t(v))\sum_{i\in\mathcal{A}}p_t^v(i)^{3/2}}
\leq \delta^{3d/2}\frac{\sqrt{p^v_{t-d}\left(I_t(v)\right)}}{\sum_{i\in\mathcal{A}}p_t^v(i)^{3/2}}
$$
where the second inequality comes from $p^v_{t-d}(I_t(v))\leq q^v_{t-d}(I_t(v))$.

This implies
\begin{equation*}
    \begin{split}
\sqrt{p_t^v(i)}\gamma 
&\geq -\eta(v) \delta^{3d/2}\frac{\sqrt{p^v_{t-d}\left(I_t(v)\right)} \sqrt{p_t^v(i)}}{\sum_{i\in\mathcal{A}}p_t^v(i)^{3/2}}  \\
&\geq -\eta(v) \delta^{3d/2}\frac{1}{\sum_{i\in\mathcal{A}}p_t^v(i)^{3/2}}  \\
& \geq -\eta(v) \delta^{3d/2}\sqrt{K} 
    \end{split}
\end{equation*}
where the last inequality is because $\sum_{j=1}^K p_t^v(j)^{3/2} = K \sum_{j=1}^K (1/K)p_t^v(j)^{3/2}\geq K (\sum_{j=1}^K (1/K) p_t^v(j))^{3/2} = 1/\sqrt{K}$.
From (\ref{equ:pt}), it follows that for every $i\neq I_t(v)$,
$$
p_{t+1}^v(i) = \frac{1}{\left(1+\sqrt{p_t^v(i)}\gamma\right)^2}p_t^v(i)
\leq \frac{1}{\left(1-\eta(v) \delta^{3d/2}\sqrt{K} \right)^2} p_t^v(i) \leq \delta p^v_t(i)
$$
whenever $\eta(v)\leq (1-1/\sqrt{\delta})/\left(\delta^{3d/2}\sqrt{K}\right)$.

Clearly, since $\delta>1$, we have
$$
p^v_{t+1}(I_t(v))\leq p^v_t(I_t(v)) \leq \delta p^v_t(I_t(v)).
$$
Hence,
$$
p_{s}^v(i) \leq \delta p_{s-1}^v(i) \quad \text{for all } i\in\mathcal{A}
$$
holds for $s=t+1$.

Now we have shown both the base step and induction step.

From the induction, for any $\delta>1$,
$$
p_{t+1}^v(i) \leq \delta p_{t}^v(i)
$$
whenever $\eta(v)\leq (1-1/\sqrt{\delta})/\left(\delta^{3d/2}\sqrt{K}\right)$.

From (\ref{equ:pt}) and $\gamma\leq 0$, for all $i\in \mathcal{A}$,
$$
p_t^v(i) \leq \left(1+\sqrt{p_t^v(i)}(\eta(v)\hat{\ell}_t^{v,obs}(i))\right)^2 p_{t+1}^v(i).
$$
Now note that 
$$
\left(1+\sqrt{p_t^v(i)}(\eta(v)\hat{\ell}_t^{v,obs}(i))\right)^2
\leq \left(1+\eta(v)\hat{\ell}_t^{v,obs}(i)\right)^2 
\leq \frac{1+\eta(v)\hat{\ell}_t^{v,obs}(i)}{1-\eta(v)\hat{\ell}_t^{v,obs}(i)}
$$
from which it follows that
$$
p_t^v(i) \leq \frac{1+\eta(v)\hat{\ell}_t^{v,obs}(i)}{1-\eta(v)\hat{\ell}_t^{v,obs}(i)} p_{t+1}^v(i).
$$
This is equivalent to
$$
\left(1-\eta(v)\hat{\ell}_t^{v,obs}(i)\right)p_t^v(i) \leq \left(1+\eta(v)\hat{\ell}_t^{v,obs}(i)\right) p_{t+1}^v(i) = p_{t+1}^v(i) + \eta(v)\hat{\ell}_t^{v,obs}(i) p_{t+1}^v(i).
$$
It follows that when $\eta(v)\leq (1-1/\sqrt{\delta})/(\delta^{3d/2}\sqrt{K})$,
$$
\left(1-\eta(v)\hat{\ell}_t^{v,obs}(i)\right)p_t^v(i) \leq p_{t+1}^v(i) + \eta(v)\hat{\ell}_t^{v,obs}(i) p_{t+1}^v(i) \leq p_{t+1}^v(i) + \delta\eta(v)\hat{\ell}_t^{v,obs}(i) p_{t}^v(i).
$$
Rearranging the terms, we have
$$
\left(1-(1+\delta)\eta(v)\hat{\ell}_t^{v,obs}(i)\right)p_t^v(i) \leq p_{t+1}^v(i).
$$
Hence, we have 
$$
\left(1-(1+\delta)\eta(v)\hat{\ell}_t^{v,obs}(i)\right)p_t^v(i) \leq p_{t+1}^v(i) \leq \delta p_{t}^v(i)
$$
whenever $\eta(v)\leq (1-1/\sqrt{\delta})/(\delta^{3d/2}\sqrt{K})$.
Letting $\delta=2$ completes the proof.

\subsubsection{Proof of Theorem~\ref{thm::CFTRL}}
\label{sec::chapter-aamas-center-based}
Recall from the proof of Lemma~3.3 that for any $\delta>1$, when $\eta(v)\leq (1-1/\sqrt{\delta})/(\delta^{3d/2}\sqrt{K})$, then it holds
$$
\left(1-(1+\delta)\eta(v)\hat{\ell}_t^{v,obs}(i)\right)p_t^v(i) \leq p_{t+1}^v(i) \leq \delta p_{t}^v(i).
$$
Let $\delta=1+1/d$. Then, we have that for each center agent $c\in \mathcal{C}$,
\begin{equation}
\label{eq::p-center-non-center}
p_{t}^c(i) \leq (1+1/d) p_{t-1}^c(i) \leq  \cdots \leq (1+1/d)^d p_{t-d}^c(i) \leq 3p_{t-d}^c(i)
\end{equation}
and
\begin{equation}
\label{eq:regret-center-non-center}
p_t^c(i)
\leq p_{t+1}^c(i) + (1+\delta)\eta(c)p_t^c(i)\hat{\ell}_t^{c,obs}(i) 
= p_{t+1}^c(i) + (2+1/d)\eta(c)p_t^c(i)\hat{\ell}_t^{c,obs}(i)
\end{equation}
when 
\begin{equation*}
    \eta(c)  \leq \frac{1-\sqrt{d/(d+1)}}{(1+1/d)^{3d/2}\sqrt{K}}.
\end{equation*}

From Algorithm~\ref{algo:CFTRL} and (\ref{eq::p-center-non-center}), it follows that
\begin{eqnarray*}
    q_t^c(i)
    &=& 1-\prod_{v\in \mathcal{N}(c)}(1-p_t^v(i)) \nonumber\\
    &=& 1-\prod_{v\in \mathcal{N}(c)}(1-p_{t-d}^c(i)) \nonumber\\
    &\geq & 1-\left(1-\frac{1}{3}p_t^c(i)\right)^{|\mathcal{N}(c)|} \nonumber\\
    &\geq & 1-\exp\left(-\frac{1}{3}p_t^c(i)|\mathcal{N}(c)|\right) \nonumber\\
    &\geq & (1-e^{-1})\min\left\{1, \frac{1}{3}p^c_t(i)|\mathcal{N}(c)|\right\} \nonumber \\
\end{eqnarray*}
from which it follows
\begin{equation}\label{eq:qc}
    \frac{1}{q^c_t(i)} \leq \frac{1}{(1-e^{-1})\min\{1, \frac{1}{3}p^c_t(i)|\mathcal{N}(c)|\}}
\leq 2 + \frac{6}{|\mathcal{N}(c)|p^c_t(i)}.
\end{equation}

For the Tsallis entropy regularizer (1) in Algorithm~1, we have
\begin{equation}
f_t^{\prime\prime}(x) = \frac{1}{2\eta(c) x^{3/2}}.
\label{equ:ft}
\end{equation}
From Lemma~\ref{lm:ftrl-coop:individual} and (\ref{equ:ft}), we have that the individual regret of center $c$ is bounded as
\begin{eqnarray}
    R_{T}^c
    &\leq & M + \frac{1}{2}  \mathbb{E} \left[\sum_{t=1}^T  \sum_{i\in \mathcal{A}} \frac{1}{q_t^c(i) f_t^{\prime\prime}(p_t^c(i))}\right] +  d\mathbb{E} \left[\sum_{t=1}^T \sum_{i\in \mathcal{A}} \frac{1}{f_t^{\prime\prime}(p_t^c(i))} \right]\nonumber\\
    &\leq & 2\frac{1}{\eta(c)} \sqrt{K} + \eta(c)  \mathbb{E} \left[\sum_{t=1}^T\sum_{i\in \mathcal{A}} \frac{p_t^c(i)^{\frac{3}{2}}}{q_t^c(i)} \right]+ 2d\eta(c)  \mathbb{E} \left[\sum_{t=1}^T \sum_{i\in \mathcal{A}} p_t^c(i)^{\frac{3}{2}}\right].\label{equ:rtc}
\end{eqnarray}
Since $p^c_t(i) \leq 1$, we have
\begin{equation}
    \label{eq:p-delay-term}
    \sum_{t=1}^T \sum_{i\in \mathcal{A}} p_t^c(i)^{\frac{3}{2}} \leq \sum_{t=1}^T \sum_{i\in \mathcal{A}} p_t^c(i) = T.
\end{equation}

Let us define $M:\mathcal{V}\rightarrow \mathbf{R}$ as follows
\begin{equation}
    \label{eq::mass}
        M(c) = \min \{|\mathcal{N}(c)|, K\}, \hbox{ for } c\in \mathcal{C} \hbox{ and }
        M(v) = e^{-\frac{1}{6} d(v)}M(C(v)) \hbox{ for } v\in \mathcal{V}\setminus \mathcal{C}.
\end{equation}

Now, note
\begin{eqnarray}
\mathbb{E} \left[\sum_{t=1}^T\sum_{i\in \mathcal{A}} \frac{p_t^c(i)^{\frac{3}{2}}}{q_t^c(i)} \right]
& \leq & \mathbb{E} \left[\sum_{t=1}^T\sum_{i\in \mathcal{A}} p_t^c(i)^{\frac{3}{2}}\left(2 + \frac{6}{|\mathcal{N}(c)|p^c_t(i)}\right) \right] \nonumber\\
& \leq & 
2T + \frac{6}{M(c)} \mathbb{E} \left[\sum_{t=1}^T\sum_{i\in \mathcal{A}} \sqrt{p_t^c(i)}\right] \nonumber\\
& \leq &
2T + \frac{6T\sqrt{K}}{M(c)}
% & \leq & \frac{2T\sqrt{K}}{(1-1/e)M(c)}.
\label{equ:pqsum}
\end{eqnarray}
where the first inequality is by (\ref{eq:qc}), the second inequality is by (\ref{eq:p-delay-term}) and (\ref{eq::mass}), and the last inequality is by Cauchy-Schwartz inequality.

From (\ref{equ:rtc}), (\ref{eq:p-delay-term}), and (\ref{equ:pqsum}), we have
$$
R_{T}^c
\leq 
2\frac{1}{\eta(c)} \sqrt{K} + \eta(c)\frac{6T\sqrt{K}}{M(c)} + 2(d+1)\eta(c)T.
$$
By letting $\eta(c) = \sqrt{M(c)/(3T)}$, it follows that
\begin{equation}
    \label{eq::center-individual}
    \begin{aligned}
    R_{T}^c
    &\leq 4\sqrt{3}\sqrt{\frac{K T}{M(c)} } + \frac{2(d+1)}{\sqrt{3}}\sqrt{M(c)T}.
    \end{aligned}
\end{equation}
Note that when
$$T\geq 36(d+1)^2K\max_{c}M(c) > 9K\max_{c}M(c) \left(\frac{1+\sqrt{1-\frac{1}{d+1}}}{\frac{1}{d+1}}\right)^2 = \frac{9K\max_{c}M(c)}{\left(1-\sqrt{\frac{d}{d+1}}\right)^2},$$
it follows that 
$$
\eta(c)= \sqrt{\frac{M(c)}{3T}}\leq \frac{1-\sqrt{\frac{d}{d+1}}}{3^{3/2}\sqrt{K}} \leq \frac{1-\sqrt{\frac{d}{d+1}}}{(1+1/d)^{3d/2}\sqrt{K}}.
$$
Hence (\ref{eq::p-center-non-center}) and (\ref{eq:regret-center-non-center}) holds for every center $c$.

Now, note that for any non-center agent $v$, $p_t^v = p^{C(v)}_{t-d(v)d}$.
Then, from (\ref{eq:regret-center-non-center}), for all $t>d(v)d$,
\begin{eqnarray}
    p_t^v(i) 
    &=& p^{C(v)}_{t-d(v)d}(i) \nonumber \\
    &\leq& p^{C(v)}_{t-d(v)d+1}(i) + (2+1/d)\eta(C(v)) p^{C(v)}_{t-d(v)d}(i) \hat{\ell}^{C(v),obs}_{t-d(v)d}(i) \nonumber \\
    &\cdots& \nonumber\\
    &\leq& p^{C(v)}_{t}(i) + (2+1/d)\eta(C(v)) \sum_{s=1}^{d(v)d} p^{C(v)}_{t-s}(i) \hat{\ell}^{C(v),obs}_{t-s}(i). \label{eq::non-center-prob}
\end{eqnarray}

For every non-center agent $v$, the individual regret is bounded as 
\begin{eqnarray}
    R_T^v 
    &\leq & (d(v)+1)d + \mathbb{E}\left[ \sum_{t=(d(v)+1)d+1}^{T} \sum_{i\in \mathcal{A}} p_{t}^{v}(i) \ell_{t}(i)-\min _{i \in \mathcal{A}} \sum_{t=(d(v)+1)d+1}^{T} \ell_{t}(i)\right] \nonumber\\
    &\leq &  (d(v)+1)d + \mathbb{E}\left[\sum_{t=(d(v)+1)d+1}^{T}\sum_{i\in \mathcal{A}} p^{C(v)}_{t}(i) \ell_{t}(i)-\min _{i \in \mathcal{A}} \sum_{t=(d(v)+1)d+1}^{T} \ell_{t}(i)\right] \nonumber \\
    & & + (2+1/d)\eta(C(v)) \mathbb{E}\left[\sum_{t=(d(v)+1)d+1}^{T} \sum_{i\in \mathcal{A}} \sum_{s=1}^{d(v)d} p_{t-s}^{C(v)}(i) \hat{\ell}_{t-s}^{C(v), obs}(i)\ell_{t}(i)\right]  \nonumber \\
    &=& (d(v)+1)d+ \mathbb{E}\left[\sum_{t=(d(v)+1)d+1}^{T} \sum_{i\in \mathcal{A}} p^{C(v)}_{t}(i) \ell_{t}(i)-\min _{i \in \mathcal{A}} \sum_{t=(d(v)+1)d+1}^{T} \ell_{t}(i)\right] \nonumber \\
    & & + (2+1/d)\eta(C(v)) \mathbb{E}\left[\sum_{t=(d(v)+1)d+1}^{T} \sum_{i\in \mathcal{A}} \sum_{s=1}^{d(v)d} p_{t-s}^{C(v)}(i) \hat{\ell}_{t-s-d}^{C(v)}(i) \ell_{t}(i)\right]  \label{eq::individual-regret} 
\end{eqnarray}
where the second inequality follows from (\ref{eq::non-center-prob}) and the equality uses $\hat{\ell}_{t-d}^{v} = \hat{\ell}_{t}^{v, obs}$ when $t>d$.

From Lemma~\ref{lm:conditional_exp}, we have
\begin{eqnarray*}
&& \mathbb{E}\left[\sum_{t=(d(v)+1)d+1}^{T} \sum_{i\in\mathcal{A}} \sum_{s=1}^{d(v)d} p_{t-s}^{C(v)}(i) \hat{\ell}_{t-s-d}^{C(v)}(i) \ell_{t}(i)\right]\\
& \leq &
\mathbb{E}\left[\sum_{t=(d(v)+1)d+1}^{T}\sum_{i\in \mathcal{A}} \sum_{s=1}^{d(v)d} p_{t-s}^{C(v)}(i)\mathbb{E}_{t-s-d}\left[ \hat{\ell}_{t-s-d}^{C(v)}(i) \right]\right]
\\
& \leq & d(v)dT.
\end{eqnarray*}

Hence, it follows that
$$
R_T^v \leq R_{T}^{C(v)}+(d(v)+1)d+(2+1/d) \eta(C(v)) d(v)dT.
$$

Combining with Eq.~(\ref{eq::center-individual}), we have
$$
    \label{eq::non-center-individual}
    \begin{aligned}
    R_T^v
    &\leq 4\sqrt{3}\sqrt{\frac{K T}{M(C(v))} } + \frac{2(d+1)}{\sqrt{3}}\sqrt{M(C(v))T} + (d(v)+1)d+(2+1/d)\eta(C(v)) d(v)dT  \\
    &= 4\sqrt{3}\sqrt{\frac{K T}{M(C(v))} } + \frac{2(d+1)}{\sqrt{3}}\sqrt{M(C(v))T} + (d(v)+1)d+(2d+1)d(v)\sqrt{M(C(v))T/3} \\
    &\leq 4\sqrt{3}\sqrt{\frac{K T}{M(v)} } + \frac{2(d+1)}{\sqrt{3}}\sqrt{M(C(v))T} + (d(v)+1)d+(2d+1)d(v)\sqrt{M(C(v))T/3} \\
    &\leq 4\sqrt{3}\sqrt{\frac{K T}{M(v)} } + \frac{2(d+1)}{\sqrt{3}}\sqrt{M(C(v))T} + 6d\log(K)+2\sqrt{3}(2d+1)\log(K)\sqrt{M(C(v))T} \\
    \end{aligned}
$$
where the penultimate inequality follows from Eq.~(\ref{eq::mass}) and the last inequality follows from Lemma~7 in \cite{bar2019individual}.

From Theorem~8 in \cite{bar2019individual}, we have for all agent $v\in \mathcal{V}$
$$
M(v) \geq (1/e) \min \{|\mathcal{N}(v)|, K\}
$$
Hence, we have
\begin{multline*}
    R_T^v
    \leq 
    4\sqrt{3}\sqrt{e}\sqrt{\frac{KT}{ \min \{|\mathcal{N}(v)|, K\}}} \\
    +  \frac{2(d+1)}{\sqrt{3}}\sqrt{M(C(v))T}+ 6d\log(K) + 2\sqrt{3}(2d+1)\log(K)\sqrt{M(C(v))T}\\
    \leq 
    12\sqrt{\frac{KT}{|\mathcal{N}(v)|}}+  \frac{2(d+1)}{\sqrt{3}}\sqrt{|\mathcal{N}(C(v))|T}+ 6d\log(K) + 2\sqrt{3}(2d+1)\log(K)\sqrt{|\mathcal{N}(C(v))|T}\\
\end{multline*}
where the second inequality comes from the fact that $K\geq\max_v{|\mathcal{N}(v)|}$.

\subsubsection{Proof of Lemma~\ref{lm:ftrl-coop-independence}}
\label{sec::chapter-aamas-upper-bound-lemma-2}

The proof follows by the following sequence of relations:

\begin{equation*}
    \begin{aligned}
    \sum_{i\in \mathcal{A}}\sum_{v \in \mathcal{V}} \frac{p^v(i)^{3/2}}{q^v(i)} &=  \sum_{i\in \mathcal{A}}\sum_{v \in \mathcal{V}} \left(\frac{p^v(i)}{q^v(i)} \cdot \sqrt{p^v(i)}\right) \\ 
    &\leq \sum_{i\in \mathcal{A}}\sqrt{\sum_{v \in \mathcal{V}}\frac{p^v(i)^2}{q^v(i)^2}}\sqrt{\sum_{v \in \mathcal{V}} p^v(i)} \\
    &\leq \sum_{i\in \mathcal{A}}\sqrt{\sum_{v \in \mathcal{V}}\frac{p^v(i)}{q^v(i)}}\sqrt{\sum_{v \in \mathcal{V}} p^v(i)} \\
    &\leq \sum_{i\in \mathcal{A}}\sqrt{\frac{1}{1-1/e}\left(\alpha(G)+\sum_{v \in \mathcal{V}} p^v(i)\right)}\sqrt{\sum_{v \in \mathcal{V}} p^v(i)} \\
    &\leq \sqrt{\sum_{i\in \mathcal{A}} \frac{1}{1-1/e}\left(\alpha(G)+\sum_{v \in \mathcal{V}} p^v(i)\right)}
    \sqrt{\sum_{i\in \mathcal{A}} \sum_{v \in \mathcal{V}} p^v(i)} \\
    &= \sqrt{\frac{N}{1-1/e}\left(K\alpha(G)+N\right)} \\
    \end{aligned}
\end{equation*}
where the first inequality comes from Cauchy-Schwarz inequality, the second inequality comes from the fact that $q^v(i) \geq p^v(i)$, the third inequality comes from Lemma~3 in \cite{cesa2020cooperative} and the last inequality comes from Cauchy-Schwarz inequality again.

\subsubsection{Proof of Theorem~\ref{thm::DFTRL}}
\label{sec::chapter-aamas-upper-bound-theorem-1}

Note that 

$$
\max_{x \in \mathcal{P}_{K-1}}\{-F_{1}(x)\}+\sum_{t=2}^{T} \max_{x \in \mathcal{P}_{K-1}}\left\{F_{t-1}(x)-F_{t}(x)\right) = -F_{T}(\mathrm{e}_{[K]}/K) \\
    = 2\frac{1}{\eta_T} \sqrt{K} + \frac{1}{\zeta_T}\log(K),
$$
and
$$
f_t^{\prime\prime}(x) \geq \max\left\{\frac{1}{2\eta_t} \frac{1}{x^{3/2}}, \frac{1}{\zeta_t}\frac{1}{x}\right\}.
$$
Hence, from Lemma~3.5, we have
\begin{equation*}
    \begin{aligned}
    \frac{1}{2N}\mathbb{E}\left[\sum_{t=1}^T \sum_{i\in \mathcal{A}} \sum_{v\in \mathcal{V}} \frac{1}{q_t^v(i) f^{\prime\prime}_t(p_t^v(i))} \right]
    &\leq \frac{1}{N} \mathbb{E}\left[  \sum_{t=1}^T \eta_t\sum_{i\in \mathcal{A}} \sum_{v\in \mathcal{V}} \frac{p_t^v(i)^{3/2}}{q_t^v(i) } \right] 
    \\
    &\leq \sqrt{\frac{1}{1-1/e}\left(\frac{K}{N}\alpha(G)+1\right)} \sum_{t=1}^T\eta_t
    \end{aligned}
\end{equation*}
and
\begin{equation*}
    \begin{aligned}
    \frac{d}{N} \mathbb{E}\left[ \sum_{t=1}^T \sum_{i\in \mathcal{A}} \sum_{v\in \mathcal{V}} \frac{1}{f^{\prime\prime}_t(p_t^v(i))}  \right] 
    &\leq \frac{d}{N}  \mathbb{E}\left[ \sum_{t=1}^T \zeta_t \sum_{i\in \mathcal{A}} \sum_{v\in \mathcal{V}} p_t^v(i) \right] \\
    &= d\sum_{t=1}^T \zeta_t.
    \end{aligned}
\end{equation*}

Combining with Lemma~3.2, we have
\begin{equation*}
    \begin{aligned}
        R_{T} 
        &= \frac{1}{N} \sum_{v\in \mathcal{V}} R_T^v \\
        &\leq M + \frac{1}{2N} \mathbb{E}\left[ \sum_{t=1}^T  \sum_{i\in \mathcal{A}} \sum_{v\in \mathcal{V}} \frac{1}{q_t^v(i) f_t^{\prime\prime}(p_t^v(i))} \right]+ \frac{d}{N} \mathbb{E}\left[ \sum_{t=1}^T \sum_{i\in \mathcal{A}} \sum_{v\in \mathcal{V}} \frac{1}{f_t^{\prime\prime}(p_t^v(i))}  \right] \\
        &\leq  2\frac{1}{\eta_T}\sqrt{K} + \sqrt{\frac{1}{1-1/e}\left(\frac{K}{N}\alpha(G)+1\right)} \sum_{t=1}^T\eta_t + \frac{1}{\zeta_T}\log(K) + d\sum_{t=1}^T \zeta_t.
    \end{aligned}
\end{equation*}
Plugging 
$\eta_t = (1/(1-1/e)) (\alpha(G)/ N + 1/K)^{-1/4} \sqrt{2/T}$ and $\zeta_t = \sqrt{\log(K) / (dt)}$
into the inequality above, we complete the proof.

\subsubsection{Proof of Theorem~\ref{thm:lower-2}}
\label{sec::chapter-aamas-lower-bound}

In each round $t$, every agent $v\in \mathcal{V}$ receives $b_t(v) = O(|\mathcal{N}(v)|)$ bits since its neighboring agents can choose at most $\mathcal{N}(v)$ distinct actions. By Theorem~4 in \cite{shamir2014fundamental}, there exists some distribution $\mathcal{D}$ over $[0,1]^K$ such that loss vectors $\ell_t\sim \mathcal{D}$ for all $t=1,2,\dots, T$ independently and $\min_{i\in \mathcal{A}} \mathbb{E}\left[\ell_t(I_t(v)) - \ell_t(i)\right] = \Omega\left(\min\{T, \sqrt{KT/|\mathcal{N}(v)|}\}\right)$
Hence, the worst-case individual regret of agent $v$ is
\begin{equation*}
    \begin{aligned}
    \sup_{\ell_1\ldots,\ell_T} R_T^v 
    &=  \sup_{\ell_1\ldots,\ell_T} \mathbb{E}\left[\sum_{t=1}^{T} \ell_{t}\left(I_{t}(v)\right)-\min _{i \in \mathcal{A}} \sum_{t=1}^{T} \ell_{t}(i)\right] \\
    &\geq \mathbb{E}_{\ell_t\sim \mathcal{D}}\left[ \mathbb{E}\left[\sum_{t=1}^{T} \ell_{t}\left(I_{t}(v)\right) \right] \right] - \mathbb{E}_{\ell_t\sim \mathcal{D}}\left[\min _{i \in \mathcal{A}} \sum_{t=1}^{T} \ell_{t}(i)\right] \\
    &\geq \mathbb{E}_{\ell_t\sim \mathcal{D}}\left[ \mathbb{E}\left[\sum_{t=1}^{T} \ell_{t}\left(I_{t}(v)\right) \right] \right] - \min _{i \in \mathcal{A}}  \mathbb{E}_{\ell_t\sim \mathcal{D}}\left[\sum_{t=1}^{T} \ell_{t}(i)\right] \\
    &= \max_{i\in\mathcal{A}} \mathbb{E}\left[\sum_{t=1}^{T} \ell_{t}\left(I_{t}(v)\right) -  \sum_{t=1}^{T} \ell_{t}(i) \right]\\
    &\geq \min_{i\in\mathcal{A}} \mathbb{E}\left[\sum_{t=1}^{T} \ell_{t}\left(I_{t}(v)\right) -  \sum_{t=1}^{T} \ell_{t}(i) \right]\\
    &\geq \Omega\left(\min\{T, \sqrt{KT/|\mathcal{N}(v)|}\}\right).
    \end{aligned}
\end{equation*}
Note that with delay $d$, the individual regret of the agent $v$ can not be smaller than the regret of the agent $v$ with access to full information of $\ell_1,\dots, \ell_{t-d}$ at each round $t$.
Following the argument in the proof of Corollary~15 in \cite{cesa2019delay}, the worst-case regret of a single-agent with delayed full information is $\Omega(\sqrt{dT\log(K)})$.
Hence,
\begin{equation*}
    \sup_{\ell_1,\ldots,\ell_T} R_T^v = \Omega\left(\max\left\{ \min \left\{T, \sqrt{\frac{KT}{|\mathcal{N}(v)|}}\right\}, \sqrt{dT \log(K)}\right\} \right).
\end{equation*}
We complete the proof by noting that $R_T = \frac{1}{N}\sum_{v\in \mathcal{V}} R_T^v$.

\subsubsection{CFTRL and the center-based Exp3 when $d=1$}
\label{sec::chapter-aamas-regret-comparison}
As noted in Theorem~3.1 with $d=1$ and \cite{bar2019individual}, when $K\geq \max_{v\in V}|\mathcal{N}(v)|$ and $T\geq \max\{144K \max_{c\in \mathcal{C}}|\mathcal{N}(c)|, K^2\log(K)\}$, the individual regret of CFTRL is 
$$
O\left(\sqrt{\left(\frac{1}{|\mathcal{N}(v)|}+\frac{|\mathcal{N}(C(v))|}{K}\log(K)^2\right)KT}\right),
$$ and the individual regret of the center-based Exp3 is
$$
\tilde{O}\left(\sqrt{\left(1+\frac{K}{\mathcal{N}(v)}\right)T}\right) = O\left(\sqrt{\left(\frac{1}{|\mathcal{N}(v)|}+\frac{1}{K}\right)\log(K)KT}\right)
$$
If we want the individual regret of CFTRL lower than the center-based Exp3, it suffices to show
$$
\frac{1}{|\mathcal{N}(v)|} + \frac{|\mathcal{N}(C(v))|}{K}\log(K)^2 \leq \left(\frac{1}{|\mathcal{N}(v)|}+\frac{1}{K}\right)\log(K) 
$$
which is equivalent to
$$
\frac{|\mathcal{N}(C(v))|}{K}\log(K)^2 - \frac{\log(K)}{K} -  \leq \frac{\log(K)-1}{|\mathcal{N}(v)|}  
$$
Note that for $K\geq 9$, we have
$$
\log(K) - 1\geq \frac{1}{2}\log(K)
$$
It is suffices to show that
$$
\frac{|\mathcal{N}(C(v))|}{K}\log(K)^2 \leq \frac{\log(K)}{2|\mathcal{N}(v)|}  
$$
which is equivalent to
$$
\frac{K}{\log(K)} \geq 2 |\mathcal{N}(v)||\mathcal{N}(C(v))|.
$$

\subsubsection{CFTRL and the lower bound when $d=1$}
\label{sec::chapter-aamas-optimal-bound-condition}
As noted in Theorem~4.1, for $T\geq K/|\mathcal{N}(v)|$, the individual regret lower bound is 
$$
\Omega\left(\sqrt{\left(\frac{1}{|\mathcal{N}(v)|}+\frac{1}{K}\log(K)\right)KT}\right).
$$
From Theorem~3.1 with $d=1$, when  $K\geq \max_{v\in V}|\mathcal{N}(v)|$ and $T\geq 144K \max_{c\in \mathcal{C}}|\mathcal{N}(c)|$, the individual regret of CFTRL is 
$$
O\left(\sqrt{\left(\frac{1}{|\mathcal{N}(v)|}+\frac{|\mathcal{N}(C(v))|}{K}\log(K)^2\right)KT}\right).
$$
Hence when $K\geq \max_{v\in V}|\mathcal{N}(v)|$ and $T\geq \max\{144K \max_{c\in \mathcal{C}}|\mathcal{N}(c)|, K/|\mathcal{N}(v)|\}$,
the ratio between the individual regret of CFTRL and the individual regret lower bound is 
$$
O\left( \sqrt{\frac{\frac{1}{|\mathcal{N}(v)|}+\frac{|\mathcal{N}(C(v))|}{K}\log(K)^2}{\frac{1}{|\mathcal{N}(v)|}+\frac{1}{K}\log(K)}} \right).
$$
Note that 
\begin{equation*}
    \begin{split}
\sqrt{\frac{\frac{1}{|\mathcal{N}(v)|}+\frac{|\mathcal{N}(C(v))|}{K}\log(K)^2}{\frac{1}{|\mathcal{N}(v)|}+\frac{1}{K}\log(K)}}
 &\leq 1 + \sqrt{\frac{|\mathcal{N}(C(v))|\log(K)-1}{\frac{K}{|\mathcal{N}(v)|\log(K)}+1}} \\
 &\leq 1 + \sqrt{\frac{|\mathcal{N}(C(v))||\mathcal{N}(v)|\log(K)^2}{K}}.        
    \end{split}
\end{equation*}
When $|\mathcal{N}(C(v))||\mathcal{N}(v)|\log(K)^2/K = O(1)$, the individual regret upper bound of CFTRL matches the lower bound up to a constant factor.

%%%%%%%%%%%%%%%%%%%%%%%%%%%%%%%%%%%%%

\newpage\onehalfspacing

\setcounter{section}{4}
\setcounter{subsection}{0}

\section*{\textbf{\Large{Chapter~4}}}
\addcontentsline{toc}{section}{Chapter~4. Doubly Adversarial Federated Bandits}

\bigskip
\bigskip

\textbf{\Large{Doubly Adversarial Federated Bandits}}

\bigskip
\bigskip
\bigskip

\noindent{There is a rising trend of research on federated learning, which coordinates multiple \emph{heterogeneous} agents to collectively train a learning algorithm, while keeping the raw data decentralized \citep{kairouz2021advances}.
We consider the federated learning variant of a multi-armed bandit problem which is one of the most fundamental sequential decision making problems.
In standard multi-armed bandit problems, a learning agent needs to balance the trade-off between exploring various arms in order to learn how much rewarding they are and selecting high-rewarding arms.
In federated bandit problems, multiple heterogeneous agents collaborate with each other to maximize their cumulative rewards.
The challenge here is to design decentralized collaborative algorithms to find a \emph{globally} best arm for all 
agents while keeping their raw data decentralized.}

\subsection{Introduction}

Finding a globally best arm with raw arm or loss sequences stored in a distributed system has ubiquitous applications in many systems built with a network of learning agents.
One application is in recommender systems where different recommendation app clients (i.e. agents) in a communication network collaborate with each other to find news articles (i.e. arms) that are popular among all users within a specific region, which can be helpful to solve the \emph{cold start} problem \citep{li2010contextual, yi2021efficient}.
In this setting, the system avoids the exchange of users' browsing history (i.e. arm or loss sequences) between different clients for better privacy protection.
Another motivation is in international collaborative drug discovery research, where different countries (i.e. agents) cooperate with each other to find a drug (i.e. arm) that is uniformly effective for all patients across the world \citep{varatharajah2022contextual}. To protect the privacy of the patients involved in the research, the exact treatment history of specific patients (i.e. arm or loss sequences) should not be shared during the collaboration.

The federated bandit problems are focused on identifying a globally best arm (pure exploration) or maximizing the cumulative  group reward (regret minimization) in face of heterogeneous feedback from different agents for the same arm, which has gained much attention in recent years \citep{dubey2020differentially, zhu2021federated, huang2021federated, shi2021federated, reda2022near}. In prior work, heterogeneous feedbacks received by different agents are modeled as samples from some unknown but fixed distributions.
Though this formulation of heterogeneous feedback allows elegant statistical analysis of the regret, it may not be adequate for dynamic (non-stationary) environments. For example, consider the task of finding popular news articles within a region mentioned above. The popularity of news articles on different topics can be time-varying, e.g. the news on football may become most popular during the FIFA World Cup but may be less popular afterwards.

In contrast with the prior work, we introduce a new \emph{non-stochastic} federated multi-armed bandit problem in which the heterogeneous feedback received by different agents are chosen by an oblivious adversary.
We consider a federated bandit problem with $K$ arms and $N$ agents. The agents can share their information via a communication network. At each time step, each agent will choose one arm, receive the feedback and exchange their information with their neighbors in the network. The problem is \emph{doubly adversarial}, i.e. the losses are determined by an oblivious adversary which specifies the loss of each arm not only for each time step but also for each agent. As a result, the agents which choose the same arm at the same time step may observe different losses. 
The goal is to find the \emph{globally} best arm in hindsight, whose cumulative loss averaged over all agents is lowest, without exchanging raw information consisting of arm identity or loss value sequences among agents.
% We will consider two variants of this problem: either the full-information feedback setting, i.e., agents has access to the losses of all arms, or the bandit feedback setting, i.e., agents only have access to the loss of arm they choose.
As standard in online learning problems, we focus on regret minimization over an arbitrary time horizon.

%\mv{Perhaps providing some more justification for why studying both full information and bandit feedback.}

\subsubsection{Related work}

The doubly adversarial federated bandit problem is related to several lines of research, namely that on federated bandits, multi-agent cooperative adversarial bandits, and distributed optimization. Here we briefly discuss these related works.

\paragraph{Federated bandits}

Solving bandit problems in the federated learning setting has gained attention in recent years. \cite{dubey2020differentially} and \cite{huang2021federated}  considered the linear contextual bandit problem and extended the LinUCB algorithm \citep{li2010contextual} to the federated learning setting.
\cite{zhu2021federated} and \cite{shi2021federated}
studied a federated multi-armed bandit problem where the losses observed by different agents are i.i.d. samples from some common unknown distribution.
\cite{reda2022near} considered the problem of identifying a globally best arm for multi-armed bandit problems in a centralized federated learning setting. All these works focus on the stochastic setting, i.e. the reward or loss of an arm is sampled from some unknown but fixed distribution. Our work considers the non-stochastic setting, i.e. losses are chosen by an oblivious adversary, which is a more appropriate assumption for non-stationary environments. 

\paragraph{Multi-agent cooperative adversarial bandit}
\cite{cesa2016delay, bar2019individual, yi2022regret} studied the adversarial case where agents receive the same loss for the same action chosen at the same time step, whose algorithms require the agents to exchange their raw data with neighbors.
\cite{cesa2020cooperative} discussed the cooperative online learning setting where the agents have access to the full-information feedback and the communication is asynchronous. In these works, the agents that choose the same action at the same time step receive the same reward or loss value and agents aggregate messages received from their neighbors. Our work relaxes this assumption by allowing agents to receive different losses even for the same action in a time step. Besides, we propose a new algorithm that uses a different aggregation of messages than in the aforementioned papers, which is based on distributed dual averaging method in \cite{nesterov2009primal, xiao2009dual, duchi2011dual}.

% \paragraph{Distributed online learning} 
%      $O(T^{3/4})$ regret upper bound appears in distributed online convex optimization \citep{wan2022projection} and distributed online linear regression \citep{DBLP:journals/tit/YuanPS21}. \mv{Here we mention $O(T^{3/4})$ while our new upper bounds are $O(T^{2/3})$?}

\paragraph{Distributed optimization} \cite{duchi2011dual} proposed the dual averaging algorithm for distributed convex optimization via a gossip communication mechanism. Subsequently,  \cite{hosseini2013online} extended this algorithm to the online optimization setting.
\cite{scaman2019optimal} found optimal distributed algorithms for distributed convex optimization and a lower bound which applies to strongly convex functions. The doubly adversarial federated bandit problem with full-information feedback is a special case of distributed online linear optimization problems. Our work complements these existing studies by providing a lower bound for the distributed online linear optimization problems. Moreover, our work proposes a near-optimal algorithm for the more challenging bandit feedback setting.

%\mv{May add punchlines for different groups of related works, discussing how our problem and/or results are different. }

\subsubsection{Organization of this chapter}

We first formally formulate the doubly adversarial federated bandit problem and the federated bandit algorithms we study in Section~\ref{sec::problem-setting}.
Then, in Section ~\ref{sec:lower-bound}, we provide two regret lower bounds for any federated bandit algorithm under the full-information and bandit feedback setting, respectively.
In Section~\ref{sec:upper-bound}, we present a federated bandit algorithm adapted from the celebrated Exp3 algorithm for the bandit-feedback setting, together with its regret upper bound.
Finally, we show results of our numerical experiments in Section~\ref{sec:numerical-experiments}.
All the proofs are available in Section~\ref{sec::chapter-icml-app}

\subsection{Problem setting} \label{sec::problem-setting}

Consider a communication network defined by an undirected graph $\mathcal{G}=(\mathcal{V}, \mathcal{E})$, where $\mathcal{V}$ is the set of $N$ agents and $(u,v)\in \mathcal{E}$ if agent $u$ and agent $v$ can directly exchange messages.
We assume that $\mathcal{G}$ is simple, i.e. it contains no self loops nor multiple edges.
The agents in the communication network collaboratively aim to solve a non-stochastic multi-armed bandit problem. In this problem, there is a fixed set $\mathcal{A}$ of $K$ arms and a fixed time horizon $T$.
%Without loss of generality, we assume $|\mathcal{V}|=N$ and $|\mathcal{A}|=K$.
Each instance of the problem is parameterized by a tensor $L = \left( \ell_t^v(i) \right)\in [0, 1]^{T\times N\times K}$ where $\ell_t^v(i)$ is the loss associated with agent $v\in \mathcal{V}$ if it chooses arm $i \in \mathcal{A}$ at time step $t$.

At each time step $t$, each agent $v$ will choose its action $a^v_t=i$, observe the feedback $I^v_t$ and incur a loss defined as the average of losses of arm $i$ over all agents, i.e., 
\begin{equation}
    \Bar{\ell}_t (i) = \frac{1}{N} \sum_{v\in \mathcal{V}} \ell_t^v(i).
\end{equation}
At the end of each time step, each agent $v\in \mathcal{V}$ can communicate with their neighbors $\mathcal{N}(v)=\{u\in \mathcal{V}: (u, v)\in \mathcal{E}\}$. 
We assume a \emph{non-stochastic} setting, i.e. the loss tensor $L$ is determined by an oblivious adversary. In this setting, the adversary has the knowledge of the description of the algorithm running by the agents but the losses in $L$ do not depend on the specific arms selected by the agents.

The performance of each agent $v\in \mathcal{V}$ is measured by its \emph{regret}, defined as the difference of the expected cumulative loss incurred and the cumulative loss of a \emph{globally} best fixed arm in hindsight, i.e.
\begin{equation}
R_T^v(\pi, L) = \mathbb{E} \left[ \sum_{t=1}^T \Bar{\ell}_t(a_t^v) - \min_{i\in \mathcal{A}}\left\{ \sum_{t=1}^T \Bar{\ell}_t(i)\right\} \right]
\label{equ:regret}
\end{equation}
where the expectation is taken over the action of all agents under algorithm $\pi$ on instance $L$. 
We will abbreviate $R_T^v(\pi, L)$ as $R_T^v$ when the algorithm  $\pi$ and instance $L$ have no ambiguity in the context.
We aim to characterize $\max_{L}R_T^v(\pi, L)$ for each agent $v\in \mathcal{V}$ under two feedback settings, 
\begin{itemize}
    \item full-information feedback: $I^v_t = \ell_t^v$, and
    \item bandit feedback: $I^v_t = \ell_t^v(a^v_t)$.
\end{itemize}
Let  $\mathcal{F}^v_t$ be the sequence of agent $v$'s actions and feedback up to time step $t$, i.e., $\mathcal{F}^v_t = \bigcup_{s=1}^t \{a^v_s, I^v_s \}$. For a graph $\mathcal{G}=(\mathcal{V}, \mathcal{E})$, we denote as $d(u, v)$ the number of edges of a shortest path connecting nodes $u$ and $v$ in $\mathcal{V}$ and $d(v,v)=0$.

We focus on the case when $\pi$ is a \emph{federated bandit algorithm} in which each agent $v\in\mathcal{V}$ can only communicate with their neighbors within a time step.
\begin{definition}[federated bandit algorithm]\label{def:federated}
    A federated bandit algorithm $\pi$ is a multi-agent learning algorithm such that for each round $t$ and each agent $v\in \mathcal{V}$, the action selection distribution $p^v_t$ only depends on $\bigcup_{u\in \mathcal{V}} \mathcal{F}^u_{t-d(u, v)-1}$.
\end{definition}
From Definition~\ref{def:federated}, the communication between any two agents $u$ and $v$ in $\mathcal{G}$ comes with a delay equal to $d(u, v)+1$. Here we give some examples of $\pi$ in different settings:
\begin{itemize}
    \item when $|\mathcal{V}| = 1$ and $I^v_t = \ell_t^v$, $\pi$ is an online learning algorithm for learning with expert advice problems \citep{cesa1997use},
    \item when $|\mathcal{V}| = 1$ and $I^v_t = \ell_t^v(a^v_t)$, $\pi$ is a sequential algorithm for a multi-armed bandit problem \citep{auer2002finite}, 
    \item when $I^v_t \in \partial f_i(x^v_t)$ for some convex function $f(x)$, $\pi$ belongs to the black-box procedure for distributed convex optimization over a simplex \citep{scaman2019optimal}, and
    \item when $\mathcal{G}$ is a star graph, $\pi$ is a centralized federated bandit algorithm discussed in \cite{reda2022near}.
\end{itemize}

% \mv{Circle notation appears strange to me as it suggests it is a binary operator. Maybe write instead $\mathbb{E}_{\pi,L}$, if it is necessary to indicate $\pi$ and $L$.}

% \subsection{Black-box online learning procedure}

\subsection{Lower bounds} \label{sec:lower-bound}

In this section, we show two lower bounds on the cumulative regret of any federated bandit algorithm $\pi$ in which all agents exchange their messages through the communication network $\mathcal{G} = \left(\mathcal{V}, \mathcal{E}\right)$, for full-information and bandit feedback setting. 
Both lower bounds highlight how the cumulative regret of the federated algorithm $\pi$ is related to the minimum time it takes for all agents in $\mathcal{G}$ to reach an agreement on a globally best arm.
% \mv{May reconsider the structure of sections. Currently, there is first a section for full-information feedback and then another section for full-bandit feedback. For the former, we only provide a lower bound, while for the latter we provide a lower and an upper bound. This appears strange. Maybe combining all theoretical results in one section? Also would need to explain why for the full information feedback we provide only a lower bound.}

Agents reaching an agreement about a globally best arm in hindsight is to find $i^\ast \in \arg\min_{i\in \mathcal{A}} \sum_{t=1}^T \Bar{\ell}_t(i)$ by each agent $v$ exchanging their private information about $\{\sum_{t=1}^T \ell^v_t(i): i\in \mathcal{A}\}$ with their neighbors. 
This is known as a distributed consensus averaging problem \citep{boyd2004fastest}. 
% \mv{The last sentence is ambiguous because reaching an agreement by exchanging information with neighbors can be for various kinds of agreement objectives while distributed consensus averaging is for a particular agreement objective.} 
% \mv{*** How about using notation $d_u$, $d_{\min}$ and $d_{\max}$ instead of $d_u$, $d_{\min}$ and $d_{\max}$? This would be more standard notation. ***} 
Let $d_v = |\mathcal{N}(v)|$ and $d_{\max} = \max_{v\in\mathcal{V}} d_v$ and $d_{\min} = \min_{v\in\mathcal{V}} d_v$ .
The dynamics of a consensus averaging procedure is usually characterized by spectrum of the Laplacian matrix $M$ of graph $\mathcal{G}$ defined as
$$
M_{u, v}:= \begin{cases} d_{u} & \text { if } u=v \\ -1 & \text { if } u \neq v \text { and } (u,v)\in \mathcal{E} %u \text { is adjacent to } v 
\\ 0 & \text { otherwise}.\end{cases}
$$
% a family of matrices $\mathcal{M}_\mathcal{G}$.
% Let $\mathcal{M}_\mathcal{G}$ be the set of all $N\times N$ square real matrix $M$ such that
% \begin{enumerate}
%     \item $M$ is symmetric and positive semi-definite;
%     \item the kernel of $M$ is the set of constant vectors, i.e. $\{x\in \mathbb{R}^N: Mx=0\} = \operatorname{Span}(\mathbb{1})$ where $\mathbb{1}=(1, \ldots, 1)^{\top}$;
%     \item $M_{i,j} \neq 0$ only if $i=j$ or $(i,j)\in \mathcal{E}$.
% \end{enumerate}
Let $\lambda_1(M)\geq \cdots\geq \lambda_N(M) = 0$ be the eigenvalues of the Laplacian matrix $M$. 
% \mv{*** Hereinafter, it should be $N$ instead of $n$, as we use capital letters to denote number agents, number of arms, and horizon time. ***}
% The convergence analysis of a consensus procedure usually depends on $\jy{\frac{1+d_{\max}}{\lambda_{N-1}(M)} =} \lambda_1(M) / \lambda_{N-1}(M)$ \citep{scaman2019optimal}, 
% \mv{Why is this called an \emph{eigengap} while being defined as a ratio of eigenvalues?}
% We denote $\frac{1+d_{\max}}{\lambda_{N-1}(M)} = \min_{M\in\mathcal{M}_\mathcal{G}} \gamma(M)$ 
% which measures the time it takes for all agents in the communication network $\mathcal{G}$ to reach a consensus using distributed averaging algorithms.
The second smallest eigenvalue $\lambda_{N-1}(M)$ is the \emph{algebraic connectivity} which approximates the sparest-cut of graph $\mathcal{G}$ \citep{arora2009expander}.
In the following two theorems, we show that for any federated bandit algorithm $\pi$, there always exists a problem instance and an agent whose worst-case cumulative regret is $\Omega(\lambda_{N-1}(M)^{-1/4} \sqrt{T})$.

\begin{theorem}[Full-information feedback]
    \label{thm:lower-bound-full}For any federated bandit algorithm $\pi$, there exists a graph $\mathcal{G}=(\mathcal{V}, \mathcal{E})$ with Laplacian matrix $M$ and a full-information feedback instance $L\in [0, 1]^{T\times N\times K}$ such that for some $v_1 \in \mathcal{V}$,
    \begin{equation}
        \label{lower-bound:full-feedback}
        R^{v_1}_T(\pi, L) = \Omega \left(\sqrt[4]{\frac{1+d_{\max}}{\lambda_{N-1}(M)}}\sqrt{T\log K}\right).
    \end{equation}
\end{theorem}

The proof, in Section~\ref{app-lower-full}, relies on the existence of a graph in which there exist two clusters of agents, $A$ and $B$, with distance $d(A, B) = \min_{u\in A, v\in B}d(u, v) = \Omega\left(\sqrt{(d_{\max}+1)/\lambda_{N-1}(M)}\right)$. Then, we consider an instance where only agents in cluster $A$ receive non-zero losses. 
Based on a reduction argument, the cumulative regrets for agents in cluster $B$ are the same as (up to a constant factor) the cumulative regret in a single-agent adversarial bandit problem with feedback of delay $d(A, B)$ (see Lemma~\ref{lm:delay} in Section~\ref{app-lower-full}). Hence, one can show that the cumulative regret of agents in cluster $B$ is $\Omega \left(\sqrt{d(A,B)}\sqrt{T\log K}\right)$. 

Note that the doubly adversarial federated bandit with full-information feedback is a special case of distributed online linear optimization, with the decision set being a $K-1$-dimensional simplex. Hence, Theorem~\ref{thm:lower-bound-full} immediately implies a regret lower bound for the distributed online linear optimization problem. To the best of our knowledge, this is the first non-trivial lower bound that relates the hardness of distributed online linear optimization problem to the algebraic connectivity of the communication network.

Leveraging the lower bound for the full-information setting, we show a lower bound for the bandit feedback setting.

\begin{theorem}[Bandit feedback]
    \label{thm:lower-bound-bandit}
    For any federated bandit algorithm $\pi$, there exists a graph $\mathcal{G}=(\mathcal{V}, \mathcal{E})$ with Laplacian matrix $M$ and a bandit feedback instance $L\in [0, 1]^{T\times N\times K}$ such that for some $v_1 \in \mathcal{V}$,
    \begin{equation} \label{lower-bound:bandit-feedback}
        R^{v_1}_T(\pi, L) = \Omega\left( \max\left\{ \sqrt{ \frac{1}{1+d_{v_1}}}\sqrt{KT}, \sqrt[4]{\frac{1+d_{\max}}{\lambda_{N-1}(M)} } \sqrt{T\log K} \right\} \right).
    \end{equation}
% \mv{Should $v$ in the right-hand side of the last equation be $v_1$?}
\end{theorem}

The proof is provided in Section~\ref{app:lower-bandit}. The lower bound contains two parts. The first part, derived from the information-theoretic argument in \cite{shamir2014fundamental},  captures the effect from bandit feedback. The second part is inherited from Theorem~\ref{thm:lower-bound-full} by the fact that the regret of an agent in bandit feedback setting cannot be smaller than its regret in full-information setting.

\subsection{\fedexp: a federated regret-minimization algorithm} 
\label{sec:upper-bound}

% \mv{*** Is qualification "near-optimal" justified given how  upper bound compares with lower bound? ***}
% \jy{*** In the discussion paragraph, we show that the ratio between regret upper bound of {\fedexp} and the lower bound is $\Tilde{O}(T^{1/6})$, I suppose this can justify the "near-optimality" of \fedexp.***} \mv{*** How about dependence on other parameters? In general, there is a danger in trying to oversell results -- I am not saying this is the case here, but better be careful with qualification of results. Maybe not stating "near-optimal" in the section title, but saying it in the text where appropriate discussion is provided. ***}

Inspired by the fact that the cumulative regret is related to the time need to reach consensus about a globally best arm,  we introduce a new federated bandit algorithm based on the gossip communication mechanism, called \fedexp. The details of {\fedexp} are described in Algorithm~\ref{algo:FedExp3}. We shall also show that {\fedexp} has a sub-linear cumulative regret upper bound which holds for all agents simultaneously.

The {\fedexp} algorithm is adapted from the Exp3 algorithm, in which each agent $v$ maintains an estimator $z^v_t \in \mathbb{R}^K$ of the cumulative losses for all arms and a tentative action selection distribution  $x^v_t\in[0, 1]^K$.
At the beginning of each time step $t$, each agent follows the action selection distribution $x^v_t$ with probability $1-\gamma_t$, or performs a uniform random exploration with probability $\gamma_t$. Once the action $a^v_t$ is sampled, the agent observes the associated loss $\ell^v_t(a^v_t)$ and then computes an importance-weighted loss estimator $g^v_t  \in \mathbb{R}^K$ using the sampling probability $p^v_t(a^v_t)$. Before $g^v_t$ is integrated into the cumulative loss estimator $z^v_{t+1}$, the agent communicates with its neighbors to average its cumulative loss estimator $z^v_t$.
% In this section, we present the {\fedexp} algorithm,  \mv{Provide some discussion for the definition of the algorithm.}
%\subsection{{\fedexp} algorithm}

\begin{algorithm}[h]
\caption{\fedexp}
\label{algo:FedExp3}
\SetKwInOut{Init}{Initialization}
\SetKwInOut{Input}{Input}
\Input{Non-increasing sequence of learning rates $\{\eta_t>0\}$ , non-increasing sequence of exploration ratios $\{\gamma_t>0\}$, and a gossip matrix $W\in [0, 1]^{N\times N}$.} 
\Init{$z_1^v(i) = 0, x^v_1(i)=1/K$ for all $i\in \mathcal{A}$ and $v\in \mathcal{V}$.}
\For{each time step $t=1,2,\dots, T$}{
    \For{each agent $v\in \mathcal{V}$}{
        compute the action distribution $p_t^v(i) = (1-\gamma_t) x^v_t(i) + \gamma_t/K$\;
        choose the action $a_t^v\sim p_t^{v}$\;
        compute the loss estimators
        $
        g^v_t(i) = \ell^v_t(i) \mathbb{I}\left\{a^v_t = i \right\} / p^v_t(i) 
        $ for all $i\in \mathcal{A}$\;
        update the gossip accumulative loss $z_{t+1}^v = \sum_{u: (u,v)\in \mathcal{E}} W_{u,v} z_t^u + g_t^v$ \tcp*{communication step}\;
        update the exploitation distribution $x^v_{t+1} = \frac{\exp\left(-\eta_t z_{t+1}^v(i)\}\right)}{\sum_{j\in A} \exp\left(-\eta_t z_{t+1}^v(j)\right)}$\;
    }
}
\end{algorithm}

The communication step is characterized by the \emph{gossip} matrix which is a doubly stochastic matrix $W\in [0, 1]^{N\times N}$ satisfying the following constraints
\begin{equation*}
    \sum_{v\in \mathcal{V}}W_{u,v} = \sum_{u\in \mathcal{V}}W_{u,v} = 1
\end{equation*}
and $W_{u,v} \geq 0$ where equality holds when $(u,v)\notin \mathcal{E}$.
This gossip communication step facilitates the agents to reach a consensus on the estimators of the cumulative losses of all arms, and hence allows the agents to identify a globally best arm in hindsight.
We present below an upper bound on the cumulative regret of each agent in \fedexp.
\begin{theorem}
\label{thm:upper-bound:static}
Assume that the network runs Algorithm~\ref{algo:FedExp3} with
$$
\gamma_t = \sqrt[3]{\frac{\left(C_W + \frac{1}{2}\right)K^2\log K}{t}}
$$
and
$$
\eta_t = \frac{\log K}{T \gamma_T} = \sqrt[3]{\frac{(\log K)^2}{ \left(C_W + \frac{1}{2}\right) K^2 T^2}}
$$
with
$
C_W = \min\{2\log T + \log N, \sqrt{N}\}/(1-\sigma_2(W)) +3.
$
Then, the expected regret of each agent $v\in \mathcal{V}$ is bounded as
$$
R_T^v = \Tilde{O} \left( \frac{1}{\sqrt[3]{1-\sigma_2(W)}} K^{2/3} T^{2/3}\right) 
$$
%$$
%R_T^v = \Tilde{O} \left( \frac{1}{\sqrt[3]{1-\sigma_2(W)}} \sqrt[3]{K^2\log K} T^{\frac{2}{3}}\right) 
%$$
where $\sigma_2(W)$ is the second largest singular value of $W$. 
% \mv{In the theorem, $\tilde{O}(\cdot)$ would hide some poly-log factors, but we write $\log(K)^{1/3}$ therein?}
\end{theorem}

\paragraph{Proof sketch}
Let $\hat{\ell}_t$ and $\bar{z}_t$ be the average instant loss estimator and average cumulative loss estimator, 
\begin{equation*}
    f_t = \frac{1}{N}\sum_{v\in \mathcal{V}} g^v_t \quad \text{and} \quad \bar{z}_t = \frac{1}{N}\sum_{v\in \mathcal{V}} z^v_t,
\end{equation*}
and let $y_t$ be action distribution that minimizes the regularized average cumulative loss estimator
$$
y_t(i) = \frac{\exp\left(-\eta_{t-1} \bar{z}_t(i)\right)}{\sum_{j\in A} \exp\left(-\eta_{t-1} \bar{z}_t(j)\right)}.
$$
The cumulative regret can be bounded by the sum of three terms
\begin{equation*}
    \begin{split}
    R^v_t &\leq 
    \underbrace{\mathbb{E}\left[\sum_{t=1}^T \left(\langle f_t, y^v_t \rangle - f_t (i^\ast) \right)\right]}_{\operatorname{FTRL}} \\ &+ 
      K \underbrace{\sum_{t=1}^T \eta_{t-1} \mathbb{E}\|z^{v}_t - \bar{z}_t\|_\ast}_{\operatorname{CONSENSUS}} + \underbrace{\sum_{t=1}^T \gamma_t}_{\operatorname{EXPLORATION}}
    \end{split}
\end{equation*}
where $i^\ast \in \arg\min_{i\in \mathcal{A}} \sum_{t=1}^T \Bar{\ell}_t(i)$ is a globally best arm in hindsight.

The $\operatorname{FTRL}$ term is a typical regret term from the classic analysis for the Follow-The-Regularized-Leader algorithm \citep{lattimore_szepesvari_2020}. The $\operatorname{CONSENSUS}$ term measures the cumulative approximation error generated during the consensus reaching process, which can be bounded using the convergence analysis of distributed averaging algorithm based on doubly stochastic matrices \citep{duchi2011dual, hosseini2013online}. The last $\operatorname{EXPLORATION}$ term can be bounded by specifying the time-decaying exploration ratio $\gamma_t$.
The full proof of Theorem~\ref{thm:upper-bound:static} is provided in Section~\ref{proof:upper-bound}.

The {\fedexp} algorithm is also a valid algorithm for the multi-agent adversarial bandit problem \citep{cesa2016delay} which is a special case of the doubly adversarial federated bandit problem when $\ell^v_t(i) = \ell_t(i)$ for all $v\in \mathcal{V}$. According to the distributed consensus process of \fedexp, each agent $v\in \mathcal{V}$ only communicates cumulative loss estimator values $z^v_t$, instead of the actual loss values $\ell^v_t(a^v_t)$, and the selection distribution $p^v_t$.
{\fedexp} can guarantee a sub-linear regret without the exchange of sequences of selected arm identities or loss sequences of agents, which resolves an open question raised in \cite{cesa2016delay}.

\paragraph{Choice of the gossip matrix} 
The gossip matrix $W$ can be constructed using the \emph{max-degree} trick in \cite{duchi2011dual}, i.e.,
$$
W  = I-\frac{D-A}{2(1+d_{\max})}
$$
where $D = \operatorname{diag}(d_1, \dots, d_N)$ and $A$ is the adjacency matrix of $\mathcal{G}$. This construction of $W$ requires that all agents have knowledge of the maximum degree $d_{\max}$, which can indeed be easily computed in a distributed system by nodes exchanging messages and updating their states using the maximum reduce operator.

% Before running the {\fedexp} algorithm, the agents can run a distributed averaging algorithm \citep{DBLP:journals/tit/BoydGPS06} to reach an consensus on $d_{\max}$. \mv{*** Why referring to this averaging algorithm for computing maximum value? This is a distributed \emph{selection problem}. It can indeed be easily computed in a distributed system by nodes exchanging messages and updating their states using maximum reduce operator. ***}

Another choice of $W$ comes from the effort to minimize the cumulative regret.
The leading factor $1/\sqrt[3]{1-\sigma_2(W)}$ in the regret upper bound of {\fedexp} can be minimized by choosing a gossip matrix $W$ with smallest $\sigma_2(W)$. Suppose that the agents have knowledge of the topology structure of the communication graph $\mathcal{G}$, then the agents can choose the gossip matrix to minimize their regret by solving the following convex optimization problem:
\begin{equation}
\begin{array}{rl}
\operatorname{minimize} & \left\|W-(1 / n) \mathbf{1 1}^T\right\|_2 \nonumber \\
\text { subject to} & W \geq 0, W \mathbf{1}=\mathbf{1}, W=W^T, \\
 & W_{i j}=0, \hbox{ for } (i, j) \notin \mathcal{E}  
\end{array}
\end{equation}
which has an equivalent semi-definite programming formulation as noted in \cite{boyd2004fastest}.
% Let $\Sigma$ be the Laplacian matrix of $\mathcal{G}$, \mv{Need to define the Laplacian matrix as there are several different definitions. Unclear why Laplacian matrix is mentioned in this context.} then $\Sigma\in \mathcal{M}_\mathcal{G}$.

\paragraph{Gap between upper and lower bounds} 
 The regret upper bound of {\fedexp} algorithm in Theorem~\ref{thm:upper-bound:static} grows sublinearly in the number of arms $K$ and horizon time $T$. There is only a small polynomial gap between the regret upper bound and the lower bound in Theorem~\ref{thm:lower-bound-bandit} with respect to these two parameters. The regret upper bound depends also on the second largest singular value $\sigma_2(W)$ of $W$. The related term in the lower bound in Theorem~\ref{thm:lower-bound-bandit} is the second smallest eigenvalue $\lambda_{N-1}(M)$ of the Laplacian matrix $M$. To compare these two terms, we point that 
 when the gossip matrix is constructed using the max-degree method, as discussed in Corollary 1 in \cite{duchi2011dual},
 $$ \frac{1}{\sqrt[3]{1- \sigma_2(W)}} \leq \sqrt[3]{2\frac{d_{\max }+1}{\lambda_{N-1}(M)}}.$$ 
 With respect to $\sqrt[4]{(d_{\max}+1)/\lambda_{N-1}(M)}$ in Theorem~\ref{thm:lower-bound-bandit}, there is only a small polynomial gap between the regret upper bound and the lower bound. 
 % \mv{May consider how to best slightly rephrase the last sentence so that it is clear. From the last above inequality, we have $(1/(1-\sigma_2(W)))^{1/3} \leq 2^{1/3} ((d_{\max}+1)/\lambda_{N-1}(M))^{1/3}$. Maybe it is sufficient to rephrase by saying " ...".}
 We note that a similar dependence on $\sigma_2(W)$ is present in the analysis of distributed optimization algorithms  \citep{duchi2011dual, hosseini2013online}.
 
\subsection{Numerical experiments}\label{sec:numerical-experiments}

We present experimental results for the {\fedexp} algorithm ($W$ constructed by the max-degree method) using both synthetic and real-world datasets. 
We aim to validate our theoretical analysis and demonstrate the effectiveness of {\fedexp} on finding a globally best arm in non-stationary environments. All the experiments are performed with 10 independent runs. The code for producing our experimental results is available online in the Github repository: \href{https://anonymous.4open.science/r/doubly-stochastic-federataed-bandit-7ECD}{[link]}.

\subsubsection{Synthetic datasets}

We validate our theoretical analysis of the {\fedexp} algorithm on synthetic datasets. The objective is two-fold. First, we demonstrate that the cumulative regret of {\fedexp} grows sub-linearly with time. Second, we examine the dependence of the regret on the second largest singular value of the gossip matrix. %is as stated in Theorem~\ref{thm:upper-bound:static}.

A motivation for finding a globally best arm in recommender systems is to provide recommendations for those users whose feedback is sparse.
In this setting, we construct a federated bandit setting in which a subset of agents will be activated at each time step and only activated agents may receive non-zero loss.
Specifically, we set $T=3,000$ with $N=36$ and $K=20$. At each time step $t$, a subset $U_t$ of $N/2$ agents are selected from $\mathcal{V}$ with replacement.
For all activated agents $U_t$, the loss for arm $i$ is sampled independently from Bernoulli distribution with mean $\mu_i = (i-1)/(K-1)$. All non-activated agents receive a loss of $0$ for any arm they choose at time step $t$.

We evaluate the performance of {\fedexp} on different networks, i.e. for a complete graph, a $\sqrt{N}$ by $\sqrt{N}$ grid network, and random geometric graphs. 
The random geometric graph RGG($d$) is constructed by uniform random placement of each node in $[0, 1]^2$ and connecting any two nodes whose distance is less or equal to $d$ \citep{penrose2003random}.
Random geometric graphs are commonly used for modeling spatial networks.

\begin{figure}[t]
    \centering
    \includegraphics[width=.4\textwidth]{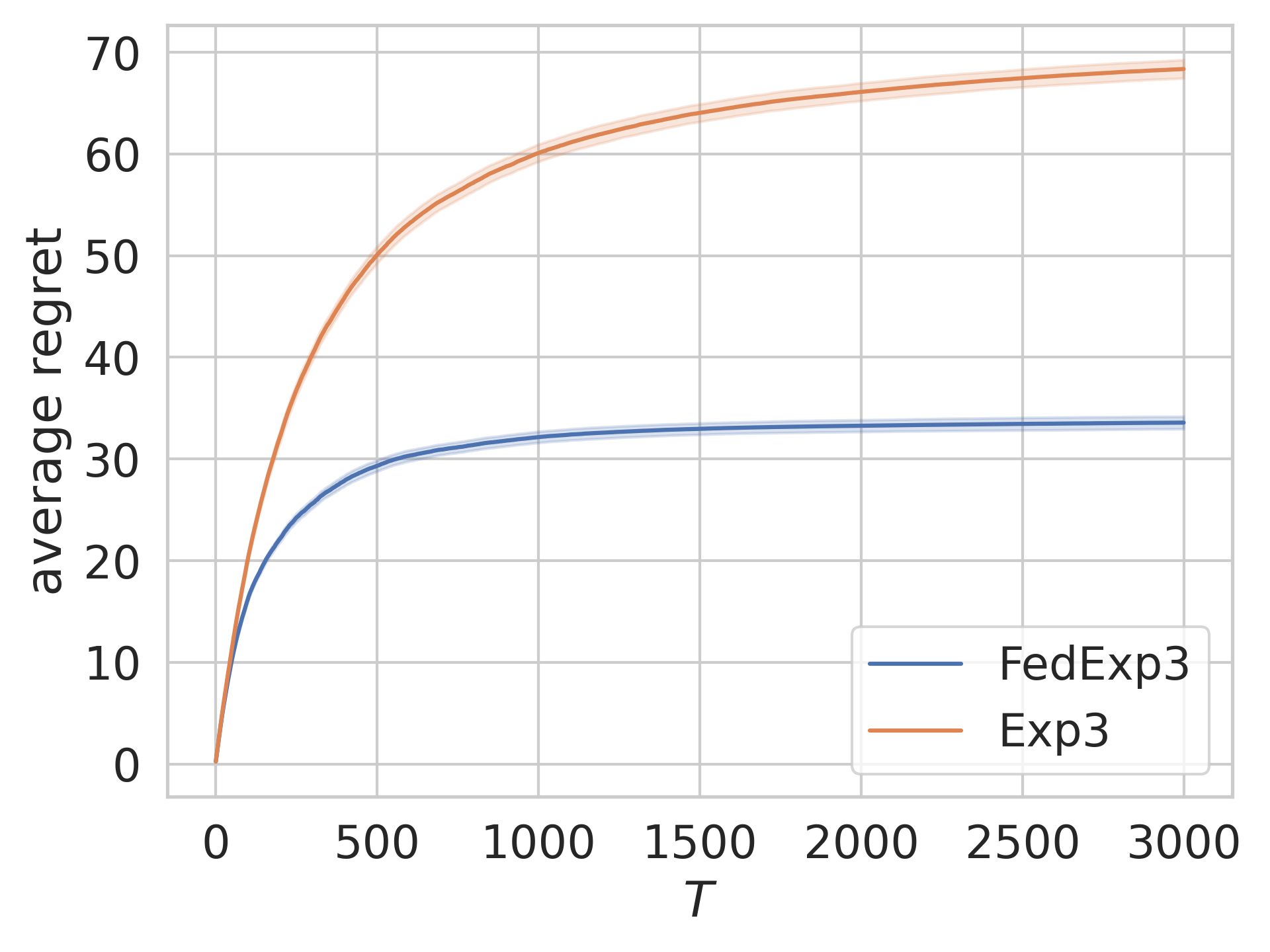}
    \includegraphics[width=.4\textwidth]{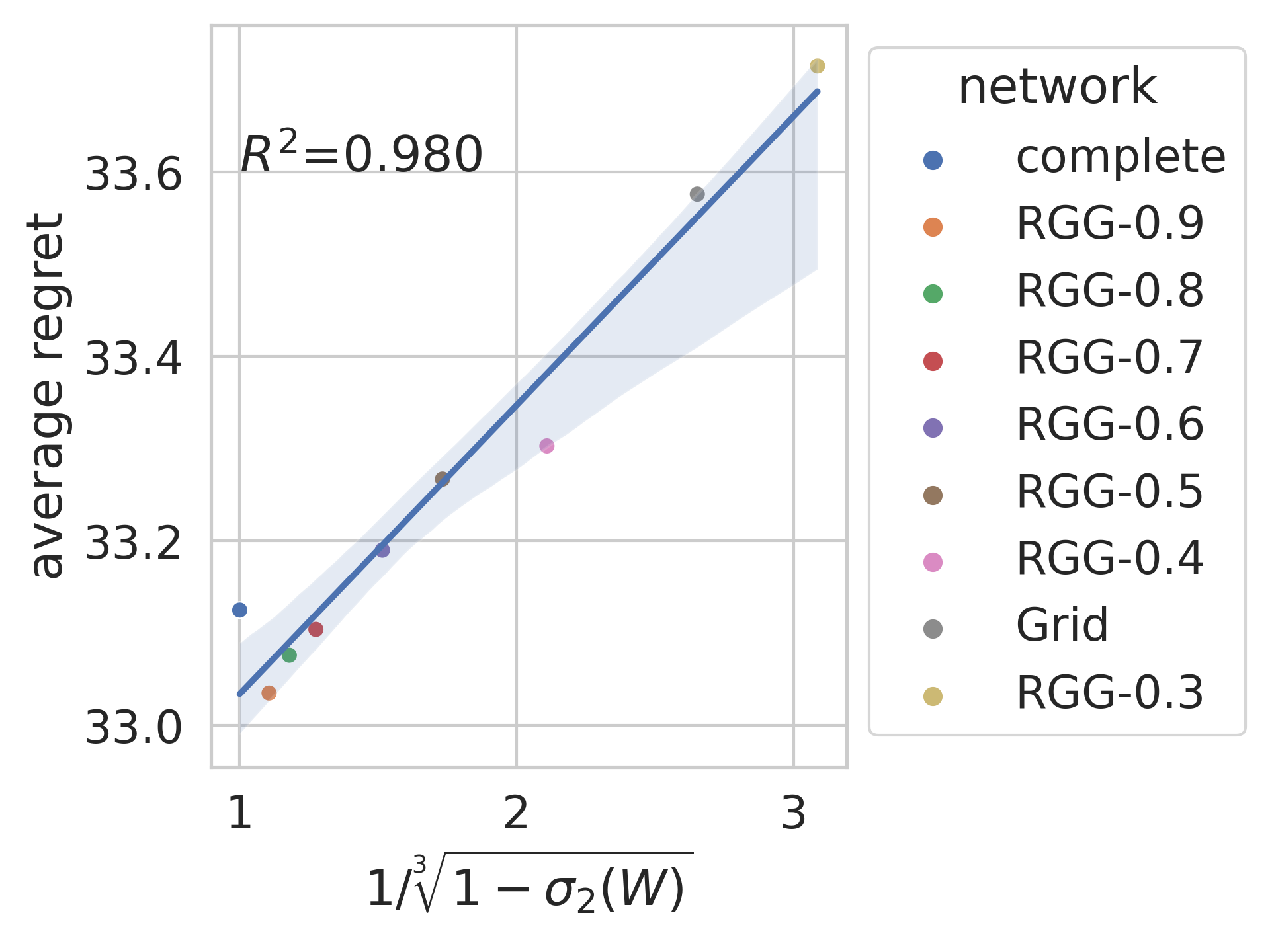}
    \caption{(Top) The average cumulative regret, i.e. $\sum_{v\in\mathcal{V}} R^v_T / N$, versus $T$, for {\fedexp} and Exp3 on the grid graph. (Down) The average cumulative regret versus $(1-\sigma_2(W))^{-1/3}$ for {\fedexp} on different networks at $T=3000$.
    }
    \label{fig:regret}
\end{figure}

In our experiments, we set $d\in \{0.3, \dots, 0.9\}$. The results in 
Figure~\ref{fig:regret} confirm that the cumulative regret of the {\fedexp} algorithm grows sub-linearly with respect to time and suggest that the cumulative regret of {\fedexp} is proportional to $\left(1-\sigma_2(W)\right)^{-1/3}$. This is compatible with the regret upper bound in Theorem~\ref{thm:upper-bound:static}. 

\subsubsection{MovieLens dataset: recommending popular movie genres}

% \mv{*** What is the goal of this experimental analysis? Figure~\ref{fig:movie-lens} shows regret curves for different networks --- what do we want to show with this? Is there something we want to show that is of interest in the specific context of movie recommendations? ***}
We compare {\fedexp} with a UCB-based federated bandit algorithm in a movie recommendation scenario using a real-world dataset. In movie recommendation settings, users' preferences over different genres of movies can change over time. In such non-stationary environments, we demonstrate that a significant performance improvement can be achieved by {\fedexp} against the GossipUCB algorithm (we refer to as \gucb) proposed in \cite{zhu2021federated}, which is defined for stationary settings.

We evaluate the two algorithms using the \href{https://grouplens.org/datasets/movielens/latest/}{MovieLens-Latest-full} dataset which contains 58,000 movies, classified into 20 genres, with 27,000,000 ratings (rating scores in $\{0.5, 1,\ldots, 5\}$) from 280,000 users. Among all the users, there are 3,364 users who rated at least one movie for every genre. We select these users as our agents, i.e. $N = 3,364$, and the 20 genres as the arms to be recommended, i.e. $K = 20$.

\begin{figure}[t]
    \centering
    \includegraphics[width=.3\textwidth]{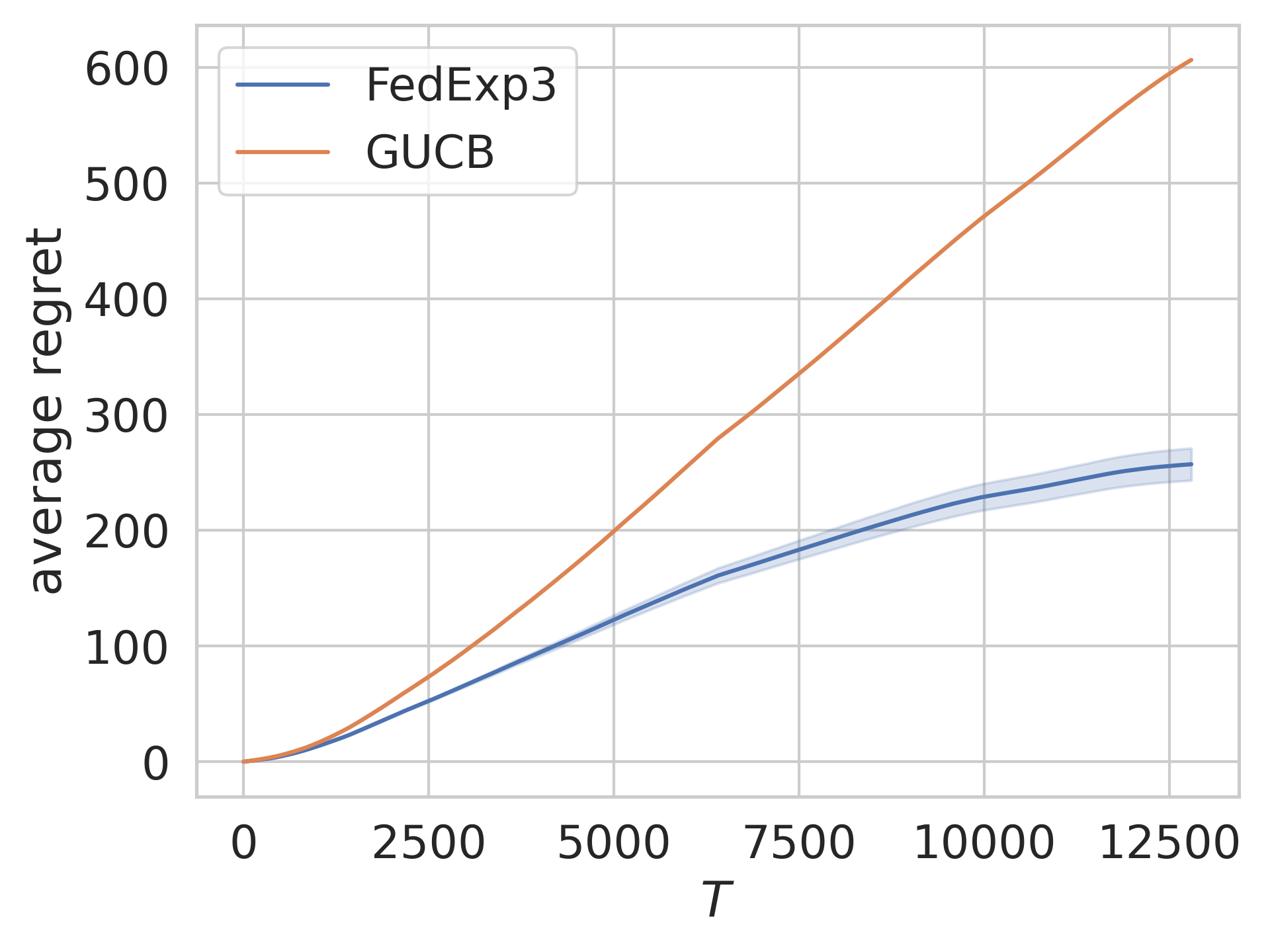}
    \includegraphics[width=.3\textwidth]{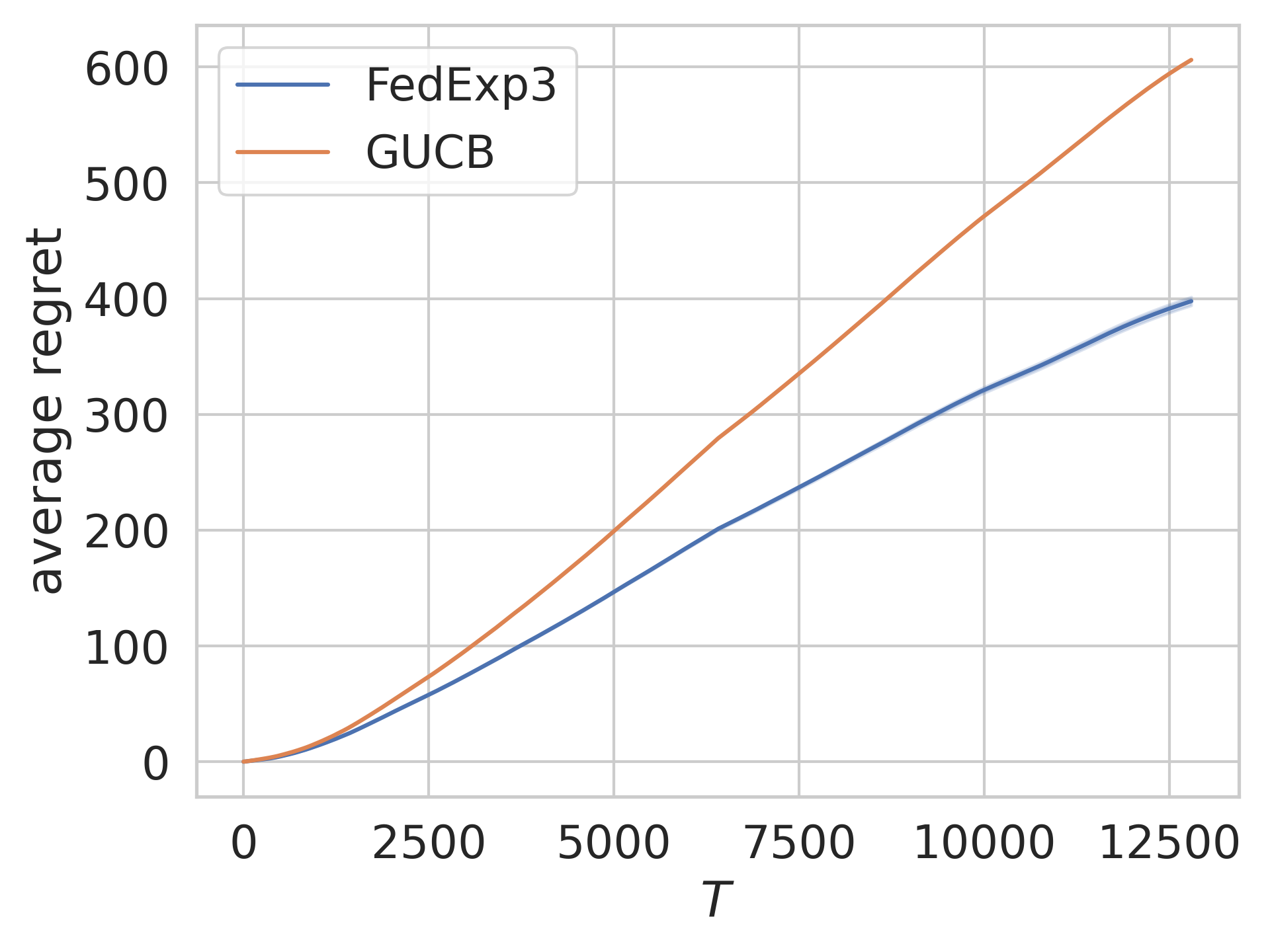}
    \includegraphics[width=.3\textwidth]{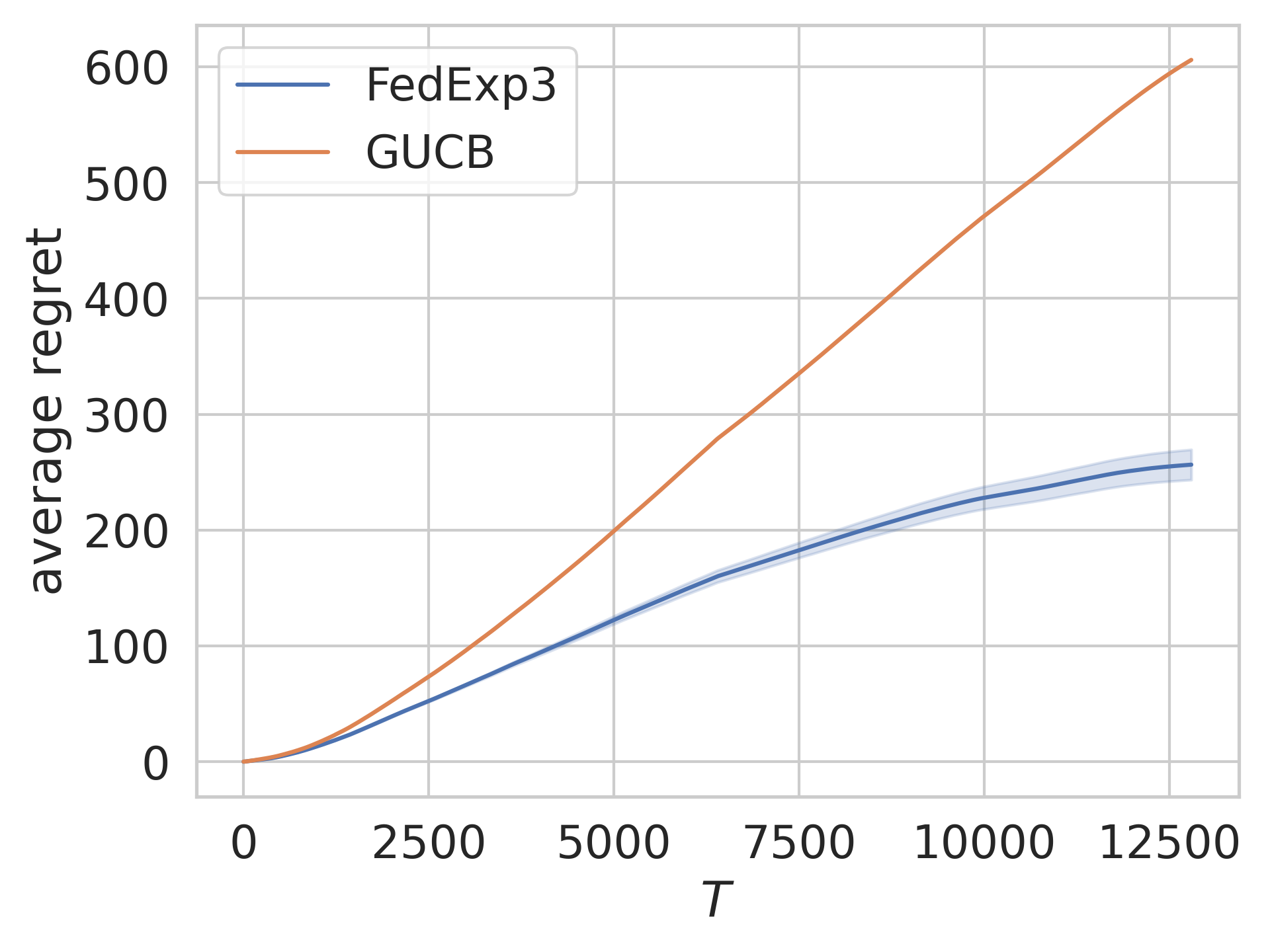}
    \caption{The average cumulative regret versus horizon time for {\fedexp} and \gucb\ in the movie recommendation setting with the communication networks: (top) complete graph, (mid) the grid network, and (down) RGG($0.5$). 
    % \mv{*** Same comments for $x$ and $y$ labels and plot titles as for Figure~1. ***}
    }
    \label{fig:movie-lens}
\end{figure}

We create a federated bandit environment for movie recommendation based on this dataset. 
Let $m^v(i)$ be the number of ratings that agent $v$ has for genre $i$. We set the horizon $T= \max_{v\in \mathcal{N}} m^v(i) =12,800 $.
To reflect the changes in agents' preferences
over genres as time evolves, we sort the ratings in an increasing order by their Unix timestamps and construct the loss tensor in the following way.
Let $r^v_j (i)$ be the $j$-th rating of agent $v$ on genre $i$, the loss of recommending an movie to agent $v$ of genre $i$ at time step $t$ is defined as 
    \begin{equation*}
        \ell^v_t(i) = \frac{5.5-r^v_j (i)}{5.5}
    \end{equation*}
for $t \in \left[(j-1)\left\lfloor \frac{T}{m^v(i)} \right\rfloor, j\left\lfloor \frac{T}{m^v(i)} \right\rfloor\right)$.
% Hence the doubly adversarial federated bandit problem is to recommend each agent the averaged best genre, e.g. \textit{Film Noir}. 
The performance of {\fedexp} and \gucb\ is shown in Figure~\ref{fig:movie-lens}. 
The results demonstrate that {\fedexp} can outperform \gucb\ by a significant margin for different communication networks.

\subsection{Proofs}
\label{sec::chapter-icml-app}

\paragraph{Notation} For a vector $x$, we use $x(i)$ to denote the $i$-th coordinate of $x$. We define $\mathcal{F}_{t} = \bigcup_{v\in \mathcal{V}} \mathcal{F}^v_{t}$ where $\mathcal{F}^v_t$ is the sequence of agent $v$'s actions and feedback up to time step $t$, i.e., $\mathcal{F}^v_t = \bigcup_{s=1}^t \{a^v_s, I^v_s \}$. 

\subsubsection{Proof of Theorem~\ref{thm:lower-bound-full}}
\label{app-lower-full}

We first define a new class of cluster-based distributed online learning procedure, referred to as \emph{cluster-based federated algorithms},  in which the delay only occurs when the communication is between different clusters.
The regret lower bound for federated bandit algorithms will be no less than the regret lower bound for cluster-based federated algorithms, as shown in Lemma~\ref{lm:monotone}. Then we show in Lemma~\ref{lm:delta-g} that there exists a special graph in which there exist two clusters of agents $A$ and $B$ with distance $d(A, B) = \min_{u\in A, v\in B}d(u, v) = \Omega\left(\sqrt{(d_{\max}+1)/\lambda_{N-1}(M)}\right)$. Then, we consider an instance where only agents in cluster $A$ receive non-zero losses. 
Based on a reduction argument, the cumulative regrets of agents in cluster $B$ are the same as (up to a constant factor) the cumulative regrets in a single-agent adversarial bandit setting with feedback delay $d(A, B)$ (see Lemma~\ref{lm:delay} in Section~\ref{app-lower-full}). Hence, one can show that the cumulative regret of agents in cluster $B$ is $\Omega \left(\sqrt{d(A,B)}\sqrt{T\log K}\right)$.  

We denote with $d(\mathcal{U}, \mathcal{U}^\prime)$ the smallest distance between any two nodes in $\mathcal{U}, \mathcal{U}^\prime \subset \mathcal{V}$, i.e.
$$
d(\mathcal{U}, \mathcal{U}^\prime) = \min_{u\in U, u^\prime\in \mathcal{U}^\prime} d(u, u^\prime)
$$
where $d(u,v)$ is the length of a shortest path connecting $u$ and $u^\prime$.
\begin{definition}[Cluster-based federated algorithms]
    A cluster-based federated algorithm is a multi-agent learning algorithm defined by a partition of graph $\bigcup_{r} \mathcal{U}_r = \mathcal{V}$ 
    where $\mathcal{U}_r$ is called cluster.
    In the cluster-based federated algorithm, at each round $t$, the action selection probability $p^v_t$ of agent $v\in \mathcal{U}_r$ depends on the history information up to round $t-d(\mathcal{U}_r, \mathcal{U}_{r^\prime})-1$ of all agents $u^\prime\in \mathcal{U}_{r^\prime}$.
\end{definition}

Note that when all agents are in the same cluster $\mathcal{V}$, the centralized federated algorithm in \cite{reda2022near} is an instance of a cluster-based federated algorithm.

\begin{lemma}[Monotonicity]\label{lm:monotone}
    Let $\Pi$ and $\Pi^\prime$ be two sets of all cluster-based federated algorithms with two partitions $\bigcup_{r} \mathcal{U}_r$ and $\bigcup_{s} \mathcal{U}^\prime_s$, respectively. Suppose for any cluster $\mathcal{U}^\prime_{s}$ of $\pi^\prime$, there exists a cluster $\mathcal{U}_r$ of $\pi$ such that $\mathcal{U}^\prime_{s}\subset \mathcal{U}_r$, then 
    $$ \Pi^\prime \subset \Pi \quad \text{and} \quad \min_{\pi\in\Pi} R^v_T(\pi, L ) \leq  \min_{\pi^\prime\in\Pi^\prime} R^v_T(\pi^\prime, L)$$
    for any $L\in [0, 1]^{T\times N\times K}$ and any $v\in \mathcal{V}$.
\end{lemma}
\begin{proof}
    It suffices to show $\Pi^\prime \subset \Pi$.
    Consider a cluster-based federated algorithm $\pi^\prime \in \Pi^\prime$. For any agent $v\in \mathcal{V}$, let $\mathcal{U}^\prime_s$ be the cluster of $v$ in $\Pi^\prime$. By definition of cluster-based procedure, agent $v$'s action selection distribution probability $p^v_t$ depends on the history information up to round $t-d(\mathcal{U}^\prime_s, \mathcal{U}^\prime_{h})-1$ of all agents $u^\prime\in \mathcal{U}^\prime_{h}$.
    
    By the assumption, there exists two subset $\mathcal{U}_{r_1}, \mathcal{U}_{r_2}\subset \mathcal{V}$ such that $\mathcal{U}^\prime_{s} \subset \mathcal{U}_{r_1}$ and $\mathcal{U}^\prime_{h} \subset \mathcal{U}_{r_2}$. Hence $d(\mathcal{U}_{r_1}, \mathcal{U}_{r_2}) \leq d(\mathcal{U}^\prime_{s}, \mathcal{U}^\prime_{h})$, from which it follows $t-d(\mathcal{U}^\prime_{s}, \mathcal{U}^\prime_{h})-1 \leq t - d(\mathcal{U}_{r_1}, \mathcal{U}_{r_2}) - 1$. Hence $\pi^\prime\in \Pi$ which completes the proof.
\end{proof}

% We present a regret lower bound on any distributed online learning procedure with 2 clusters. \mv{This is a strange sentence as what immediately follows is a lemma from literature.}

\begin{lemma}\label{lm:delta-g}
There exists a graph $\mathcal{G}=(\mathcal{V}, \mathcal{E})$ with $N$ nodes and a matrix $M\in \mathcal{M}_{\mathcal{G}}$, together with two subsets of nodes $I_0, I_1 \subset \mathcal{V}$ of size $\left|I_0\right|=\left|I_1\right| \geq N/4$ and such that
$$
d\left(I_0, I_1\right) \geq \Tilde{\Delta},
$$
where $d\left(I_0, I_1\right)$ is the shortest-path distance in $\mathcal{G}$ between the two sets and $$\Tilde{\Delta} =  \frac{\sqrt{2}}{3} \sqrt{\frac{1 + d_{\max}} {\lambda_{N-1}(M)}}.$$
\end{lemma}

\begin{proof}
    From Lemma~24 in \cite{scaman2019optimal}, there exists exists a graph $\mathcal{G}=(\mathcal{V}, \mathcal{E})$ with $N$ nodes and a matrix $M\in \mathcal{M}_{\mathcal{G}}$, together with two subsets of nodes $I_0, I_1 \subset \mathcal{V}$ of size $\left|I_0\right|=\left|I_1\right| \geq N/4$ and such that
$$
d\left(I_0, I_1\right) \geq \frac{\sqrt{2}}{3} \sqrt{\frac{\lambda_1(M)} {\lambda_{N-1}(M)}}.
$$
We show that $\lambda_1(M) \geq 1 + d_{\max}$.
To see this, note that $\lambda_1(M)$ is the Rayleigh quotient $\max _{x \neq 0} \frac{x^T M x}{x^T x}$.
By the definition of the Laplacian matrix,
$$
x^T M x=\sum_{(v, u) \in \mathcal{E}}\left(x_v-x_u\right)^2.
$$
Let $v$ be a vertex whose degree is $d_{\max}$ and
$$
x_u:= \begin{cases} \sqrt{\frac{d_{\max}}{1+d_{\max}}} & \text { if } u=v \\ - \frac{1}{\sqrt{d_{\max}}\sqrt{1+d_{\max}}} & \text { if } u \neq v \text { and } v_i \text { is adjacent to } v_j \\ 0 & \text { otherwise }\end{cases}
$$
then
\begin{equation*}
    \begin{split}
        \sum_{(v, u) \in \mathcal{E}}\left(x_v-x_u\right)^2
        &= 
        d_{\max} \left( \sqrt{\frac{d_{\max}}{1+d_{\max}}} +  \frac{1}{\sqrt{d_{\max}}\sqrt{1+d_{\max}}} \right)^2 \\
        &= 
        d_{\max} \left( \frac{1}{\sqrt{d_{\max}}\sqrt{1+d_{\max}}} \right)^2 \left(d_{\max}+1\right)^2 \\
        &= 1 + d_{\max} \\
    \end{split}
\end{equation*}
% \mv{*** The last equation seems incorrect. ***}
and 
$$
\sum_{u\in \mathcal{V}} x_u^2 = \frac{d_{\max}}{1+d_{\max}} + d_{\max} \frac{1}{d_{\max}(1+d_{\max})} = 1.
$$
Hence, $\lambda_1(M) \geq 1+ d_{\max}$.
\end{proof}

Let $I_0, I_1$ be two subsets of nodes satisfying
$$d(I_0, I_1)\geq \Tilde{\Delta}\quad \text{and} \quad |I_0| = |I_1| = N/4.$$
The number of rounds needed to communicate between any node in $I_0$ and any node $I_1$ is at least $\Tilde{\Delta}$.
\begin{lemma} \label{lm:delay}
    Let  $v_0\in I_0$ and $v_1\in \mathcal{V}\backslash I_0$. Consider a cluster-based federated algorithm with clusters
    $I_0$ and $V\backslash I_0$. Then, any distributed online learning algorithm $\sigma$ for full information feedback setting has an expected regret 
    $$
    R^{v_1}_T \geq \frac{1-o(1)}{4} \sqrt{\frac{\left(\Tilde{\Delta}+1\right)}{2}T\log K}
    $$
    as $T\to\infty$.
\end{lemma}

\begin{proof}
Consider an online learning with expert advice problem with the action set $\mathcal{A}$ over $B$ rounds \citep{cesa1997use}.
Let $\ell_1^\prime, \dots, \ell_B^\prime$ be an arbitrary sequence of losses and $p^\prime_b$ be the action selection distribution at round $b$.
We show that $\sigma$ can be used to design an algorithm for this online learning with expert advice problem, adapted from \cite{cesa2016delay}.

Consider the loss sequences $\{\ell^v_t\}_{t=1}^T$ for each $v\in \mathcal{V}$ with $T = (\Tilde{\Delta}+1)B$ such that
$$
\ell^v_t =
\begin{cases}
    \ell^\prime_{\lceil t/(\Tilde{\Delta}+1) \rceil}, & v\in I_0 \\
    0 & \text{otherwise.} \\
\end{cases}
$$
Let $p^v_t$ be the action select distribution of agent $v\in \mathcal{V}$ running the algorithm $\sigma$.
Define the algorithm for the online learning with expert advice problem as follows:
$$
p^\prime_b =\frac{1}{\Tilde{\Delta}+1} \sum_{s=1}^{\Tilde{\Delta}+1} p^{v_1}_{(\Tilde{\Delta}+1)(b-1)+s}
$$
where $p^v_t = (1/k,\dots,1/k)$ for all $t\leq 1$ and $v\in \mathcal{V}$.

Note that $p^\prime_b$ is defined by $p^{v_1}_{(\Tilde{\Delta}+1)(b-1)+1}, \dots, p^{v_1}_{(\Tilde{\Delta}+1)b}$.
These are in turn defined by $\ell_1^{v_0},\dots, \ell_{(\Tilde{\Delta}+1)(b-1)}^{v_0}$ by the definition of cluster-based communication protocol. Also note that $\lceil t/(\Tilde{\Delta}+1) \rceil \leq b - 1$ for $t\leq (\Tilde{\Delta}+1)b$, hence $p^\prime_b$ is determined by $\ell_1^\prime, \dots, \ell_{b-1}^\prime$.
Therefore $p^\prime_1,\dots, p^\prime_B$ are generated by a legitimate algorithm for online learning with expert advice problem.

Note that the cumulative regret of agent $v_1$ is 
\begin{eqnarray}
    \label{eq:lower-1}
    \sum_{t=1}^T \langle p^{v_1}_t, \bar{\ell}_t \rangle 
    & = & \frac{1}{N} \sum_{t=1}^T \left[ \sum_{v\in I_0} \langle p^{v_1}_t, \ell^v_t \rangle + \sum_{v\not\in I_0} \langle p^{v_1}_t, \ell^v_t \rangle\right]  \nonumber \\ 
    & = & \frac{1}{4} \sum_{t=1}^T \langle p^{v_1}_t, \ell^\prime_{\lceil t/(\Tilde{\Delta}+1) \rceil} \rangle \nonumber \\
    & = & \frac{1}{4} \sum_{b=1}^B \sum_{s=1}^{\Tilde{\Delta}+1} \langle p^{v_1}_{(\Tilde{\Delta}+1)(b-1)+s}, \ell^\prime_{b} \rangle \nonumber \\
    & = & \frac{\Tilde{\Delta}+1}{4} \sum_{b=1}^B  \langle p^{\prime}_{b}, \ell^\prime_{b} \rangle
\end{eqnarray}
where the second equality comes from the definition of $\ell^v_t$ and the fourth equality comes from the definition of $p^{\prime}_b$.

Also note that 
\begin{eqnarray}
\label{eq:lower-2}
\min_{i\in \mathcal{A}} \sum_{t=1}^T \bar{\ell}_t(i)
&=& \frac{1}{4} \min_{i\in \mathcal{A}} \sum_{t=1}^T \ell^\prime_{\lceil t/(\Tilde{\Delta}+1) \rceil} (i) \nonumber \\
&=& \frac{\Tilde{\Delta}+1}{4} \min_{i\in \mathcal{A}} \sum_{b=1}^B \ell^\prime_{b} (i).
\end{eqnarray}
From (\ref{eq:lower-1}) and (\ref{eq:lower-2}), it follows that
$$
\sum_{t=1}^T \langle p^{v_1}_t, \bar{\ell}_t \rangle  - \min_{i\in \mathcal{A}} \sum_{t=1}^T \bar{\ell}_t(i)
= \frac{\Tilde{\Delta}+1}{4}
\left[ \sum_{b=1}^B  \langle p^{\prime}_{b}, \ell^\prime_{b} \rangle - \min_{i\in \mathcal{A}} \sum_{b=1}^B \ell^\prime_{b} (i)\right].
$$
There exists a sequence of losses $\ell_1^\prime, \dots, \ell_B^\prime$ such that for any algorithm for online learning with expert advice problem, the expected regret satisfies \citep[Theorem~3.7]{cesa2006prediction}, 
% \mv{Strange citation --- could you refer to a specific theorem in \cite{cesa1997use}?}
$$
\sum_{b=1}^B  \langle p^{\prime}_{b}, \ell^\prime_{b} \rangle - \min_{i\in \mathcal{A}} \sum_{b=1}^B \ell^\prime_{b} (i)
\geq
(1-o(1))\sqrt{\frac{B}{2} \ln K}.
$$
Hence, we have 
$$
\sum_{t=1}^T \langle p^{v_1}_t, \bar{\ell}_t \rangle  - \min_{i\in \mathcal{A}} \sum_{t=1}^T \bar{\ell}_t(i)
\geq
\frac{1-o(1)}{4} \sqrt{(\Tilde{\Delta}+1)\frac{T}{2}  \ln K}.
$$
\end{proof}

\subsubsection{Proof of Theorem~\ref{thm:lower-bound-bandit}} \label{app:lower-bandit}

% \mv{Same comment here as for the proof in the preceeding section - should have a paragraph discussing main steps of the proof, pointing how known results are leveraged, and higlighting any novel ideas.}
The lower bound contains two parts. The first part is derived by using information-theoretic arguments in \cite{shamir2014fundamental} and it  captures the effect of bandit feedback. The second part is inherited from the full-information feedback lower bound in Theorem~\ref{thm:lower-bound-full} by the fact that the regret of an agent in the bandit feedback setting cannot be smaller than the regret in the full-information setting.

Consider a centralized federated algorithm with all the agents in the same cluster $\mathcal{V}$, denoted as $\Pi^C$.
Note that by Lemma~\ref{lm:monotone}, for a federated bandit algorithm $\Pi^G$,
$$ \Pi^G \subset \Pi^C \quad \text{and} \quad \min_{\pi^\prime\in\Pi^C} R^v_T(\pi^\prime, L ) \leq  \min_{\pi\in\Pi^G} R^v_T(\pi, L)$$
for any $L\in [0, 1]^{T\times N\times K}$ and any $v\in \mathcal{V}$.

% For any federated bandit algorithm $\pi\in \Pi^G$, there exists a centralized federated algorithm $\pi^\prime\in \Pi^C$ such that 
% $$ R^v_T(\pi^\prime, L ) \leq  R^v_T(\pi, L)$$
% for any $L\in [0, 1]^{T\times N\times K}$ and any $v\in \mathcal{V}$. \mv{Should this para be removed?}

For any $\pi^\prime\in \Pi^C$, at each round $t$, every agent $v\in \mathcal{V}$ receives $O(|\mathcal{N}(v)|)$ bits since its neighboring agents can choose at most $|\mathcal{N}(v)|$ distinct actions. By Theorem~4 in \cite{shamir2014fundamental}, there exists some distribution $\mathcal{D}$ over $[0,1]^K$ such that loss vectors $\Bar{\ell}_t \overset{i.i.d}{\sim} \mathcal{D}$ for all $t=1,2,\dots, T$  and $\min_{i\in \mathcal{A}} \mathbb{E}\left[ \sum_{t=1}^T \Bar{\ell}_t(a_t(v)) - \sum_{t=1}^T\Bar{\ell}_t(i)\right] = \Omega\left(\min\{T, \sqrt{KT/(1+|\mathcal{N}(v)|)}\}\right)$.

Hence, it follows that
\begin{eqnarray}
    \min_{L} R^v_T(\pi, L)
    &\geq& \min_{L} R^v_T(\pi^\prime, L ) \nonumber \\ 
    & \geq &
    \mathbb{E}_{\Bar{\ell}_t\sim \mathcal{D}}\left[\sum_{t=1}^T\Bar{\ell}_t(a_t(v)) - \min_{i\in \mathcal{A}} \sum_{t=1}^T \Bar{\ell}_t(i)\right] \nonumber \\
    &\geq& 
    \max_{i\in \mathcal{A}} \mathbb{E}_{\Bar{\ell}_t\sim \mathcal{D}}\left[\sum_{t=1}^T\Bar{\ell}_t(a_t(v)) -  \sum_{t=1}^T \Bar{\ell}_t(i)\right] \nonumber \\
    &\geq& \min_{i\in \mathcal{A}}
    \mathbb{E}_{\Bar{\ell}_t\sim \mathcal{D}}\left[\sum_{t=1}^T\Bar{\ell}_t(a_t(v)) - \sum_{t=1}^T \Bar{\ell}_t(i)\right] \nonumber \\ 
    &=& \Omega\left(\min\{T, \sqrt{KT/(1+|\mathcal{N}(v)|)}\}\right) \nonumber 
\end{eqnarray}
where the third inequality comes from Jensen's inequality.

Also note that any federated bandit algorithm for bandit feedback setting is also a federated bandit algorithm for full-information setting, from which it follows 
\begin{eqnarray}
    \min_{L} R^v_T(\pi, L) 
    &\geq&   
     \max\left\{ \Omega\left(\min\{T, \sqrt{KT/(1+|\mathcal{N}(v)|)}\}\right), \Omega \left(\sqrt[4]{\frac{1+d_{\max}}{\lambda_{N-1}(M)} } \sqrt{ T\log K} \right) \right\} \nonumber \\
    &=&
    \Omega\left( \min\left\{T, \max\left\{\sqrt{K/(1+|\mathcal{N}(v)|)}, \sqrt[4]{\frac{1+d_{\max}}{\lambda_{N-1}(M)} } \sqrt{\log K} \right\}\sqrt{T}\right\} \right). \nonumber 
\end{eqnarray}

\subsubsection{Auxiliary lemmas}
Here we present some auxiliary lemmas which are used in the proof of Theorem~\ref{thm:upper-bound:static}.
% \mv{Explain for what are lemmas presented in this section used for.}
Recall that $\hat{\ell}_t$ and $\bar{z}_t$ are the average instant loss estimator and average cumulative loss, 
\begin{equation*}
    f_t = \frac{1}{N}\sum_{v\in \mathcal{V}} g^v_t \quad \text{and} \quad \bar{z}_t = \frac{1}{N}\sum_{v\in \mathcal{V}} z^v_t
\end{equation*}
and $y_t$ is action distribution to minimize the regularized average cumulative loss
$$
y_t(i) = \frac{\exp\left(-\eta_{t-1} \bar{z}_t(i)\right)}{\sum_{j\in A} \exp\left(-\eta_{t-1} \bar{z}_t(j)\right)}.
$$
\begin{lemma}
\label{lm:z-bar}
For each time step $t=1,\dots,T$,
\begin{equation*}
    \bar{z}_{t+1} = \bar{z}_t +  f_t
\end{equation*}
and
$$\max\{ \|g_t^v\|_{\ast}, \|f_t\|_{\ast}\} \leq \frac{K}{\gamma_t}.
$$
\end{lemma}
\begin{proof}
\begin{eqnarray}
\bar{z}_{t+1} 
&=& \frac{1}{N}\sum_{v\in \mathcal{V}} z^v_{t+1}\nonumber \\
&=& \frac{1}{N}\sum_{v\in \mathcal{V}} \sum_{u: (u,v)\in \mathcal{E}} W_{u,v} z_t^u + \frac{1}{N}\sum_{v\in \mathcal{V}}g_t^v\nonumber \\
&=& \frac{1}{N}\sum_{v\in \mathcal{V}} z^v_{t} + \frac{1}{N}\sum_{v\in \mathcal{V}}g_t^v\nonumber \\
&=& \bar{z}_t + f_t \nonumber
\end{eqnarray}
where the second equality comes from Line~7 in Algorithm \ref{algo:FedExp3} and the third equality comes from the double-stochasticity of $W$.

Noting that $p^v_t(i) \geq \gamma / K$ for all $v\in \mathcal{V}$, $i\in \mathcal{A}$ and $t\in \{1,\ldots, T\}$, it follows that
$$
\|g^v_t\|_\ast = \frac{\ell^v_t(a^v_t)}{p^v_t(a^v_t)} \leq \frac{K}{\gamma_t}
\quad
\text{and}
\quad
\|f_t\|_\ast \leq \frac{1}{N}\sum_{v\in \mathcal{V}}  \| g^v_t\|_\ast \leq \frac{K}{\gamma_t}.
$$
\end{proof}

\begin{lemma}\label{lm:loss-estimator}
For any $v\in \mathcal{V}$ and $t\geq 1$, it holds that
$$
\mathbb{E}\left[g_t^v \mid \mathcal{F}_{t-1}\right] = \ell^v_t \text { and } \mathbb{E}\left[f_t \mid \mathcal{F}_{t-1}\right] = \Bar{\ell}_t
$$
with
$$
\mathbb{E}\left[\|f_t\|_\ast \right] \leq K \text { and } \mathbb{E}\left[\|f_t\|_\ast^2 \right] \leq \frac{K^2}{\gamma_t}.
$$
\end{lemma}
\begin{proof}
Note that $p_t^v$ is determined by $\mathcal{F}_{t-1}$, hence
$$
\mathbb{E}\left[g_t^v(i) \mid \mathcal{F}_{t-1}\right] = \frac{\ell^v_t(i) }{p^v_t(i) }
\mathbb{E}\left[\mathbb{I}\left\{a^v_t = i \right\} \mid \mathcal{F}_{t-1}\right]
=
\frac{\ell^v_t(i) }{p^v_t(i) } p^v_t(i)
= \ell^v_t(i) 
$$
and
$$
\mathbb{E}\left[\|g_t^v\|_\ast \right]
= \mathbb{E}\left[\frac{\ell_t^v(a_t^v)}{p_t^v(a^v_t)} \right]
= \mathbb{E}\left[ \mathbb{E}\left[\frac{\ell_t^v(a_t^v)}{p_t^v(a^v_t)}  \mid \mathcal{F}_{t-1} \right] \right]
= \mathbb{E}\left[ \sum_{i\in \mathcal{A}} p^v_t(i) \frac{\ell_t^v(i)}{p_t^v(i)} \right]
= \sum_{i\in \mathcal{A}} \ell_t^v(i) \leq K
$$
where the last inequality comes from $\ell_t^v(i) \leq 1$.
Since $f_t (i) = \frac{1}{N} \sum_{v\in \mathcal{V}} g^v_t$,
it follows that
$$
\mathbb{E}\left[f_t (i) \mid \mathcal{F}_{t-1}\right]
= \frac{1}{N}\sum_{v\in \mathcal{V}} \ell^v_t(i) = \Bar{\ell}_t(i)
$$
and
$$
\mathbb{E}\left[\|f_t\|_\ast \right] \leq \frac{1}{N}\sum_{v\in \mathcal{V}} \mathbb{E}\left[\|g_t^v\|_\ast \right] \leq K
$$
which comes from Jensen's inequality.
Notice that
\begin{equation*}
    \begin{split}
        \mathbb{E}\left[\|g_t^v\|_\ast^2 \right]
        &= \mathbb{E}\left[\frac{\ell_t^v(a_t^v)^2}{p_t^v(a^v_t)^2} \right] \\
        &= \mathbb{E}\left[ \mathbb{E}\left[\frac{\ell_t^v(a_t^v)^2}{p_t^v(a^v_t)^2}  \mid \mathcal{F}_{t-1} \right] \right] \\
        &=\mathbb{E}\left[ \sum_{i\in \mathcal{A}} p^v_t(i) \frac{\ell_t^v(i)^2}{p_t^v(i)^2} \right]
\leq \mathbb{E}\left[ \sum_{i\in \mathcal{A}} \frac{1}{p_t^v(i)} \right] \leq \frac{K^2}{\gamma_t}
    \end{split}
\end{equation*}
where the last inequality comes from $p_t^v(i) \geq \gamma_t / K$.
Again, from Jensen's inequality, it follows
$$
\mathbb{E}\left[\|f_t\|_\ast^2 \right] \leq \frac{1}{N}\sum_{v\in \mathcal{V}} \mathbb{E}\left[\|g_t^v\|_\ast^2 \right] \leq \frac{K^2}{\gamma_t}.
$$
\end{proof}

Before presenting the next lemma, we recall the definition of strongly-convex functions and Fenchel duality. A function $\phi$ is said to be $\alpha$-strongly convex function on a convex set $\mathcal{X}$ if 
$$
\phi(x^\prime) \geq \phi(x) + \langle \nabla\phi(x), x^\prime- x \rangle + \frac{1}{2} \alpha\|x^\prime- x\|^2
$$
for all $x^\prime, x\in \mathcal{X}$, for some $\alpha \geq 0$.

Let $\phi^\ast$ denote the \emph{Fenchel conjugate} of $\phi$, i.e.,
$$
\phi^{\ast}(y)=\max_{x \in \mathcal{X}}\{\langle x, y\rangle-\phi(x)\}
$$
with the projection, 
$$
\nabla \phi^{\ast}(y)=\arg\max_{x \in \mathcal{X}} \{\langle x, y\rangle-\phi(x)\}.
$$

\begin{lemma}
\label{lm:lipschitz}
Let $\psi$ the normalized negative entropy function \cite{lattimore_szepesvari_2020} on $\mathcal{P}_{K-1} = \{x\in [0,1]^K: \sum_{i=1}^K x(i) = 1\}$ 
% \mv{It may be better to provide this definition before the lemma.}
,
$$
\psi_\eta (x) = \frac{1}{\eta} \sum_{i=1}^k x(i) \left(\log(x(i)) - 1\right).
$$
For all $t=1,\dots, T$,
it holds that
$$
x^{v}_{t} = \underset{x\in \mathcal{P}_{K-1}}{\operatorname{argmin}}\{\langle x, z^{v}_t \rangle + \psi_{\eta_{t}}(x)\} = \nabla \psi_{\eta_{t-1}}^\ast(-z^{v}_{t})
$$
with $\mathcal{X} = \mathcal{P}_{K-1}$ and 
$$
y_t = \underset{x\in \mathcal{P}_{K-1}}{\operatorname{argmin}}\{\langle x, \bar{z}_t \rangle + \psi_{\eta_{t-1}}(x)\}  = \nabla \psi_{\eta_{t-1}}^\ast(-\bar{z}_t).
$$
Furthermore, it holds
$$
\|x^{v}_t - y_t\| \leq \eta_{t-1} \|z^{v}_t - \bar{z}_t\|_\ast.
$$
\end{lemma}

\begin{proof}
We prove for $y_t = \underset{x\in \mathcal{P}_{K-1}}{\operatorname{argmin}}\{\langle x, \bar{z}_t \rangle + \psi_{\eta_{t-1}}(x)\}$ whose argument also applies to $x^v_t$.

Notice it suffices to consider the minimization problem
\begin{eqnarray}
    \min_{x\in  \mathcal{P}_{K-1}} & \eta_{t-1} \sum_{k=1}^K x(i) \Bar{z}_t(i) +  \sum_{i=1}^k x(i) \log(x(i)) \nonumber \\
    \text{subject to} & \sum_{k=1}^K x(i) = 1. \nonumber
\end{eqnarray}
It suffices to consider the Lagrangian, 
$$
\mathcal{L} = - \eta_{t-1} \sum_{k=1}^K x(i) \Bar{z}_t(i) -  \sum_{i=1}^k x(i) \log(x(i))  - \lambda \left(\sum_{k=1}^K x(i) - 1\right). 
$$
Consider the first-order conditions for all $i=1,\dots, K$
\begin{eqnarray}
    \frac{\partial \mathcal{L}}{\partial x(i)} &=& -\eta_{t-1} \Bar{z}_t(i) - \log(x(i)) - 1 - \lambda = 0 \nonumber
\end{eqnarray}
which gives $x(i) = \exp\left(-\eta_{t-1}\Bar{z}_t(i)\right) / \exp\left(1+\lambda\right)$ for all $i=1,\dots, K$.
Plugging into the constraint $\sum_{k=1}^K x(i) = 1$ together with the definition of Fenchel duality \cite{hiriart-urruty_convex_2010} completes the proof for $y_t$.

Note that the normalized negative entropy $\psi(x)$ is $1$-strongly convex,
$$
\psi(x^\prime) \geq \psi(x) + \langle \nabla\psi(x), x^\prime- x \rangle + \frac{1}{2}\|x^\prime- x\|^2.
$$
Multiplying $1/\eta_{t-1}$ both sides of the inequality yields that $\psi_{\eta_{t-1}}(x)$ is $1/\eta_{t-1}$-strongly convex. 
% \mv{The concept of "modulus" does not appear to be defined earlier. Should define somewhere $\alpha$-strong convexity for parameter $\alpha$.}
By Theorem~4.2.1 in \cite{hiriart-urruty_convex_2010}, we have that $\nabla \psi^\ast_{\eta_{t-1}}(z)$ is $\eta_{t-1}$-Lipschitz. 

It follows that
$$
\|p^{v}_t - \bar{p}_t\| = \|\nabla \psi_{\eta}^\ast(-z^{v}_t) - \nabla \psi_{\eta}^\ast(-\bar{z}_t)\| \leq \eta_{t-1} \|\bar{z}_t - z^{v}_t\|_\ast .
$$
\end{proof}

We state an upper bound on the network disagreement on the cumulative loss estimators from \cite{duchi2011dual} and \cite{hosseini2013online}.
\begin{lemma}
\label{lm:gossip}
For any $v\in \mathcal{V}$ and $t=1,2,\dots,T$, $$
\|\bar{z}_t - z^{v}_t\|_\ast
\leq \frac{K}{\gamma_T}\left( \frac{\min\{2\log T + \log n, \sqrt{n}\}}{1-\sigma_2(W)} +3\right) = \frac{K}{\gamma_T} C_W
$$
where $\sigma_2(W)$ is the second largest singular value of $W$.
\end{lemma}
\begin{proof}
    From Lemma~\ref{lm:z-bar}, it follows that $\|g^v_t\|_\ast \leq K/\gamma_t$. Since $\{\gamma_{t}\}$ is non-increasing, let $L = K/\gamma_T$ in Eq.~(29) in \cite{duchi2011dual} and Lemma~6 in \cite{hosseini2013online} completes the proof.
\end{proof}

\subsubsection{Proof of Theorem~\ref{thm:upper-bound:static}} \label{proof:upper-bound}

Let $i^\ast = \arg\min_{i\in \mathcal{A}} \sum_{t=1}^T \Bar{\ell}_t(i)$.
Note that $p^v_t$ is determined by $\mathcal{F}_{t-1}$ and $\mathbb{E}\left[f_t \mid \mathcal{F}_{t-1}\right] = \Bar{\ell}_t$ from Lemma~\ref{lm:loss-estimator}. It follows that for each agent $v\in \mathcal{V}$
\begin{eqnarray} \label{eq:jensen}
    R^v_T
&=& \mathbb{E}\left[\sum_{t=1}^T \langle \Bar{\ell}_t, p^v_t \rangle -  \sum_{t=1}^T \Bar{\ell}_t(i^\ast ) \right] \nonumber \\
&=&  \mathbb{E}\left[
\sum_{t=1}^T \langle \mathbb{E}\left[ f_t \mid \mathcal{F}_{t-1} \right], p^v_t \rangle - \sum_{t=1}^T  \mathbb{E}\left[ f_t (i^\ast) \mid \mathcal{F}_{t-1} \right] \right]\nonumber \\
&= & 
\mathbb{E}\left[\sum_{t=1}^T \langle  f_t  , p^v_t \rangle  -  \sum_{t=1}^T   f_t (i^\ast) \right]. \nonumber \\
\end{eqnarray}
By the definition of $p^v_t$, it follows
\begin{eqnarray}
    R^v_T
    &=& \mathbb{E}\left[\sum_{t=1}^T \left( \langle f_t, (1-\gamma) x^v_t + \gamma x^v_1 \rangle - f_t (i^\ast)  \right) \right] \nonumber \\
    &=& \mathbb{E}\left[\sum_{t=1}^T (1-\gamma_t) \left(\langle f_t, x^v_t \rangle - f_t (i^\ast) \right)\right] + \sum_{t=1}^T \gamma_t  \mathbb{E}\left[ \left( \langle f_t, x^v_1 \rangle - f_t (i^\ast) \right) \right] \nonumber \\
    &=&  \mathbb{E}\left[\sum_{t=1}^T (1-\gamma_t) \left(\langle f_t, x^v_t \rangle - f_t (i^\ast) \right)\right] + \sum_{t=1}^T \gamma_t  \left( \langle \Bar{\ell}_t, x^v_1 \rangle - \Bar{\ell}_t (i^\ast) \right) \nonumber \\
    &\leq& \mathbb{E}\left[\sum_{t=1}^T \left(\langle f_t, x^v_t \rangle - f_t (i^\ast) \right)\right] + \sum_{t=1}^T \gamma_t \nonumber \\
    &=& 
    \underbrace{\mathbb{E}\left[\sum_{t=1}^T \left(\langle f_t, y^v_t \rangle - f_t (i^\ast) \right)\right]}_{(\rom{1})} + 
    \underbrace{\mathbb{E}\left[\sum_{t=1}^T \langle f_t, x^v_t - y_t^v \rangle\right]}_{(\rom{2})} + \sum_{t=1}^T \gamma_t \nonumber
\end{eqnarray}
where the first inequality comes from the fact that $\gamma_t>0$ and the fact that $\|\Bar{\ell}_t\|_\ast \leq \sum_{v\in \mathcal{V}} 1/N \|\Bar{\ell}_t^v\|_\ast \leq 1$.

From Lemma~\ref{lm:z-bar}, it follows
$\Bar{z}_t = \sum_{s=1}^{t-1} f_s$.
Hence, it follows from Lemma~\ref{lm:lipschitz} that
$$
y_t = \underset{x \in \mathcal{P}_{K-1}}{\arg \min }\left\{\sum_{s=1}^t\langle f_s, x\rangle+\frac{1}{\eta_{t-1}} \psi(x)\right\}.
$$
From Lemma 3 in \cite{duchi2011dual} and  Corollary~28.8 in \cite{lattimore_szepesvari_2020}, we have
\begin{equation}
    \label{eq:rom-1}
    (\rom{1}) \leq \frac{1}{2} \sum_{t=1}^T \eta_{t-1} \mathbb{E}\left[ \|f_t\|_*^2 \right]+\frac{1}{\eta_{T}} \log(K)
\end{equation}
which is because $\{\eta_t\}$ is a non-increasing sequence.
Note that 
$$
\|x^{v}_t - y_t\| \leq \eta_{t-1} \|z^{v}_t - \bar{z}_t\|_\ast
$$
by Lemma~\ref{lm:lipschitz}. This yields that
\begin{equation}
\label{eq:rom-2}
    (\rom{2}) \leq \sum_{t=1}^T \eta_{t-1} \mathbb{E}\left[ \|f_t\|_\ast   \|z^{v}_t - \bar{z}_t\|_\ast \right].
\end{equation}
Plugging Equations~(\ref{eq:rom-1}) and (\ref{eq:rom-2}) into (\rom{1}) yields that
\begin{eqnarray}
    \label{eq:before-tuning}
    R^v_T  
    &\leq&
    \frac{1}{2} \sum_{t=1}^T \eta_{t-1} \mathbb{E}\left[ \|f_t\|_*^2 \right]+\frac{1}{\eta_{T}} \log(K) + \sum_{t=1}^T \eta_{t-1} \mathbb{E}\left[ \|f_t\|_\ast   \|z^{v}_t - \bar{z}_t\|_\ast \right] + \sum_{t=1}^T \gamma_t  \nonumber \\
    &\leq& \frac{1}{2} \sum_{t=1}^T \eta_{t-1} \mathbb{E}\left[ \|f_t\|_*^2 \right] + \frac{K}{\gamma_T} C_W \sum_{t=1}^T \eta_{t-1} \mathbb{E}\left[ \|f_t\|_\ast \right] + \sum_{t=1}^T \gamma_t +\frac{1}{\eta_{T}} \log(K) \nonumber \\
    &\leq&
    \frac{K^2}{2} \sum_{t=1}^T \frac{\eta_{t-1}}{\gamma_t}+ \frac{K^2}{\gamma_T} C_W \sum_{t=1}^T \eta_{t-1}  + \sum_{t=1}^T \gamma_t +\frac{1}{\eta_{T}} \log(K) \nonumber 
\end{eqnarray}
where the second inequality comes from Lemma~\ref{lm:gossip} and the third inequality comes from Lemma~\ref{lm:loss-estimator}.

Let 
$$
\gamma_t = \sqrt[3]{\frac{\left(C_W + \frac{1}{2}\right)K^2\log K}{t}}
\quad
\text{and}
\quad
\eta_t = \frac{\log K}{T \gamma_T} = \sqrt[3]{\frac{(\log K)^2}{ \left(C_W + \frac{1}{2}\right) K^2 T^2}}.
$$
Then, for every $v\in \mathcal{V}$, we have
\begin{eqnarray}
    R^v_T &\leq& \frac{3}{8} \sqrt[3]{\frac{K^2\log K}{\left(C_W + \frac{1}{2}\right)^2}} T^{\frac{2}{3}} + \sqrt[3]{K^2\log K \frac{C_W^3}{\left(C_W + \frac{1}{2}\right) ^2}} T^{\frac{2}{3}} \\ &+& \frac{3}{2} \sqrt[3]{\left(C_W + \frac{1}{2}\right)K^2\log K} T^{\frac{2}{3}} + \sqrt[3]{\left(C_W + \frac{1}{2}\right)K^2\log K }  T^{\frac{2}{3}} \nonumber \\
    &\leq & \frac{3}{4}  \sqrt[3]{K^2\log K} T^{\frac{2}{3}} + \sqrt[3]{C_WK^2\log K} T^{\frac{2}{3}} + \frac{5\sqrt[3]{2}}{2} \sqrt[3]{C_WK^2\log K} T^{\frac{2}{3}}  \nonumber \\
    &\leq& 5\sqrt[3]{C_W K^2\log K} T^{\frac{2}{3}}. \nonumber
\end{eqnarray}

%%%%%%%%%%%%%%%%%%%%%%%%%%%%%%%%%%%%%%%%%%%%%%%%%%%%%%%%%%%

\newpage \onehalfspacing
\setcounter{section}{5}
\setcounter{subsection}{0}

\section*{\textbf{\Large{Chapter~5}}}
\addcontentsline{toc}{section}{Chapter~5.  Communication-Efficient Federated Online Convex Optimization}

\bigskip
\bigskip

\textbf{\Large{Communication-Efficient Federated Online Convex Optimization}}

\bigskip
\bigskip
\bigskip

\noindent{In a cooperative learning system, multiple agents exchange information to make predictions and update their models, and the amount and frequency of communication can greatly impact the accuracy, speed, and scalability of the system. If the communication complexity is high, it can lead to increased communication overhead, which can result in slower and less efficient decision-making. This is especially important in real-world applications, where communication resources are limited and the volume of data generated is vast.
The challenge here is to design cooperative learning algorithms which can achieve the desired performance while minimizing the communication cost.}

\subsection{Introduction}
Large-scale federated learning systems have garnered significant research attention, due to their advantages in terms of privacy, security, and scalability \citep{kairouz2021advances}.
Prominent technology companies have implemented federated learning systems for various applications, such as learning from wearable devices, sentiment analysis, and location-based services \cite{bonawitz2019towards}.
However, a major challenge in the deployment of high-performance large-scale federated learning systems is the high communication overhead associated with the training process. Designing a communication-efficient federated learning method that balances the trade-off between communication costs and convergence rate remains a challenging task.

Removing the communication bottleneck for federated learning methods has been the focus of much research \citep{hamer2020fedboost, reisizadeh2020fedpaq, rothchild2020fetchsgd}. 
A number of quantization and sketching methods have been proposed to reduce the \emph{per-message communication costs}, that is, the coding length of each message transmitted within the federated learning systems. In this chapter, we aim to decrease the \emph{communication complexity}, which is measured as the number of messages transmitted during the training process. Prior work has relied on deterministic communication protocols, which facilitate elegant analysis of communication complexity. However, the system lacks the flexibility to balance the trade-off between communication complexity and convergence rate.

In this chapter, we investigate the problem of online convex optimization in the federated learning setting. The agents in the network can exchange information with their neighbors. At each time step, each agent makes a decision, receives feedback, and communicates with its neighbors. We consider a heterogeneous feedback scenario, where different agents receive feedback from different convex functions at each time step. The goal is to identify the best global solution retrospectively, with the lowest cumulative loss in terms of the averaged convex function over all agents in the network.

\subsubsection{Related work}

In this chapter, we examine the field of communication-efficient algorithms for Distributed Online Convex Optimization (DOCO) \citep{duchi2011dual, hosseini2013online, wan2020projection, cesa2020cooperative}. Our study builds upon the previous research in this area and provides a comprehensive review of the recent literature in this field. The objective of this review is to showcase the unique contributions of our work and differentiate it from the prior research.

The focus of communication-efficient algorithms for DOCO is to simultaneously minimize the amount of information transmitted between nodes in the distributed system and the cumulative regret over a given horizon. Previous works in this area have explored two main directions. The first direction is to minimize the coding length of the transmitted message, while the second direction is to reduce the rounds of communication between nodes.

One popular method to reduce the coding length of each message is the use of quantization methods, which perform lossy compression on the original information contained in the message. Simple quantization methods, such as 1BitSGD \citep{seide20141, strom2015scalable}, were initially proposed for the distributed training of deep neural networks with Stochastic Gradient Descent (SGD). Subsequently, more advanced methods, such as QSGD \citep{alistarh2017qsgd} and the random-dropping strategy \citep{wangni2018gradient}, have been developed with convergence guarantees. Quantization methods have also been studied in the context of federated learning \citep{hamer2020fedboost, reisizadeh2020fedpaq, rothchild2020fetchsgd}.

Reducing the rounds of communication, referred to as \emph{communication complexity}, has become a crucial area of research in communication-efficient DOCO in recent years \citep{zhang2017projection, wan2020projection, reisizadeh2020fedpaq}. \cite{zhang2017projection} proposed the first conditional gradient descent algorithm for the distributed online setting, which reduces the cost of expensive projection operations by using simpler linear optimization steps. Subsequently, \cite{wan2020projection} presented the D-OCG method, which divides the total rounds into $\sqrt{T}$ equal-sized blocks. \cite{reisizadeh2020fedpaq} proposed the periodic communication strategy to minimize the communication complexity.
Both methods enjoy a $O(\sqrt{T})$ communication complexity. It remains unknown before this thesis whether the communication complexity can be further reduced to $o(\sqrt{T})$.

In this chapter, we focus on reducing the communication complexity for distributed online convex optimization problems. Specifically, our algorithm builds on the Dual Average algorithm \citep{duchi2011dual, nesterov2009primal, hosseini2013online}. We propose a stochastic communication protocol instead of a deterministic communication protocol. By setting the communication probability as $T^{-\alpha}$ for any $\alpha\in[0, 1)$, we show that the resulting communication complexity scales as $\Theta(T^{1-\alpha})$ and the regret upper bound scales as $\Tilde{O}(\sqrt{T^{1+\alpha}})$ with high probability.
Therefore, our algorithm balances a trade-off between the communication complexity and the regret.

The minimization of communication complexity has also been discussed in the context of Distributed Offline Convex Optimization and Stochastic Multi-Agent Multi-Armed Bandits. Interested readers are referred \cite{smith2018cocoa} and  \cite{agarwal2022multi}  for a comprehensive review of these topics.

\subsubsection{Organization of this chapter}

In Section~\ref{sec::comm-complexity}, the problem setting is elucidated and the metric of communication complexity is rigorously defined. Subsequently, Section~\ref{sec::ceodda} presents a communication-efficient federated online convex optimization algorithm, equipped with both regret and communication complexity bounds. The validity of our results is demonstrated in Section~\ref{sec::trade-off} through a rigorous mathematical proof, which can be found in Section~\ref{app::comm-efficient}.

\subsection{Federated online convex optimization}
\label{sec::comm-complexity}

Consider a communication network defined by an undirected graph $\mathcal{G}=(\mathcal{V}, \mathcal{E})$, where $\mathcal{V}$ is the set of $N$ agents and $(u,v)\in \mathcal{E}$ if agent $u$ and agent $v$ can directly exchange messages.
We assume that $\mathcal{G}$ is simple, i.e. it contains no self loops nor multiple edges.
The agents in the communication network collaboratively aim to solve a distributed online convex optimization problem. In this problem, there is a convex decision set $\mathcal{X}\subset \mathbb{R}^K$ and a fixed time horizon $T$.

At each time step $t$, each agent $v$ will choose its action $x^v_t\in \mathcal{X}$, observe the feedback $g^v_t$.
At the end of each time step, each agent $v\in \mathcal{V}$ can communicate with their neighbors $\mathcal{N}(v)=\{u\in \mathcal{V}: (u, v)\in \mathcal{E}\}$. 

Let $f^v_t(\cdot)$ be a convex function defined on $\mathcal{X}$ for agent $v$ at time step $t$, and $f_t$ be the network loss function defined as
$$
f_t(x) = \frac{1}{N}\sum_{v\in\mathcal{V}} f_t^v(x) \quad \text{for } x\in\mathcal{X}.
$$
Each agent $v$ receives a loss of $f_t(x^v_t)$ at the time step $t$.

The performance of each agent $v\in \mathcal{V}$ is measured by its \emph{regret}, defined as the difference of the expected cumulative loss incurred and the cumulative loss of a \emph{globally} best fixed solution in hindsight, i.e.
$$
R^v_t = \sum_{t=1}^v f_t(x^v_t) - \min_{x\in\mathcal{X}} \sum_{t=1}^T f_t(x).
$$ 
Here we add two assumptions on the problem we consider, which are standard in online convex optimization:
\begin{hyp}[Gradient-based optimization]
    $g^v_t\in \partial f^v_t(x^v_t)$
\end{hyp}

\begin{hyp}[Lipschitz-continuity]
    There exists some $L>0$ such that $\|g^v_t\|_\ast\leq L$ for all $v\in\mathcal{V}$ and $t=1,2,\dots, T$.
\end{hyp}

Except for the regret of each agent, we also want to minimize the \emph{communication complexity} of the system, defined as
$$
Q_T = \sum_{t=1}^T m_t
$$
where $m_t$ is the number of messages transmitted in the communication network at time step $t$.

The ODDA algorithm proposed by \cite{hosseini2013online, duchi2011dual}, which is designed to minimize the regret, enjoys a regret upper bound of $O(T^{1/2})$ with communication complexity of $\Theta(T)$.
We show in the next section a stochastic algorithm which has sub-linear regret upper bounds and communication complexity in high probability simultaneously.

\subsection{{\pfedexp}: a communication-efficient stochastic regret-minimization algorithm}
\label{sec::ceodda}

\begin{algorithm}[h]
\caption{\pfedexp}
\label{algo:ceodda}
\SetKwInOut{Init}{Initialization}
\SetKwInOut{Input}{Input}
\Input{Non-increasing sequence of learning rates $\{\eta_t>0\}$, a regularizer $\psi$ and a skipping parameter $\alpha\in [0, 1)$.} 
\Init{$z_1^v = 0$ and $ x^v_1 = \max_{x\in\mathcal{X}} \psi(x)$ for all $v\in\mathcal{V}$.}
\For{each time step $t=1,2,\dots, T$}{
    sample $X_t\sim \operatorname{Bernoulli}(T^{-\alpha})$\;
    \If{$X_t=1$}{
        uniformly sample an edge $(u, u^\prime)\in\mathcal{E}$\;
    }
    \For{each agent $v\in \mathcal{V}$}{
        select $x^v_t$\;
        receive feedback $g^v_t$\;
        update the accumulative loss \tcp*{communication step}
        $$
        z_{t+1}^v = 
        \begin{cases}
        \frac{1}{2}\left(z_t^{u}+z_t^{u^\prime}\right)  + g_t^v, & v\in \{u, u^\prime\} \quad \text{and} \quad X_t = 1 \\
        z_t^v  + g_t^v, & \text{otherwise} \;
        \end{cases}
        $$
        update $x^v_{t+1} = \Pi_{\mathcal{X}}^\psi \left( z^v_{t+1}, \eta_t \right)$\;
    }
}
\end{algorithm}

In this section, we present a novel communication-efficient stochastic algorithm, named {\pfedexp}. The proposed algorithm enables the communication complexity to grow sub-linearly with the horizon $T$, while ensuring that the cumulative regret of each agent remains sub-linear in the same horizon. This innovative approach offers a compelling solution to the challenge of limiting communication while maintaining the performance of the system.

The {\pfedexp} algorithm is parameterized by $\alpha$, called \emph{skipping parameter}, which lies in the interval [0, 1). The selection of $\alpha$ enables the {\pfedexp} algorithm to strike a balance between the cumulative regret of each agent and the communication complexity of the system.

{\pfedexp} takes a \emph{regularizer} function $\psi$ as an input argument.
The regularizer $\psi$ is a strongly convex function defined on $\mathcal{X}$ with $\max_{x\in \mathcal{X}} \psi(x) <\infty$. 
And the decision variable chosen by agent $v$ is computed by solving the minimization algorithm
$$
x^v_t = \Pi_\mathcal{X}^\psi(z^v_t, \eta_{t-1})=\arg \min _{x \in \mathcal{X}}\left\{\langle z^v_t, x\rangle+\frac{1}{\eta_{t-1}} \psi(x)\right\}.
$$

In the communication step of {\pfedexp}, each agent has the opportunity to either be matched with a single neighbor to communicate with, or skip the communication step.
Specifically, at the start of time step $t$, the {\pfedexp} will decide whether this time step is a communication step or skipping step. 
When it is a communication step, the system will select one edge from the network at random, and the agents in the edge will average their estimates before integrating the new observation $g_t^v$.
In a skipping step, the agents will not communicate with others.

The probability of a communication step is inversely proportional to a polynomial function in the horizon $T$.
Consequently, the communication complexity scales sub-linearly with $T$ with high probability.
The details of {\pfedexp} are described in Algorithm~\ref{algo:ceodda}.

\subsection{Trade-off between communication complexity and regret}
\label{sec::trade-off}

In this section, a theoretical examination of the cumulative regret and communication complexity of each agent in the {\pfedexp} algorithm is performed. The objective is to derive upper bounds for these metrics.

Recall that $\delta_{\min} = \min\{\delta_v: v\in\mathcal{V}\}$.
Let $D = \operatorname{diag}\{\delta_v: v\in\mathcal{V}\}$, A be the adjacent matrix of $\mathcal{G}$ and $M$ be its Laplacian matrix.
\begin{theorem}
\label{thm:upper-bound:comm-regret}
Let $R>0$ such that $\psi(x^\ast) \leq R^2$.
Assume that the network runs Algorithm~\ref{algo:ceodda} with $\alpha\in [0, 1)$ and
$$
\eta_t = \frac{R\sqrt{1-\lambda_2(W)}}{L\sqrt{t}}.
$$
Then, for any $\delta\in (0, 1)$, with probability at least
$$
1-\frac{1}{T} - 2\exp\left(-\frac{\delta^2}{3}T^{1-\alpha}\right)
$$
the communication complexity
$$  Q_T \in \left[2(1-\delta) T^{1-\alpha}, \quad 2(1+\delta) T^{1-\alpha} \right]$$
and
the cumulative regret of each agent $v\in \mathcal{V}$ is bounded as
$$
R^v_T \leq \Tilde{O}\left( \sqrt{\frac{|\mathcal{E}|}{\delta_{\min}\lambda_{N-1}(M) }} \sqrt{T^{1+\alpha}} \right)
$$
where $\lambda_{N-1}(M)$ is the second smallest eigenvalue value of the Laplacian matrix.
\end{theorem}

\paragraph{Proof sketch}
The proof of Theorem~\ref{thm:upper-bound:comm-regret} relies on a decomposition used in the proof of Theorem~\ref{thm:upper-bound:static}, replacing the expectation argument with a high probability argument.

Let 
$\bar{z}_t=\frac{1}{N} \sum_{v\in\mathcal{V}} z^v_t$ and $y_{t+1}=\Pi_\mathcal{X}^\psi(\bar{z}_{t+1}, \eta_t)$.
The cumulative regret of agent $v$ can be bounded by the sum of two terms
\begin{equation*}
    \begin{split}
    R^v_T &\leq 
    \underbrace{\sum_{t=1}^T \frac{1}{N} \sum_{v\in\mathcal{V}} \left\langle g^v_t, y_t-x^*\right\rangle}_{\operatorname{FTRL}} + 
      \underbrace{\sum_{t=1}^T\left( \frac{2L}{N} \sum_{v\in\mathcal{V}}\left\|x^v_t-y_t\right\|+L\left\|x^v_t-y_t\right\| \right)}_{\operatorname{CONSENSUS}}.
    \end{split}
\end{equation*}

With the probabilistic peer-to-peer communication mechanism in Algorithm~\ref{algo:ceodda}, the CONSENSUS term is upper bounded by $\Tilde{O}(1/ (1-\lambda_2(W)))$ where
$$
W = I - \frac{1}{2T^\alpha |\mathcal{E}|} (D-A).
$$
Note that $D^{1/2} M D^{1/2} = D-A$, and from Lemma~4 in \cite{duchi2011dual}, 
$$
\lambda_2(W) \leq 1 - \frac{\delta_{\min}}{|\mathcal{E}|T^\alpha} \lambda_{N-1}(M).
$$
Choosing the adequate learning rate $\eta_t$ completes the proof. 
The details of the proof is presented in Section~\ref{app::comm-efficient}.

\paragraph{Communication complexity and regret trade-off}

Note that both the communication complexity and regret of {\pfedexp} scale sub-linearly with respect to the horizon.
As the skipping parameter $\alpha$ approaches 1, the number of communication steps decreases. As a result, with high probability, the communication complexity decreases as $T^{1-\alpha}$ and the cumulative regret upper bound increases as $\sqrt{T^{(1+\alpha)}}$ (ignoring logarithmic terms).
Note that the case when $\alpha = 1/2$ recovers the communication complexity of $O(\sqrt{T})$ and the regret upper bound of $O(T^{3/4})$ presented in the analysis of online conditional gradient algorithms in the distributed setting \citep{wan2020projection}.

\subsection{Proofs}
\label{app::comm-efficient}

\subsubsection{Proof of Theorem~\ref{thm:upper-bound:comm-regret}}

\paragraph{Communication complexity}
Note that at each time step, with probability $T^{-\alpha}$, there are two agents exchanging their messages; otherwise there is no message exchanged.
Hence the cumulative number of messages exchanged
$$
Q_T = \sum_{t=1}^T m_t = 2 \sum_{t=1}^T X_t.
$$
Note that $\{X_t: t=1,2,\dots, T\}$ are i.i.d. Bernoulli random variables with mean $T^{-\alpha}$.
From the Chernoff bound for binomial distribution \citep{mitzenmacher2017probability}, we have for any $\delta\in (0, 1)$
$$
\mathbb{P}(Q_T \geq 2(1+\delta) T^{1-\alpha}) \leq \exp \left(-\frac{\delta^2 T^{1-\alpha}}{3}\right)
$$
and
$$
\mathbb{P}(Q_T \leq 2 (1-\delta)  T^{1-\alpha}) \leq \exp \left(-\frac{\delta^2 T^{1-\alpha}}{2}\right).
$$

\paragraph{Cumulative regret}
The proof follows the general approach in \cite{duchi2011dual, hosseini2013online} adapted to the stochastic gossip matrix setting.
It follows that , it follows that 
\begin{equation*}
    \begin{split}
        R^v_T
        &\leq \sum_{t=1}^T \left(f_t(y_t)-f_t\left(x^*\right)+L\left\|x^v_t-y_t\right\|\right) \\
        & \leq \sum_{t=1}^T\left(\frac{1}{N} \sum_{v\in\mathcal{V}}\underbrace{\left \langle g^v_t, x^v_t-x^*\right\rangle}_{(\ast)} 
 + \frac{L}{N} \sum_{v\in\mathcal{V}}\left\|x^v_t-y_t\right\|+L\left\|x^v_t-y_t\right\|\right) 
    \end{split} 
\end{equation*}
where the first inequality comes from the fact that $f_t^v$ is $L$-Lipschitz continuous, the second inequality comes from the fact that $f^v_t$ is convex function.

For term ($\ast$), we can expand as 
\begin{equation*}
    \begin{split}
\left \langle g^v_t, x^v_t-x^*\right\rangle
&=\left\langle g^v_t, y_t-x^*\right\rangle + \left\langle g^v_t, x^v_t-y_t\right\rangle \\
&\leq \left\langle g^v_t, y_t-x^*\right\rangle + L \|x^v_t-y_t\|
    \end{split}
\end{equation*}
the inequality comes from the fact that $\|g^v_t\|_\ast\leq L$.

It follows that
\begin{equation}
    \label{eq-fedoco-decomp}
    R^v_T \leq \underbrace{\sum_{t=1}^T \frac{1}{N} \sum_{v\in\mathcal{V}} \left\langle g^v_t, y_t-x^*\right\rangle}_{(\rom{1})} + \underbrace{\sum_{t=1}^T\left( \frac{2L}{N} \sum_{v\in\mathcal{V}}\left\|x^v_t-y_t\right\|+L\left\|x^v_t-y_t\right\| \right)}_{(\rom{2})}
\end{equation}

Note that
$\Bar{z}_t = \sum_{s=1}^{t-1} \frac{1}{N} \sum_{v\in\mathcal{V}} g^v_s$.
Hence, it follows from Lemma~\ref{lm:lipschitz} that
$$
y_t = \underset{x \in \mathcal{X}}{\arg \min }\left\{\sum_{s=1}^t\langle \frac{1}{N} \sum_{v\in\mathcal{V}} g^v_s, x\rangle+\frac{1}{\eta_{t-1}} \psi(x)\right\}.
$$
From Lemma 3 in \cite{duchi2011dual} and  Corollary~28.8 in \cite{lattimore_szepesvari_2020}, we have
\begin{equation}
    \label{eq:fedoco-rom-1}
    (\rom{1}) \leq \frac{L^2}{2} \sum_{t=1}^T \eta_{t-1}  +\frac{1}{\eta_{T}} \psi(x^\ast)
\end{equation}
which is because $\{\eta_t\}$ is a non-increasing sequence.
Note that 
$$
\|x^{v}_t - y_t\| \leq \eta_{t-1} \|z^{v}_t - \bar{z}_t\|_\ast
$$
by Lemma~\ref{lm:lipschitz}. This yields that
\begin{equation}
\label{eq:fed-oco-rom-2}
    (\rom{2}) \leq L \sum_{t=1}^T \eta_t\left(\left\|\bar{z}_t-z^v_t\right\|_*+\frac{2}{N} \sum_{v\in\mathcal{V}}\left\|\bar{z}_t-z^v_t\right\|_*\right).
\end{equation}

Plugging Equations~(\ref{eq:fedoco-rom-1}) and (\ref{eq:fed-oco-rom-2}) into Equation~(\ref{eq-fedoco-decomp}) yields that
\begin{eqnarray}
    \label{eq:fedoco-before-tuning}
    R^v_T  
    &\leq&
    \frac{L^2}{2} \sum_{t=1}^T \eta_{t-1}  +\frac{1}{\eta_{T}} \psi(x^\ast) + L \sum_{t=1}^T \eta_t\left(\left\|\bar{z}_t-z^v_t\right\|_*+\frac{2}{N} \sum_{v\in\mathcal{V}}\left\|\bar{z}_t-z^v_t\right\|_*\right)  . \nonumber 
\end{eqnarray}

Let $e_v\in \{0, 1\}^K$ be the base vector whose coordinates are all 0 except that the $v$-th
coordinate is $1$, and $(u_t, u^\prime_t)$ be the edge selected at time step $t$ when $X_t = 1$.
Define the gossip matrix at time step $t$ 
$$
W_t = I - \frac{1}{2} (e_{u_t} - e_{u^\prime_t}) (e_{u_t} - e_{u^\prime_t})^\top \mathbb{I}\{ X_t = 1 \}
$$
then
$$
z_{t+1}^v = \sum_{u: (u,v)\in \mathcal{E}} [W_t]_{u,v} z_t^u + \ell_t^v
$$
and
$$
\mathbb{E}\left[W_t\right] = W = I - \frac{1}{T^\alpha |\mathcal{E}|} (D-A)
$$
which comes from the fact that $\sum_{u: (u,v)\in \mathcal{E}} (e_{u_t} - e_{u^\prime_t}) (e_{u_t} - e_{u^\prime_t})^\top = 2(D-A)$.

From Theorem~3 in \cite{duchi2011dual}, we have
$$
\mathbb{P}\left[\max _{t \leq T, v\in\mathcal{V}}\left\|\bar{z}_t-z^v_t\right\|_* / L>\frac{6 \log \left(T^2 N\right)}{1-\lambda_2(W)}+\frac{1}{T \sqrt{N}}+2 \right] \leq \frac{1}{T}.
$$
Let 
$$
\eta_t = \frac{R\sqrt{1-\lambda_2(W)}}{L\sqrt{t}}.
$$
Then,  with probability at least $1-1/T$, for every $v\in \mathcal{V}$,
$$
R^v_T \leq c RL \cdot \frac{\log(T\sqrt{N})}{\sqrt{1-\lambda_2(W)}} \sqrt{T}
$$
for some universal constant $c>0$.

Note that $D-A = D^{1 / 2} \mathcal{L} D^{1 / 2}$, hence
$$
\lambda_2\left(W\right)
= \lambda_2\left(I - \frac{1}{T^\alpha |\mathcal{E}|} D^{1 / 2} \mathcal{L} D^{1 / 2}\right)
\leq 1 - \frac{\delta_{\min}}{T^\alpha |\mathcal{E}|} \lambda_{N-1}(M).
$$
It follows that with probability at least $1-1/T$, for every agent $v$,
$$
R^v_T \leq c RL \cdot \log(T\sqrt{N}) \sqrt{\frac{|\mathcal{E}|}{\delta_{\min}\lambda_{N-1}(M)}} \sqrt{T^{1+\alpha}}
$$
for some universal constant $c>0$.

Applying the union bound on $Q_T$ and $R^v_T$ completes the proof.
%%%%%%%%%%%%%%%%%%%%%%%%%%%%%%%%%%%%%%%%%%%%%%%%%%%%%%%%%%%%%%%%%%%%%%%%%%%%%%%
%%%%%%%%%%%%%%%%%%%%%%%%%%%%%%%%%%%%%%%%%%%%%%%%%%%%%%%%%%%%%%%%%%%%%%%%%%%%%%%

%%%%%%%%%%%%%%%%%%%%%%%%%%%%%%%%%%%%%

\newpage \onehalfspacing
\setcounter{section}{6}
\setcounter{subsection}{0}

\section*{\textbf{\Large{Chapter~6}}}
\addcontentsline{toc}{section}{Chapter~6.  Greedy Bayes Incremental Matching Strategies}

\bigskip
\bigskip

\textbf{\Large{Greedy Bayes Incremental Matching Strategies}}

\bigskip
\bigskip
\bigskip

\noindent{Efficiently finding a maximum value matching for a given set of nodes based on observations of values of matches over a sequence of rounds is a fundamental sequential learning problem. Here a matching refers to a set of matches of node pairs (or matches in short) such that each node is matched to at most one other node. This problem arises in many applications including team formation, online social networks, online gaming, online labor, and other online platforms. We may think of the nodes as agents in MACL systems and the value of a matched node pair to represent the reward or welfare generated from interaction or joint production of the two agents, contributing to the total reward or welfare. For example, in team production, the value of a matched pair of nodes may represent the quality of the work produced. In online social networks, the value of a matched pair of nodes may represent user satisfaction or social capital gained through interaction. 
}

\subsection{Introduction}

We consider the case when the value of each matched pair of nodes is according to a function -- we refer to as a value function -- of latent node features in some domain. We focus on the case of a finite domain such that each node is either of low or high type, which can be represented by each node having value $0$ or $1$ -- we refer to this as latent binary types. For value functions, we consider two Boolean functions, AND and OR. The types of nodes are a priori unknown and partial information about node types is acquired as values of matches are observed. The underlying value function plays a key role in how much information is revealed about individual node types from observed values of matches. For example, for Boolean OR function, having observed value of a match to be $0$, we can immediately deduce that both nodes are of low type, and, otherwise, the only conclusion that can be drawn is that at least one node is of high type. The decision maker gains partial information about individual node types as outcomes of matches are observed over a sequence of rounds. We consider this decision problem under the constraint that sequence of matchings must be incremental, meaning that in each round the selected matching is derived by a limited change of the matching in the previous round. Specifically, we consider the case when in each round the decision maker can only choose a matching that is identical to the matching selected in the previous round, except for possibly re-matching any two previously matched pairs. Learning how to match through an incremental sequence of matchings limits how much information can be gained about individual node types in each round. This constraint is motivated by applications in practice when matched node pairs remain matched over several rounds, allowing only a small change of matching to be made in each round. 

We consider the underlying incremental matching problem for a set of $n$ nodes in a Bayesian setting where latent binary types of nodes are according to a prior product-form Bernoulli distribution with parameter $p$. Here $p$ is the probability that a node is of high type according to the prior distribution. We study greedy Bayes incremental matching strategies that in each round select a matching that has highest marginal value from the set of feasible matchings (obeying incremental matching constraint), according to posterior distribution, given the information observed up to the current round. We evaluate performance of greedy Bayes incremental matching strategies by considering expected Bayesian regret, defined as the difference between cumulative expected value of an optimal matching and expected value of the sequence of matchings chosen by the algorithm asymptotically for large number of rounds. Our goals are to identify greedy Bayes incremental matching strategies for for both AND and OR value functions, and characterize expected Bayesian regret under these two functions.

\subsubsection{Related work}

Team formation problems were studied under various assumptions in different scientific fields and application domains. \cite{KM18}, \cite{SVY20}, \cite{ts-dabeen} and \cite{notes} studied individual skill scoring methods for the best team selection, or assigning individuals to teams, for various functions determining team performance from individual performances. \cite{NWN99} studied the relationship between work team effectiveness and two aspects of the personality traits of team members, namely, the average level of a given trait within a team, and the variability in traits within a team. \cite{CE10} studied the problem of formations of teams through a pricing mechanism, where individuals are partitioned into groups of equal size and each team requires exactly one member from each group. When there are two groups, the problem corresponds to the classic marriage problem. \cite{liemhetcharat} studied the team formation problem where each individual has its own capabilities and the performance of a team depends on capabilities of team members as well their pairwise synergies. \cite{RMKT17} studied the problem of partitioning individuals into disjoint groups for different objectives that account for pairwise compatibility of individuals. \cite{FT20} studied the problem of inferring individual from team productivity using a scoring method leveraging varying membership and level of success of all teams. Popular skill rating systems used in online gaming platforms, such as TrueSkill, assume team performances to be sum of individual performances with some added noise, and matching strategies are used aiming to match players of similar skills in order to make interesting competitions \cite{GMH07}. 

The most closely related work to ours is \cite{johari18} which introduced the problem of incremental matching with match outcomes according to a function of binary latent types. They considered regret for maximum value and minimum value functions. For maximum value (Boolean OR) function, their 1-chain algorithm is a greedy Bayes incremental matching algorithm, for which we show a tight regret characterization. For minimum value (Boolean AND), they studied an algorithm and established some comparison results. Our work identifies the greedy Bayes incremental matching algorithm and establishes a regret bound. \cite{synergy} studied the same problem, but without the constraint that the sequence of matchings must be according to an incremental matching sequence.

The problem we consider has some connections with combinatorial bandits that have been studied under various assumptions, e.g. \cite{comb}, \cite{CTPL15}, \cite{KZAC15}, \cite{combgenreward}, \cite{combsumfull}, \cite{combstochdom}. For combinatorial bandits, in each round, the decision maker selects an arm that corresponds to a feasible subset of base arms. For example, given a set of $N$ base arms, an arm is an arbitrary subset of base arms of cardinality $K$. Under full bandit feedback, the reward of the selected arm is observed, which is according to an underlying function of the  corresponding base arms' values. For example, this may be sum of values \cite{comb,CTPL15,combsumfull}, or some non-linear function of values (such as maximum value) \cite{combgenreward,combstochdom}. Under semi-bandit feedback, individual values of all selected base arms are observed. In our setting, we may think of node pairs to be base arms, matchings to be arms, and we have semi-bandit feedback. Alternatively, we may regard nodes to be base arms, matchings to be arms, and we have full-bandit feedback for each match in a matching. Note that requiring for sequence of matching selections to be according to an incremental matching sequence is not studied in the line of work on combinatorial bandits. 

\cite{SKB16} studied the best-of-$K$ bandit problem in which at each time step, the decision maker chooses a subset of $K$ arms from a set of $N$ arms and observers reward equal to maximum of individual arm's values drawn from a joint distribution. Here the objective is to identify the subset that achieves the highest expected reward using as few queries as possible. 

Regret minimization for maximum weight matching problems with stochastic edge weights whose mean values obey a rank-1 structure was studied in \cite{flore}. Here the mean weight $w_{u,v}$ for two nodes $u$ and $v$ is of the form $w_{u,v} = \theta_u \theta_v$, where $\theta_1, \ldots, \theta_n$ are hidden node parameters with values in $[0,1]$. The stochastic weight of an edge is defined as the sum of mean edge weight and a sub-Gaussian random noise. An adaptive matching algorithm is shown to have regret of order $n\log(T)$ up to a poly-logarithmic factor. Note that the rank-1 assumption along with assuming that node parameters are binary and zero noise boil, down to weights according to Boolean OR function. Our work is different in that we consider noiseless weights, binary node types, and we admit the incremental matchings constraint.   

\subsubsection{Organization of this chapter}

In Section~\ref{sec:problem} we present problem formulation and notation used in this chapter. Section~\ref{sec:results} contain our results, starting with a regret lower bound, and then showing greedy Bayes incremental matching algorithm and its analysis for each value function studied separately in different subsections. Section~\ref{sec::greedy-bayes-proofs} contains proofs for theorems stated in this chapter. 

\subsection{Problem formulation}
\label{sec:problem}

Let $V = \{1,\ldots, n\}$ be a set of $n$ nodes, where $n$ is an even, positive integer. Each node $v\in V$ has a latent variable $\theta_v$ that takes value in $\Theta = \{0, 1\}$. We refer to the latent variable of a node as the \emph{type} of node. Type $0$ is referred to as \emph{low type} and type $1$ is referred to as  \emph{high type}.

The latent variables $\theta = (\theta_1, \ldots, \theta_n)$ are assumed to have a prior distribution of product-form with identical Bernoulli marginal distributions, i.e. $\P[\theta_v = 1] =1-\P[\theta_v = 0]= p$, for all $v\in V$, for some $p \in [0,1]$. Intuitively, parameter $p$ characterises the population of nodes with respect to their types--we may think of as the portion of high-type nodes in the population. 

Each match of a pair of nodes $(u,v)$, has \emph{weight} $w_{u,v} = \f(\theta_u, \theta_v)$ where $\f$ is a symmetric function $\f: \Theta^2 \rightarrow \tilde{\Theta}$. A function is said to be symmetric if its output is invariant to permutations of its inputs. Set $\tilde{\Theta}$ is assumed to be fixed---its elements do not depend on either $n$ or $p$. We refer to $\f$ as a \emph{value function}, quantifying the value of a pair of nodes. When $\tilde{\Theta} = \{0,1\}$, then $\f$ is a Boolean function. We consider two Boolean functions with dual input, which are defined as follows:
\begin{itemize}
    \item AND: $1$ if both inputs are $1$, otherwise $0$
    \item OR: $0$ if both inputs are $0$, otherwise $1$.
\end{itemize}
We refer to Boolean OR as \emph{maximum} value function, Boolean AND as \emph{minimum} value function. 

These value functions are natural choices and have been studied in the context of production systems. Boolean OR value function requires at least one node of a pair of nodes to be of high type for the value of the pair to be high. In other words, the value of a pair is the value of the highest type node of this pair. This function is sometimes referred to as \emph{strongest link} function. In the other extreme, Boolean AND value function requires both nodes of a pair of nodes to be of high type for the value of the pair to be high. This value function is sometimes referred to as the \emph{weakest link} function, indicating that the value of a pair of nodes is the value of the lowest type node of this pair. 

The value $w_{u,v} =\f(\theta_u,\theta_v)$ of a pair of nodes $(u,v)$ may only reveal a \emph{partial information} about values of node types $\theta_u$ and $\theta_v$. We discuss this for functions $\f$ that we consider as follows. For Boolean OR value function, if $w_{u,v} = 0$, then we can deduce that \emph{both} $u$ and $v$ are of low type, and, otherwise, if $w_{u,v} = 1$, then we can deduce that \emph{either} $u$ or $v$ is of high type.
For minimum (Boolean AND) value function, if $w_{u,v}=0$, then we can deduce that \emph{either} $u$ or $v$ is of low type, and , otherwise, if $w_{u,v}=1$, then we can deduce that \emph{both} $u$ and $v$ are of high type. 

\begin{figure}[t]
\begin{center}
\includegraphics[width=14cm]{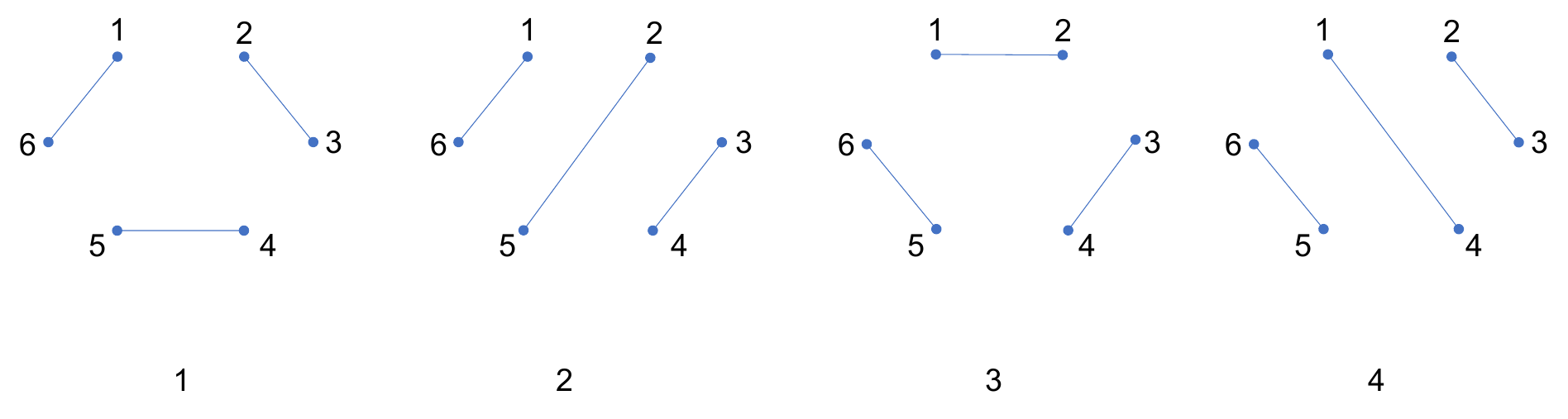}
\caption{An example of an incremental matching sequence: for example, the second matching is derived from the first matching by re-matching node pairs $(2,3)$ and $(4,5)$.}
\label{fig:inc-match}
\end{center}
\end{figure}

We consider the problem of finding a maximum value matching of nodes in a sequential decision making setting. An adaptive matching algorithm $\alg$ selects a matching $M_t^{\alg}$ in each round $t = 1, \ldots,T$ from a set of \emph{permissible matchings} $\mathcal{M}_t$, observes the value of each pair in this matching, i.e. $Y_t = \{\f(\theta_u, \theta_v): (u, v)\in M_t^{\alg}\}$, and receives an instantaneous reward $\mu_{\f}(M_t^{\alg}) = \sum_{(u,v)\in M_t^{\alg}} \f(\theta_u, \theta_v)$. The set of permissible matchings has to satisfy the \emph{incremental matching constraint}: each $M\in \mathcal{M}_t$ is such that $M$ has at most two new matched pairs with respect to $M_{t-1}^{\alg}$. Formally, $\mathcal{M}_t = \{M\in \mathcal{M}: |M\Delta M_{t-1}^{\alg}| \leq 4\}$ where $\mathcal{M}$ is the set of all matchings, and $\Delta$ is the symmetric set difference operator $A\Delta B = \left(A \backslash B\right) \cup \left(B \backslash A\right)$. The algorithm $\alg$ specifies a mapping from the history  $\mathcal{H}_{t}$ to $M_t^{\alg}$, where $\mathcal{H}_1 = \emptyset$ and $\mathcal{H}_{t} = \{M_1^{\alg}, Y_1, \dots, M_{t-1}^{\alg}, Y_{t-1}\}$, for $t > 1$. See Figure~\ref{fig:inc-match} for an illustration.

An adaptive matching obeying the incremental matching constraint is referred to as an incremental matching algorithm. An adaptive incremental matching algorithm that in each round re-matches pairs of nodes greedily with respect to the conditional expected instantaneous regret in the next round according to the posterior belief about node states is referred to as a \emph{greedy Bayes algorithm}. For a given value function $\f$, we let $\gb$ denote a greedy Bayes algorithm. Note that we omit explicit referral to $\f$ in the notation $\gb$ for simplicity.

For any given value function $\f$, let $M^\ast$ be a maximum value matching of nodes in $V$, i.e. $M^\ast\in \arg\max_M \mu_w(M)$. Then, the \emph{regret} of an adaptive algorithm $\alg$ is defined as
$$
\breg_{\f}^{\alg}(T) = T \E[\mu_{\f}(M^\ast)] -  \E\left[\sum_{t=1}^T \mu_{\f}(M_t^{\alg})\right] 
$$
where in the right-hand side of the last equation, the expectations are taken with respect to the prior distribution of the node types and the prior distribution of node types and any randomization of algorithm $\alg$, respectively. Note that $\breg_{\f}^\alg(T)$ is the \emph{expected Bayesian regret}. We aim to characterise the asymptotic regret of greedy Bayes algorithm $\gb$ for some value functions $\f$ as the time horizon $T$ goes to infinity and the number of nodes $n$ goes to infinity. 

Note that for any value function $\f$ and any adaptive algorithm $\alg$, the initial matching is a random matching of nodes in $V$ because the prior distribution of node types is non-informative for maximum value matching. 

\subsection{Algorithms and regret bounds}
\label{sec:results}

We first show a regret lower bound that holds for any adaptive matching algorithm under mild assumptions. Let $\tilde{R}_{\f}$ denote the expected instantaneous regret incurred for a random matching of nodes and value function $\f$. The following lemma provides a regret lower bound that holds for any adaptive incremental matching algorithm.

\begin{lemma}[lower bound]  For any adaptive incremental matching algorithm for value function $\f$, there exists a constant $c>0$ such that the expected regret is at least $c\tilde{R}_{\f}^2$, provided that $\tilde{R}_{\f}$ is larger than some constant value that depends only on $c$ and $\f$.
\label{lm:lb}
\end{lemma}

The proof of the lemma is provided in Section~\ref{greedy-bayes-proof-lm:lb}. The proof follows by the fact that any adaptive incremental matching algorithm reduces the instantaneous regret in each round for at most a constant amount. From this and some simple arguments, it follows that the cumulative regret is lower bounded by the quadratic function of the expected instantaneous regret of initial random matching.

\begin{table}[t]
\caption{Regret lower bounds for different value functions.}
\begin{center}
\begin{tabular}{c | c}
$w$ & Regret lower bound\\\hline
OR & $\Omega(\min\{p,1-p\}^4 n^2)$\\
AND & $\Omega((p(1-p))^2 n^2)$\\\hline
\end{tabular}
\end{center}
\label{tab:lb}
\end{table}

Regret lower bounds for two value functions are shown in Table~\ref{tab:lb}. These lower bounds follow from Lemma~\ref{lm:lb} and computations given in Section~\ref{sec:rnd}.

We will compare the regret upper bounds (which in some cases are tight) developed for greedy Bayes algorithms in the next sections and compare then with given lower bounds. Any additional factor in a regret upper bound that depends on the population parameter $p$ to the one that is present in the corresponding lower bound quantifies the search cost of a greedy Bayes strategy. 

We will see that different value functions $\f$ can be separated into two classes depending on whether or not they satisfy the constant improvement property. A value function is said to satisfy \emph{a constant improvement property} if in each round of the execution of the greedy Bayes algorithm, the instantaneous regret is guaranteed to be decreased for at least a constant amount $L$ over a constant number of next $M$ rounds, where $L$ and $M$ are constants that do not depend on either $n$ or $p$. For the class of value functions that satisfy the constant improvement property, the regret of a greedy Bayes algorithm is essentially determined by the instantaneous regret incurred in the initial random matching and the regret differs from the corresponding lower bound only for a constant factor. For value functions that do not satisfy the constant improvement property, the regret has an additional factor depending on the population parameter $p$ that is due to the search cost of the greedy Bayes algorithm.

\subsubsection{Maximum value function (OR)}

In this section, we consider the maximum value function, defined as follows
$$
\f(x,y) = x \vee y.
$$

In this case, any optimal matching has the maximum number of matches of nodes of different types. Hence, an adaptive algorithm has to efficiently identify pairs of nodes of different types. Note that based on the observed value of a node pair, this node pair belongs to one of the following two sets $S_0 = \{(u, v): \f(\theta_u, \theta_v) = 0\}$ or $S_1 = \{(u, v): \f(\theta_u, \theta_v) = 1\}$. Here, $S_0$ is the set of all $0$-$0$ type pairs, and $S_1$ is the set of all $0$-$1$ and $1$-$1$ type pairs. Under the incremental matching constraint, the conditional expected reward is maximized by choosing one pair from $S_0$ and one pair from $S_1$ and rematching them. This strategy underlies the greedy Bayes algorithm. 

The main idea behind the algorithm is to exploit the \emph{known} low type nodes to explore the \emph{unknown} high type nodes. The algorithm maintains a set of node pairs known to be all of low type (thus each pair in this set has two nodes of low type) and a set of node pairs whose values were observed to be of high type (thus each pair in this set has at least one node of high type). In each round, the algorithm chooses a pair from the known set and a pair from the unknown set and then changes pairing of nodes of these two pairs. From observed pair values, the algorithm can identify the type of all nodes of the two chosen pairs. The algorithm terminates as soon as the types of all nodes in the unknown set are identified. The greedy Bayes algorithm corresponds to the 1-chain algorithm in \cite{KKL18}.

The asymptotic regret of the greedy Bayes algorithm is characterized in the following theorem.

\begin{theorem} \label{thm:max} For maximum value function $\f$, the regret of greedy Bayes algorithm satisfies
\begin{eqnarray*}
\lim_{T\rightarrow \infty}\breg_{\f}^{\gb}(T) & \sim & \frac{1-(1-p)^2}{8p^2}\min\{p,1-p\}^4 n^2, \hbox{ for large } n.
\end{eqnarray*}
\end{theorem}

The proof of Theorem~\ref{thm:max} is provided in Section~\ref{app:max}. Here we provide the proof sketch. The key observation in the proof is that the instantaneous regret in each round is a linear function of the number of matches of $1$-$1$ type in the next round. We can show that the number of matches of $1$-$1$ type evolves according to a time-inhomogeneous Markov chain. The regret analysis boils down to computing the expected values of the states of this Markov chain until a stopping time at which either the number of matches known to be of $0$-$0$ type or the number of matches known to be either of $0$-$0$ or $0$-$1$ type is equal to zero. The analysis uses the Wormald's convergence theorem for bounding the expected value of some nonlinear functions of the Markov chain state. 

Under maximum value function, matching nodes of different type is preferred and the difficulty of finding the optimal matching is in removing all pairs of nodes of same type whenever possible. If the nodes are of predominantly of one type, i.e. when $p$ or $1-p$ is small, a minority-type node is matched with a node of the same type with a small probability in the initial random matching. This means that the initial random matching would typically be close to an optimal matching and, hence, the regret of the algorithm is small. In the other extreme, when there is a balanced number of nodes of different types, i.e. when $p$ is close to $1/2$, any node type is equally likely to be matched with a node of either type in the initial random matching. The initial random matching can thus be far from an optimal matching, and hence, the regret will be high. 

Theorem~\ref{thm:max} gives us a precise characterization of how the regret depends on the population parameter $p$. It indeed confirms the intuition that regret would be small when the population is predominantly of one type, and that the worst-case regret is achieved when there is a balanced number of node types. Observe that the dependency of the asymptotic regret on the population parameter $p$ is not symmetric around $1/2$. It can be readily checked that for any $0 < \gamma < 1/2$, the asymptotic regret when $p = \gamma$ is larger than the asymptotic regret when $p = 1-\gamma$. In the limit of small $p$, the asymptotic regret is approximately $(1/4) p^3 n^2$, while in the limit of small $1-p$, the asymptotic regret is approximately $(1/8)(1-p)^4 n^2$. This shows tƒhat it is harder to find an optimal matching by greedy search when $p$ is small (many low-type nodes) than when $1-p$ is small (many high-type nodes). 

Compared with the lower bound shown in Table~\ref{tab:lb}, the regret upper bound has the extra factor $(1-(1-p)^2)/p^2 = (2-p)/p$. This extra factor quantifies the regret due to the search cost of the greedy Bayes algorithm. This extra factor scales as $1/p$ for small value of $p$. 

\begin{figure}[t]
\begin{center}
\includegraphics[width=7cm]{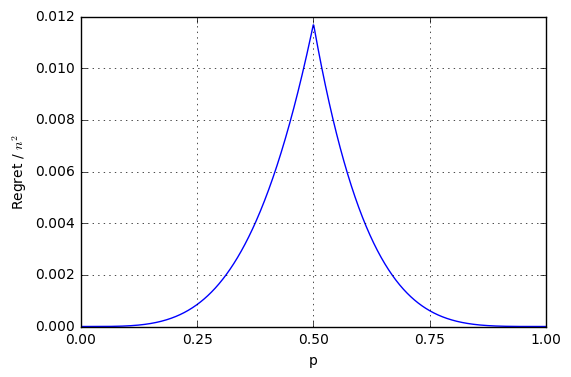}
\caption{The asymptotic normalized regret for maximum value function, in the limit of large number of nodes $n$, versus population parameter $p$.}
\label{fig:regret-max}
\end{center}
\end{figure}

\subsubsection{Minimum value function (AND)} \label{sec::and}

In this section we consider the case of minimum value function defined as
$$
\f(x,y) = x\wedge y.
$$
Under minimum value function, a greedy Bayes algorithm aims at searching for high-type nodes in order to match them together, as only in this case a positive reward is accrued.

The posterior distribution of node types can be represented by a collection of sets, each of which we refer to as a  \emph{matching set}, and the identities of nodes whose types are determined with certainty from observed pair values. Each matching set contains a set of matches such that there is \emph{at most one node} of high type in this set. The greedy Bayes algorithm explores pairs of nodes that reside in different matching sets, and, as we will show, prioritizes the search to matching sets of smallest size.

For any two matching sets, evaluation of all pairs of nodes belonging to these two matching sets results in two possible outcomes: either (a) a $1$-$1$ match is found, in which case the two matching sets are removed from further consideration, or (b) a $1$-$1$ match is not found, in which case the two matching sets are merged into a new matching set. In the former case, each matching set has exactly one high-type node, and having found a match between the two high-type nodes, all other nodes in the two matching sets are identified to be of low-type with certainty. On the other hand, in the latter case, having not found a $1$-$1$ match of nodes spanning the two matching sets, the only information deduced from observed pair evaluations is that there is at most one high-type node in the union of the two matching sets. Hence, in this case we merge the two matching sets into a new matching set. 

\begin{figure}[t]
\begin{center}
\includegraphics[width=10cm]{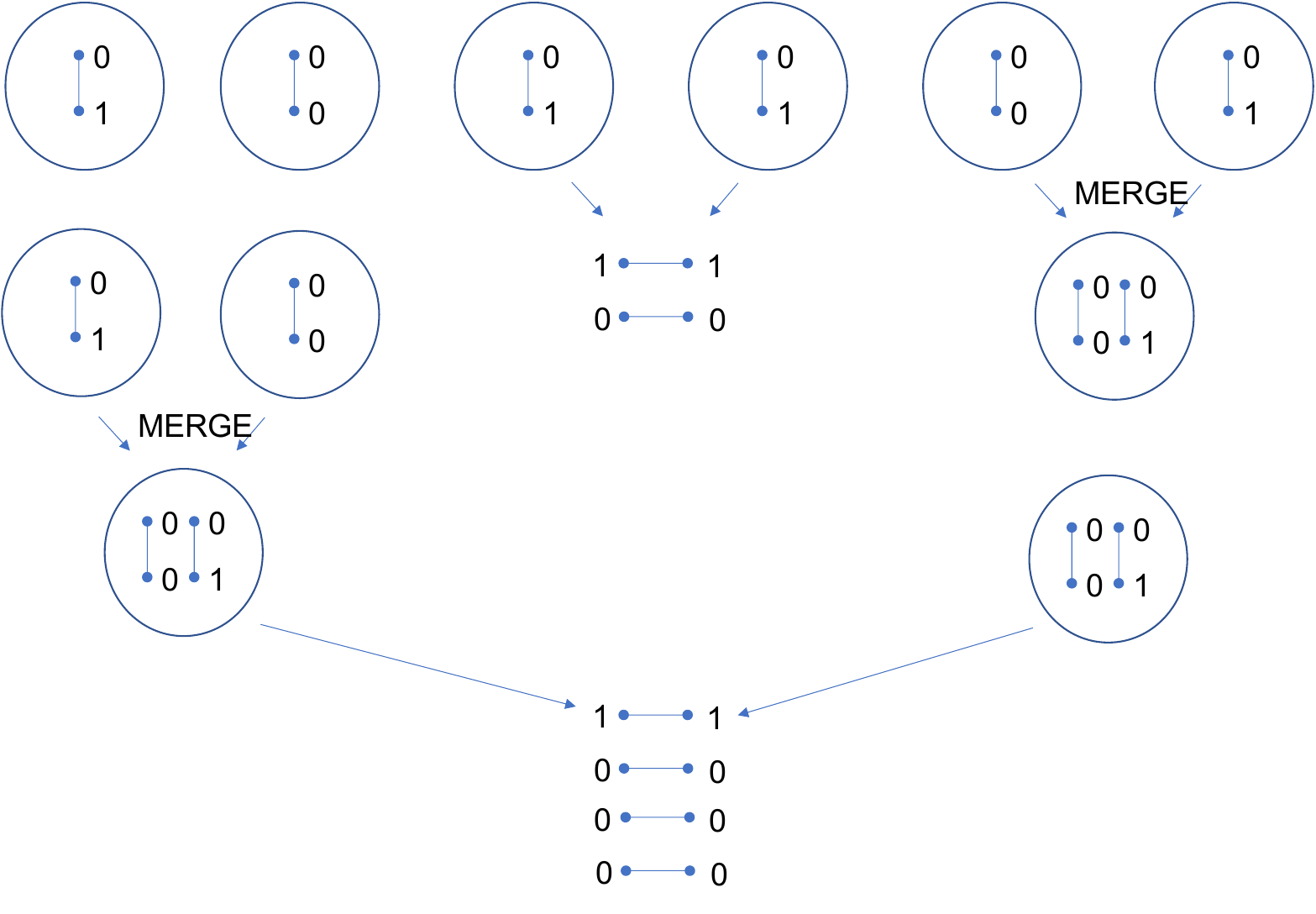}
\caption{An illustration of evolution of matching sets of greedy Bayes matching strategy for AND value function.}
\label{fig:and-merge}
\end{center}
\end{figure}

This exploration procedure ends when matching sets coalesce into one set or no matching set remains. In the former case, the remaining matching set contains at most one high-type node, which cannot be matched with any other high-type node as all of them are already matched to a high-type node. 

The following is a key lemma to show that greedy Bayes strategy explores matching sets of smallest sizes first. 

\begin{lemma} For every round of greedy Bayes algorithm and every set of matches $A$ of cardinality $a$ for which the available information is that $A$ contains at most one high-type node, the posterior probability of $A$ containing a high-type node is 
$$
\pp(a) = \frac{pa}{1-p+pa}.
$$ 
\label{lm:pa}
\end{lemma}

The proof of the lemma is provided in Section~\ref{sec:pa}. The proof is by induction over cardinality of matching sets. 

Note that the posterior probability of a matching set containing a high-type node depends only on its cardinality. Intuitively, the larger the cardinality of a matching set, the larger the posterior probability of this matching set containing a high-type node. Note that any specific node in a matching set $A$ of cardinality $a$ is a high-type node with probability $\pp(a)/(2a) = p/(2(1-p+pa))$, which is decreasing in the cardinality of $A$. This implies that the greedy Bayes strategy is to explore pairs of nodes residing in two different matching sets of smallest size. 

We refer to greedy Bayes algorithm for minimum value function as {\sc Least-Size-Merge} algorithm, which is formally defined in Algorithm~\ref{alg:min}. This algorithm uses a procedure {\sc Is-Merge} defined in Algorithm~\ref{alg::ismerge}. This procedure uses as input two matching sets and constructs matchings of nodes by exploring pairs of nodes residing in the two matching sets.   

\begin{algorithm}[ht]
\caption{Least-Size-Merge}
\label{alg:min}
\SetKwInOut{Init}{Initialization}
\SetKwInOut{Input}{Input}
\Input{nodes set $N$.} 
\Init{iteration $t=1$, $M_1\gets$ a random matching on nodes $N$ and $S\leftarrow \{\{(i, j)\}\in M_1 : Y_1(i,j)= 0\}$.}
\While{$|S| > 1$}{
    $A \leftarrow \underset{S'\in S}{\arg\min} |S'|$\;
    $B \leftarrow \underset{S'\in S\backslash \{A\}}{\arg\min} |S'|$\;
    $S \leftarrow S \backslash \{A, B\}$\;
    \If{IS-MERGE$(A, B) ==$ True}{
        $S \leftarrow S \cup \{A \cup B\}$
    }
}
\end{algorithm}

From the above discussion, it is clear that greedy Bayes algorithm results in a stochastic process that describes evolution of a collection of matching sets as pairs of matching sets are examined, which results in either merging two matching sets or removing them from the collection of matching sets.  

The initial matching is a random matching. From this initial matching, the algorithm identifies high-type nodes that belong to matches yielding value $1$, and a set of matching sets, each consisting of a pair of nodes for which the evaluation is observed to be of value $0$. Indeed, all pairs in these matching sets are either of $0$-$0$ or $0$-$1$ type. Let $X_1$ be the number of matching sets at step $1$. It is readily observed that
$$
X_1 \sim \mathrm{Bin}\left(\frac{n}{2}, 1-p^2\right).
$$

We divide rounds into epochs. In each epoch $r$, let $S_r$ denote the set of matching sets at the beginning of epoch $r$. Initially, as we discussed, $S_1$ consists of pairs of type $0$-$0$ or $0$-$1$ defined by the initial random matching and the cardinality of this set is $X_1$. At epoch $r$, a pair of matching sets $A$ and $B$ of the smallest size is removed from $S_r$ and pairs of nodes spanning $A$ and $B$ are evaluated until either a match $1$-$1$ is found, or not such a match is found. In the latter case, we deduce that nodes in $A\cup B$ contain at most one high-type node. The set of matching sets in the next epoch,  $S_{r+1}$, consists of all unexplored matching sets from $S_r$, and $A\cup B$ if the outcome of examining $A$ and $B$ resulted in not finding a $1$-$1$ match.

In each epoch, the posterior distribution of node types is fully defined by the identities of nodes inferred to be of high-type, and the set of matching sets. The posterior probability of a matching set containing a high-type node is given in Lemma~\ref{lm:pa}. 

\begin{algorithm}[ht]
\caption{IS-MERGE}
\label{alg::ismerge}
\SetKwInOut{Init}{Initialization}
\SetKwInOut{Input}{Input}
\Input{two matching sets $A, B$.}
\Init{create a candidate set $P_i \leftarrow B$ for all $i\in A$}
\For{all node $u\in A$}{
    \While{$P_u \neq \emptyset$}{
    $t\leftarrow t+1$\;
    choose an node $v\in P_u$ and let $(u, u'), (v, v') \in M_{t-1}$\;
    $M_t\gets (M_{t-1}\backslash\{(u, u'), (v, v')\} \cup \{(u, v), (u', v')\}$\;
    \If{$Y_t(u, v) + Y_t (u', v') = 1$}{
        \textbf{return} False\;
    }
    $P_{u} \leftarrow P_{u} \backslash \{v\}$\;
    \If{$u'\in A$}{
        $P_{u'} \leftarrow P_{u'} \backslash \{v'\}$\;
    }
    \If{$v'\in A$}{
        $P_{v'} \leftarrow P_{v'} \backslash \{u'\}$
    }
    }
    \textbf{return} True\;
}

\end{algorithm}

It is readily observed that at the beginning of each epoch, all matching sets are of the same cardinality except for at most one set of larger cardinality. We call this latter matching set a \emph{special matching set}, if one exists, and all other as \emph{regular matching sets}.

As the greedy strategy prioritizes evaluating nodes residing in smallest matching sets, we can divide epochs in super-epochs. In each super-epoch, disjoint pairs of regular matching sets are created and are then examined in arbitrary order, and if there is a special matching set, it is paired at the end of the super-epoch with a regular set for examination. For super-epoch $s$, let $X_s$ denote the number of regular matching sets and let $Y_s$ denote the cardinality of the special matching set; if there is no special matching set, we define $Y_s = 0$. 

Stochastic process $\{(X_s, Y_s)\}_{s\geq 0}$ is a Markov chain with initial state$(X_0,Y_0) = (n/2, 0)$, and for each super-epoch $s\geq 0$, 
\begin{equation}
X_{s+1} \mid (X_s,Y_s) \sim \mathrm{Bin}\left(\left\lfloor\frac{X_s}{2}\right\rfloor, 1-\pp(2^s)^2\right)
\label{equ:X}
\end{equation}
and
\begin{equation}
Y_{s+1} \mid (X_s,Y_s,X_{s+1}) \sim 
\left\{\begin{array}{ll}
(Y_s+\xi_s)\mathrm{Ber}\left(1-\pp(\xi_s)\pp(Y_s)\right) & \hbox{ if } Y_s > 0\\
(2^s + 2^{s+1})\mathrm{Ber}\left(1-\pp(2^s)\pp(2^{s+1})\right) & \hbox{ if } Y_s = 0, X_s \hbox{ is odd}, X_{s+1}>0\\
0, & \hbox{ otherwise}
\end{array}\right .
\label{equ:Y}
\end{equation} 
where 
$$
\xi_s = \left\{
\begin{array}{ll}
2^s & \hbox{ if } X_s \hbox{ is odd}\\
2^{s+1} & \hbox{ if } X_s \hbox{ is even, and } X_{s+1} > 0\\
0 & \hbox{ otherwise}
\end{array}
\right . .
$$
%and, if $Y_s = 0$, $X_s$ is odd, and $X_{s+1} > 0$,
%$$
%Y_{s+1} \mid (X_s,Y_s) \sim (2^s + 2^{s-1})\mathrm{Ber}\left(1-p(2^s)p(2^{s+1})\right)
%$$
%and, otherwise, $Y_{s+1} \mid (X_s, Y_s) = 0$.

\begin{figure*}[t]
\begin{center}
\includegraphics[width=0.7\linewidth]{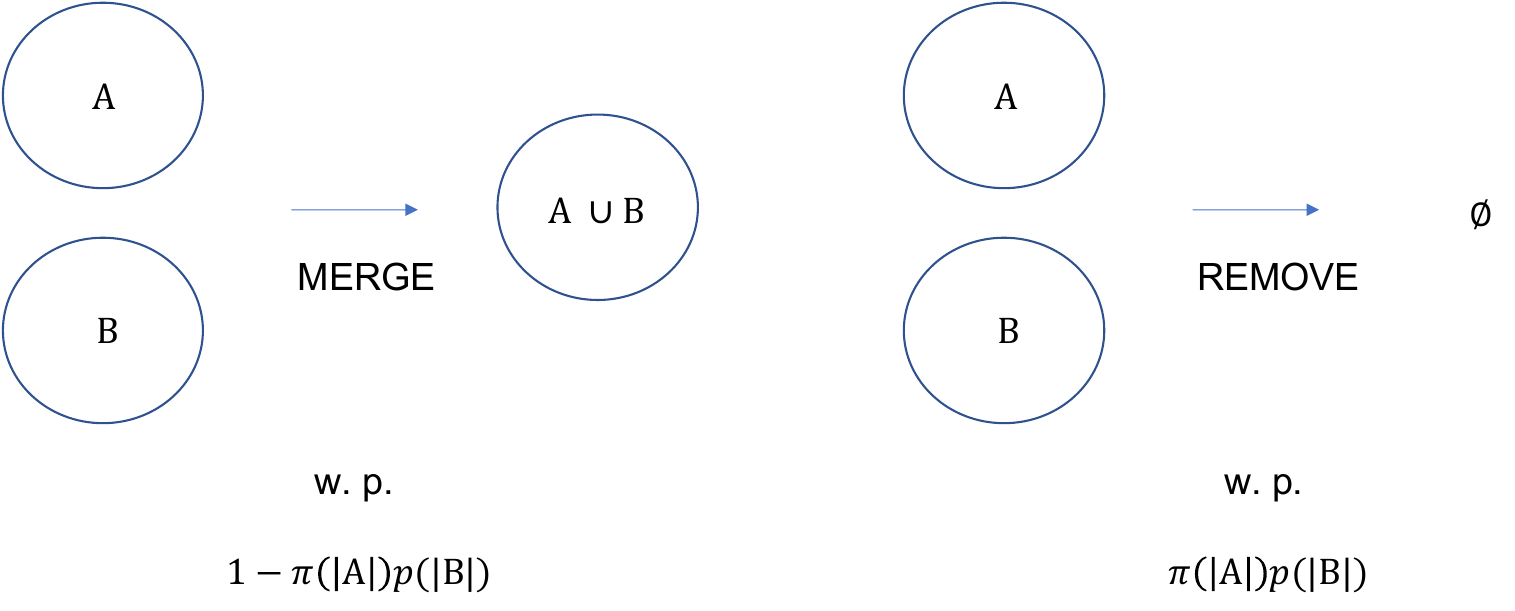}
\caption{Two possible outcomes: merge or remove two matching sets.}
\label{fig:merge}
\end{center}
\end{figure*}

The transition probabilities of the Markov chain $\{(X_s,Y_s\}_{s\geq 0}$ can be interpreted as follows. In each super-epoch, matchings are first created by swapping matches for nodes belonging to different regular matchings sets, by examining one pair of regular matching sets after other. There are $\lfloor X_s/2\rfloor$ pairs of regular matching sets that are examined in super-epoch $s$. Each such examination of a pair of regular matching sets $A$ and $B$ results in either $A$ and $B$ being merged, which occurs with probability $1-\pp(|A|)\pp(|B|)$, or $A$ and $B$ being removed from further consideration which occurs with probability $\pp(|A|)\pp(|B|)$.  Observe that $\pi(|A|)\pi(|B|)$ is the probability that both $A$ and $B$ contain a high-type node. The outcomes of this procedure are illustrated in Figure~\ref{fig:merge}. All regular matching sets in super-epoch $s$ are of cardinality $2^s$. Hence, in super-epoch $s$, two regular matching sets are merged with probability $1-\pp(2^s)^2$, and are removed from further consideration with probability $\pp(2^s)^2$. This explains the conditional distribution of the number of regular matching sets in super-epoch $s+1$, $X_{s+1}$, given in (\ref{equ:X}). 

We next explain the conditional distribution of the cardinality of the special set in super-epoch $s+1$, $Y_{s+1}$, conditional on the values of $X_s, Y_s$ and $X_{s+1}$ given by (\ref{equ:Y}). There are three possible cases. First, if $Y_s > 0$, then at the beginning of super-epoch $s$ there is a special matching set of cardinality $Y_s$. This set is examined with a matching set $A$ of cardinality $\xi_s$ at the end of super-epoch $s$ after regular matching sets are examined, if there is at least one matching set left for consideration besides the special matching set. The set $A$ is either (a) one of regular matching sets of cardinality $2^{s}$ if at the beginning of super-epoch $s$ the number of regular matching sets is odd, or (b) a matching set of cardinality $2^{s+1}$ that resulted from merging two regular matching sets in super-epoch $s$ if the number of regular matching sets at the beginning of super-epoch $s$ is even and at least one merging resulted from examinations of regular matching sets. Second, if $Y_{s}=0$, then all matching sets at the beginning of super-epoch $s$ are regular matching sets. In this case, a special matching set can exist in super-epoch $s+1$, only if the number of regular matching sets at the beginning of super-epoch $s$ is odd and at least one merging resulted from examination of regular matching sets in super-epoch $s$. The cardinality $Y_{s+1}$ of this special set is clearly $2^s+2^{s+1}$. Finally, if $Y_s = 0$ and either number of regular matching sets at the beginning of super-epoch $s$ is even or no examination of regular matching sets in super-epoch $s$ resulted in merging, there is no special matching set in super-epoch $s+1$. Note that cardinality of the special matching set in super-epoch $s$ is of cardinality $Y_1=0$ for $s = 1$ and $Y_s \leq 2^1 + \cdots + 2^s \leq 2^{s+1}$ for $s > 1$. 

The Markov chain $\{(X_s, Y_s)\}_{s\geq 0}$ has absorbing states $(0,0)$, $(1,0)$, $(0,z)$, for $0<z\leq n$. In state $(0,0)$, all high-type nodes are identified. In states $(1,0)$ or $(0,z)$, there is at most one unidentified high-type node. It is not needed to identify the type of this node, as all other high-type nodes are already matched with high-type nodes. 

The number of rounds for examination of a pair of matching sets is discussed as follows. For any two matching sets $A$ and $B$, with cardinalities $|A|=a$ and $|B| = b$, let $N(a,b)$ be the smallest number of adjacent matchings that cover all pairs $(u,v) \in A\times B$. It can be readily noted that $ab/2\leq N(a,b)\leq ab$. When exploring pairs in matching sets $A$ and $B$ such that either $A$ or $B$ contains only low-type nodes, the number of exploration steps is equal to $N(a,b)$. On the other hand, when exploring pairs in matching sets $A$ and $B$ such that both $A$ and $B$ contain a high-type node, then the number of exploration steps is a random variable with expected value $N(a,b)/2$.

The number of steps until {\sc Least-Size-Merge} algorithm terminates is characterised in the following proposition.

\begin{proposition} The expected number of steps $\tau$ until termination of algorithm {\sc Least-Size-Merge} is bounded as follows
$$
\E[\tau] = O\left(\min\left\{\frac{n}{p},n^2\right\}\right).
$$
\label{pro:epochs}
\end{proposition}

Proposition~\ref{pro:epochs} characterizes the expected number of steps used by {\sc Least-Size-Merge} algorithm, which leads to an upper bound on the regret of {\sc Least-Size-Merge} algorithm. The details of the proof of Proposition~\ref{pro:epochs} is in Section~\ref{sec::greedy-bayes:min:theorem-1}.

\begin{figure*}[t]
\begin{center}
$n = 128$ \hspace*{5.5cm} $n = 256$\\
\includegraphics[width=7cm]{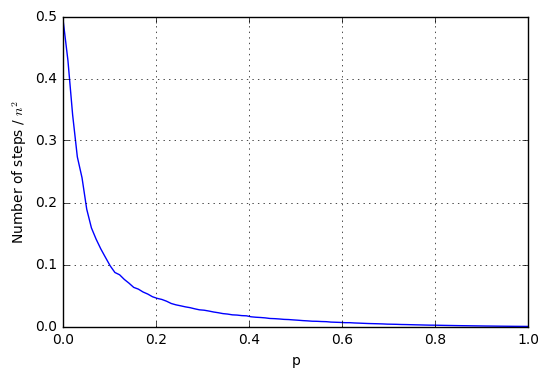}
\includegraphics[width=7cm]{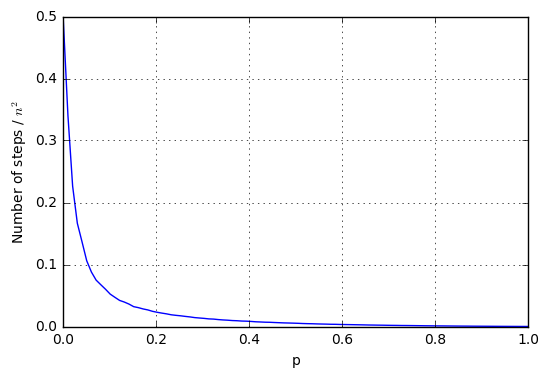}\\
$n = 512$ \hspace*{5.5cm} $n = 1024$\\
\includegraphics[width=7cm]{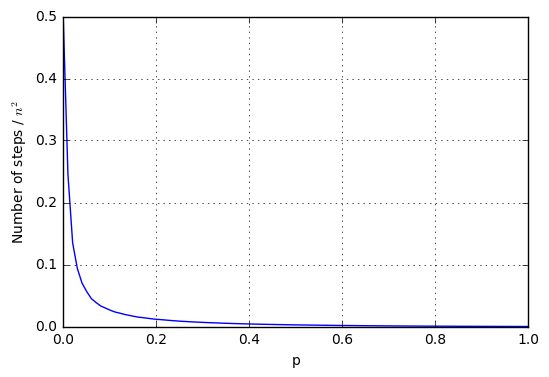}
\includegraphics[width=7cm]{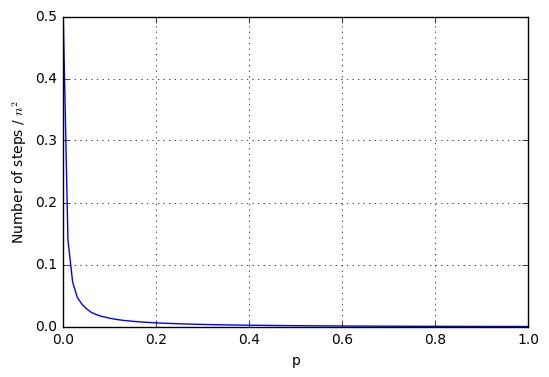}
\caption{The number of rounds until greedy Bayes incremental matching algorithm terminates for minimum (Boolean AND) value function versus population parameter $p$, for different numbers of nodes $n$. %from left to right $n = 128,256,512,1024$. 
%\mv{*** Run the new code to obtain the plots to update the plots ***)}
}
\label{fig:min-ite}
\end{center}
\end{figure*}

%The length of the super-epoch $s$, denoted as $T_s$, can be expressed as follows. 
%\begin{eqnarray*}
%T_s &=& N(2^s,2^s)X_{s+1} + \sum_{i=1}^{\left\lfloor \frac{X_s}{2}\right\rfloor - X_{s+1}} N_i \\
%&& + N(\xi_s, Y_s)\ind_{\{Y_{s+1} > 0\}}\\x
%&& + N_0 \ind_{\{Y_{s+1} = 0\}}
%\end{eqnarray*}
%where $N_i$ are independent and identically distributed random variables with distribution $N_i \sim \mathrm{Bin}(N(2^s,2^s),1/2)$, and $N_0\sim \mathrm{Bin}(N(\xi_s,Y_s),1/2)$ conditional on $\xi_s$ and $Y_s$. 

%\mv{*** Check the proof whether we use any assumptions on $p$, e.g. $p$ being a fixed constant, and then letting $n$ grow. Also for our other theorems. ***}
%\jy{checked. this holds as long as $p\leq 1$.}

\begin{theorem} \label{thm:min-reg} For the minimum value function $\f$, the regret of greedy Bayes algorithm {\sc Least-Size-Merge} satisfies
$$
\lim_{T\rightarrow \infty}\breg_{\f}^{\gb}(T) \leq \left( a_{n, p} + b_{n,p} + c_{n,p}\right) n^2 + O(n\log(n))
$$
where
$$
a_{n, p}=\frac{1}{4} \sum_{s=1}^{\log _{2}(n)} \left(1- \frac{1}{2}p_s\right) \left(1-\frac{1}{2}p_s^2\right) p_s\left(\prod_{i=0}^{s-1} q_{i}^{2}+\frac{1}{n} \sum_{i=0}^{s-1} 2^{i+1}\left(\prod_{j=0}^{i} q_{j}\right)\left(\prod_{j=i+1}^{s-1} q_{j}^{2}\right)\right),
$$
$$
b_{n,p}= \frac{1}{n} \sum_{s=1}^{\log _{2}(n)}  2^{s+1}  p_{s}%\left(1-p_{s}\right) 
\prod_{i=0}^{s-1} q_{i},
$$
and
$$
c_{n,p} = \min\left\{\frac{c}{np},\frac{4}{3}\right\}
$$
with $p_s = p2^s/\left(1-p+p2^s\right)$, $q_s = 1 - p_s^2$, and $c$ is a constant such that $c>16/(\sqrt{2}-1)$. 
Moreover, for any $n\geq 1, p\in(0, 1]$, there exists some constant $M>0$ such that $\max\{a_{n,p}, b_{n,p}, c_{n,p}\} < M$.
\label{thm:min}
\end{theorem}

Proof of the theorem is provided in Section~\ref{proof-min-reg}. The proof separately considers contributions to regret over rounds at which regular matching sets are examined and contributions to regret over rounds at which a special set is examined. 
In the regret bound, $a_{n,p}$ is due to regret accumulated while examining regular matching sets while $b_{n,p}$ and $c_{n,p}$ are due to regret accumulated while examining a special matching set.
Theorem~\ref{thm:min} shows that the asymptotic regret bound of greedy Bayes algorithm scales {\sc Least-Size-Merge} in $O(n^2)$.

Here we give more detailed characterizations of $a_{n, p}$ and $ b_{n,p}$ when the population parameter $p$ depends on the size of population $n$. First, when $p=0$ or $p=1$, any algorithm algorithm has a zero regret.
On the one hand, when the high-type nodes are scarce, i.e. $p\in (0, 1/2]$, it holds that $a_{n,p} = O(1)$ and $b_{n, p} = O(1)$. The search cost of greedy Bayes algorithm {\sc Least-Size-Merge} remains limited and independent of the scarcity of the high-type nodes.
On the other hand, when the low-type nodes are scarce, i.e. $p\in (1/2, 1]$, it holds that $a_{n,p} = O((1-p)^2)$ and $b_{n,p} = O((1-p)1/n)$. The search cost of greedy Bayes algorithm {\sc Least-Size-Merge} increases quadratically in the scarcity of the low-type nodes.
The detailed explicit bounds of $a_{n,p}$ and $b_{n,p}$ for various scenarios are provided in Section~\ref{greedy-Bayes-anp-bnp}.

\begin{figure*}[t!]
\begin{center}
$n = 1024$ \hspace*{5.5cm} $n = 2048$\\
\includegraphics[width=7cm]{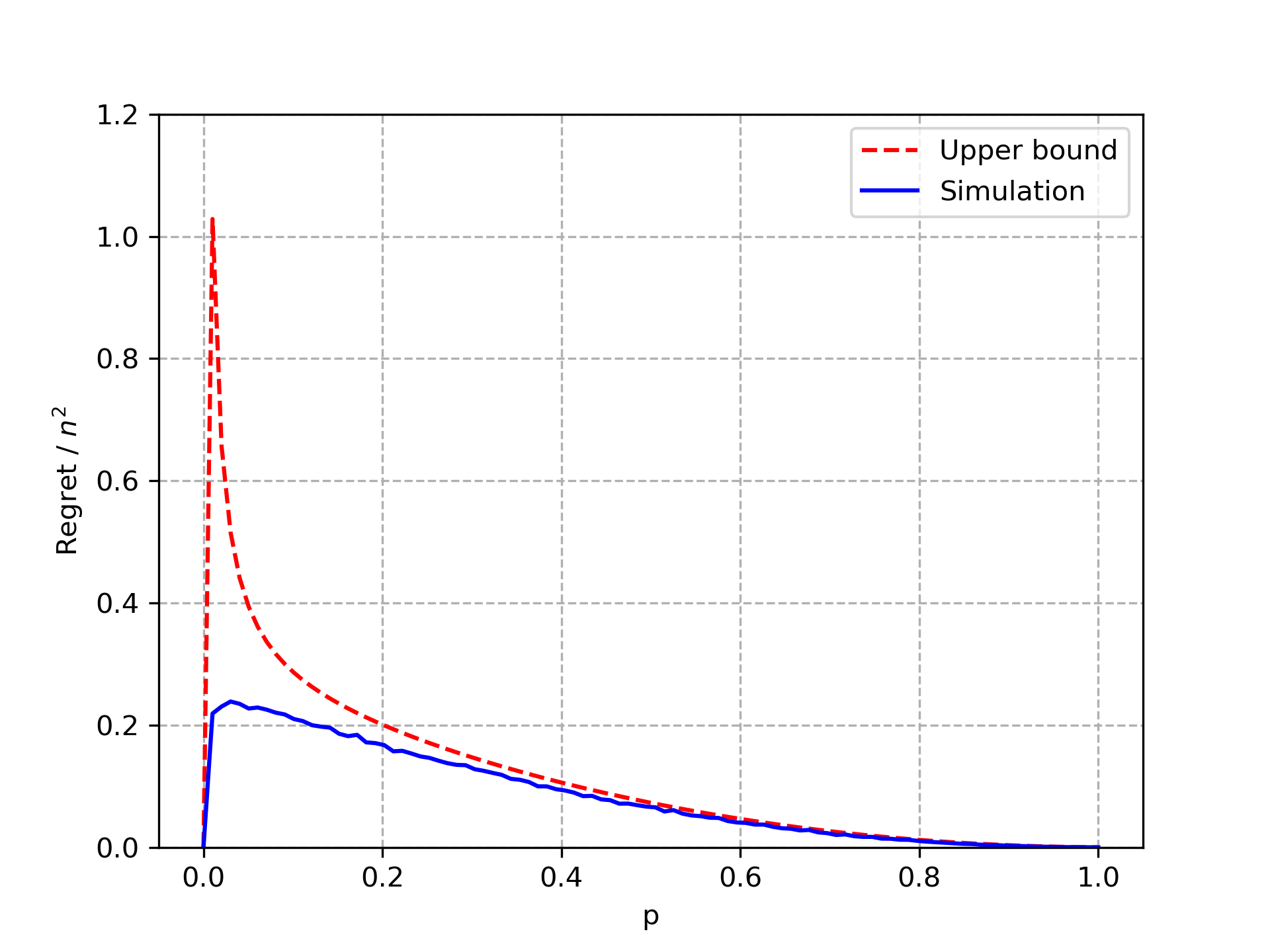}
\includegraphics[width=7cm]{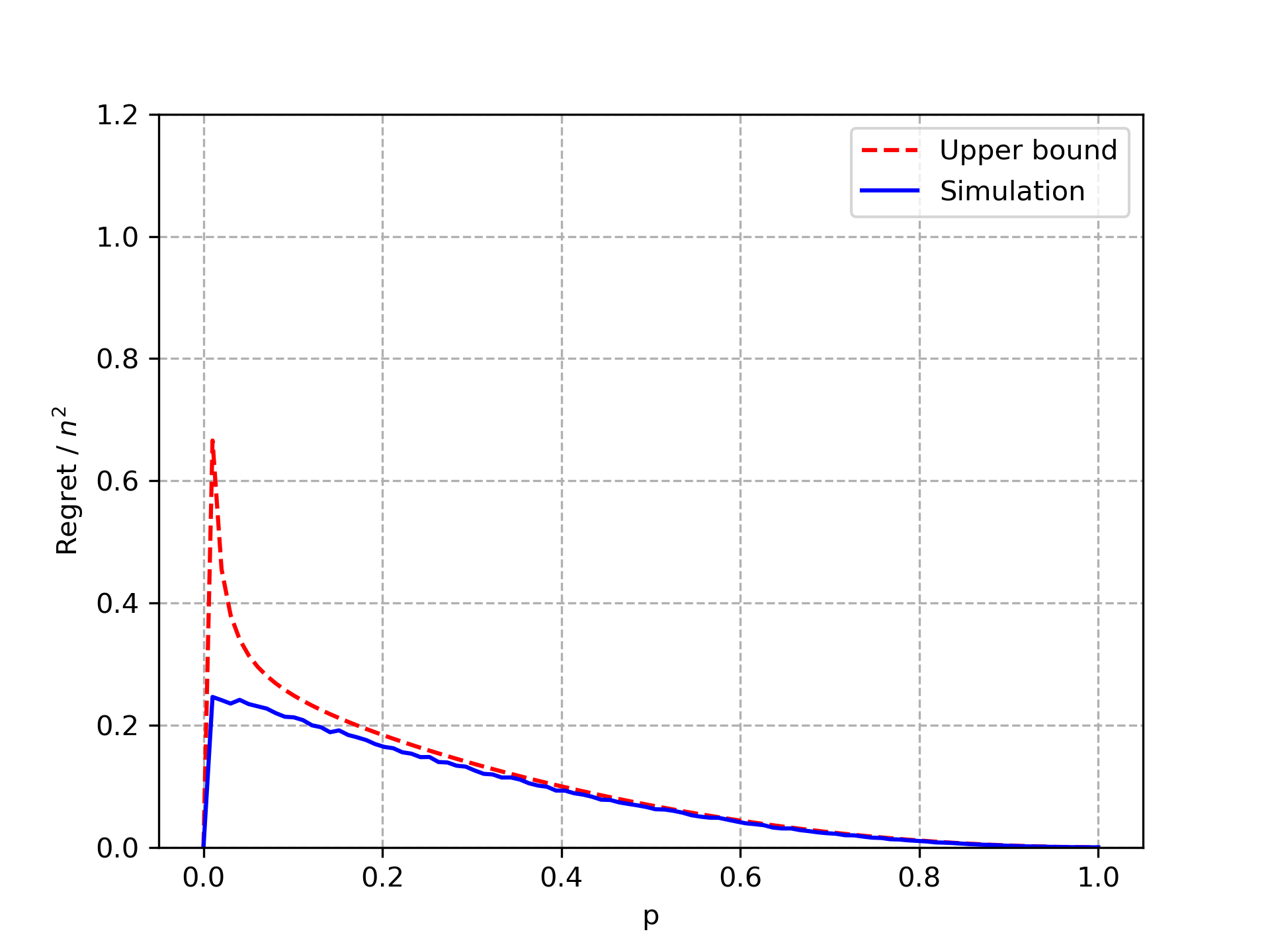}\\
$n = 4096$ \hspace*{5.5cm} $n = 8192$\\
\includegraphics[width=7cm]{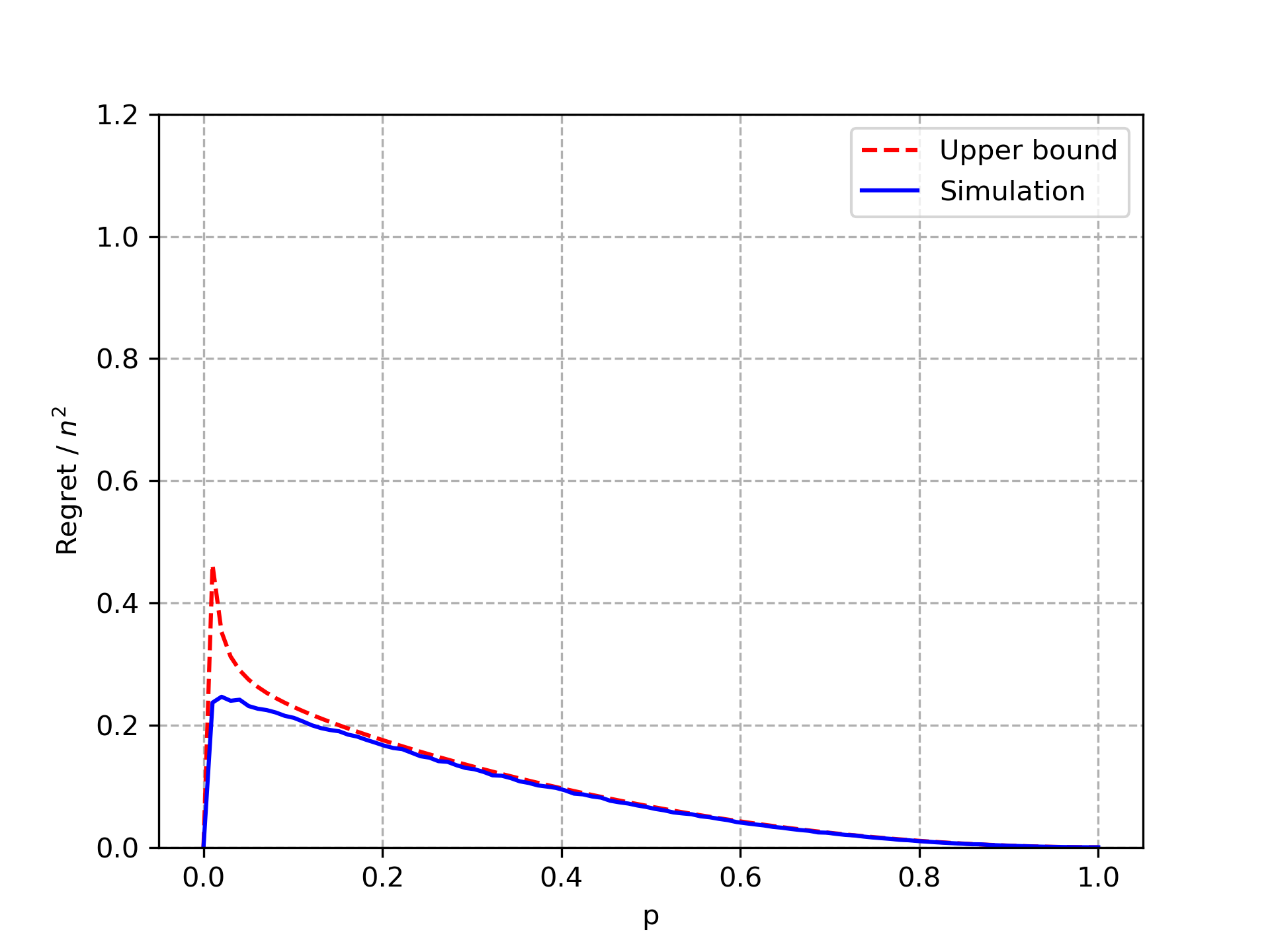}
\includegraphics[width=7cm]{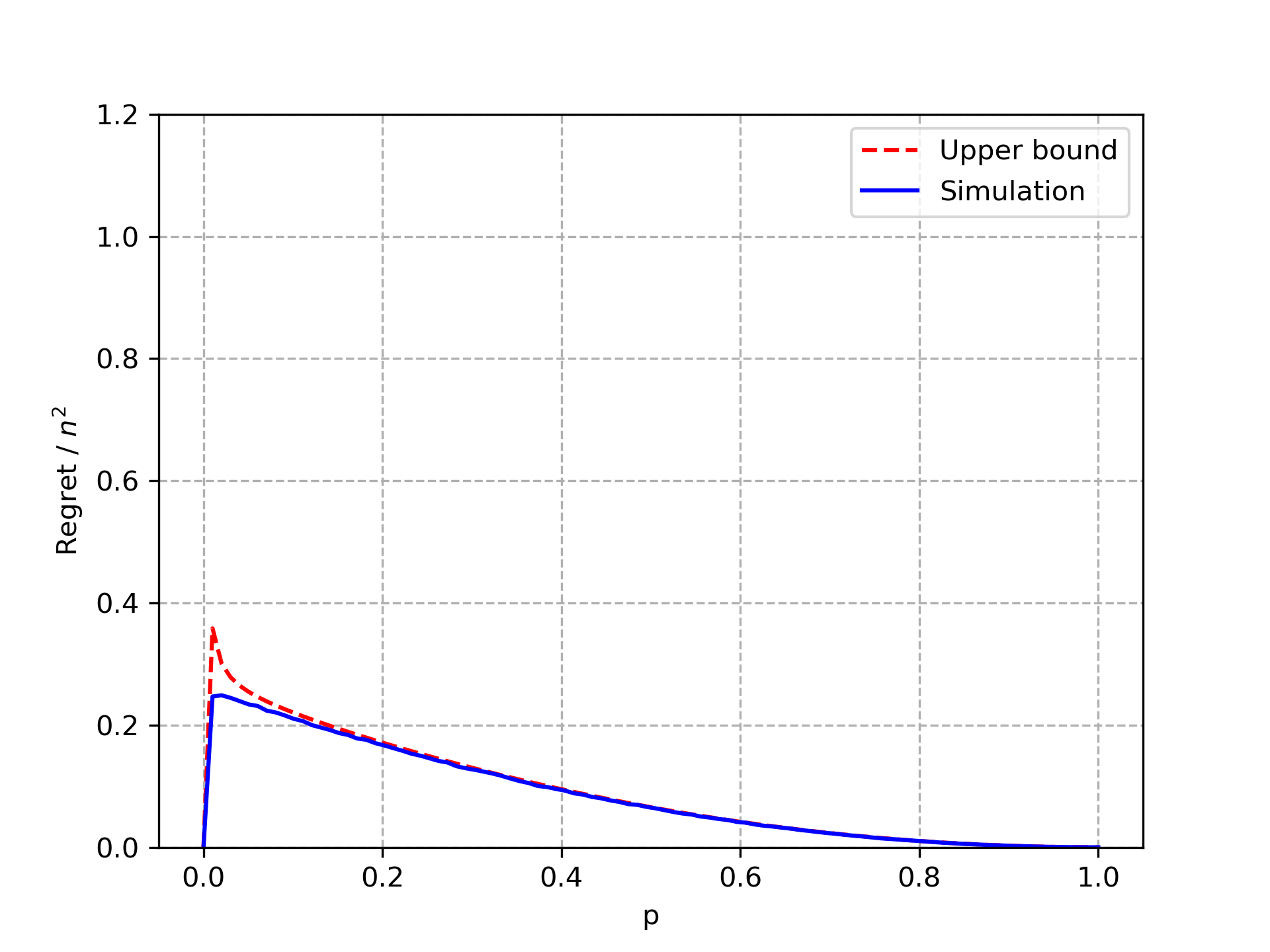}
\caption{Regret of the greedy Bayes algorithm for the MIN value function versus the population parameter $p$, for different values of the number of nodes $n$.}
\label{fig:min-reg}
\end{center}
\end{figure*}

In Figure~\ref{fig:min-reg}, we show the theoretical regret bound in Theorem~\ref{thm:min} along with the regret estimated through simulations versus the value of population parameter $p$ for several different values of the number of nodes $n$. 
We observe that the theoretical bound is tight for large values of $n$. Observe that the regret decreases with the value of the population parameter $p$ for any sufficiently large value of this parameter. This is intuitive as for pair values according to the MIN function, the goal is to maximize the number of $1$-$1$ type matches, which  easier for greedy search when there is a larger number of high-type nodes.

\subsection{Proofs} 
\label{sec::greedy-bayes-proofs}

\subsubsection{Proof of Lemma~\ref{lm:lb}} \label{greedy-bayes-proof-lm:lb}

Let $\f$ be an arbitrary value function and $\alg$ be an arbitrary local switching adaptive matching algorithm for value function $\f$. Algorithm $\alg$ reduces the instantaneous regret in each round for at most some constant value $L > 0$. Recall that any adaptive matching algorithm selects random matching for initial matching because the prior distribution is non-informative for maximum value matching. 

Let $R_t^\alg$ denote the instantaneous regret in round $t$, for $t\geq 1$. Note that it holds 
$$
R_t^\alg \geq \max\{R_1 - L(t-1),0\}, \hbox{ for all } t\geq 1
$$
where $R_1$ denotes the instantaneous regret of initial random matching (we omit the subscript $\alg$ in $R_1$ because this instantaneous regret does not depend on $\alg$).

Hence, for any $T\geq \lfloor \E[R_1]/L\rfloor$,
\begin{eqnarray*}
\E\left[\sum_{t=1}^T R_t^\alg\right] 
&\geq & \sum_{t=1}^T\E[\max\{R_1-L(t-1),0\}]\\
&\geq & \sum_{t=1}^T\max\{\E[R_1]-L(t-1),0\}\\
&\geq & \sum_{t=0}^{\lfloor \E[R_1]/L\rfloor -1} (\E[R_1]-Lt)\\
&=& \E[R_1]\left\lfloor \frac{\E[R_1]}{L}\right\rfloor - L\frac{1}{2} \left(\left\lfloor\frac{\E[R_1]}{L}\right\rfloor-1\right)\left\lfloor\frac{\E[R_1]}{L}\right\rfloor\\
&\geq & \frac{L}{2}\left\lfloor\frac{\E[R_1]}{L}\right\rfloor^2
\end{eqnarray*}
where the second inequality is by Jensen's inequality. 

For any fixed $c\in (0,1/(2L))$, assume that $\E[R_1] \geq L/(1 - \sqrt{2c L})$. Then,
$
\E[R_1]/L-1\geq \sqrt{2cL}(\E[R_1]/L)
$
and, thus, it follows
$$
\frac{L}{2}\left\lfloor\frac{\E[R_1]}{L}\right\rfloor^2 
\geq \frac{L}{2}\left(\frac{\E[R_1]}{L}-1\right)^2
\geq  c\E[R_1]^2.
$$

\paragraph{Regret for random matching}
\label{sec:rnd}

%\mv{*** TBD - at different places may need to impose conditions on $p$ such as $pn = \Omega(1)$ or $pn = \omega(1)$ and similarly for $1-p$. Note that Lemma~\ref{lm:lb} requires $\tilde{R}_w = \omega(1)$ which will require conditions on $p$ and $1-p$. ***}

Let $M$ be a random matching and let $N^{00}$, $N^{01}$ and $N^{11}$ denote the number of matches in $M$ of type $0$-$0$, $0$-$1$ and $1$-$1$, respectively. Let $\f$ be an arbitrary value function. Then, we have
$$
\E[\mu_{\f}(M)] = \f(0,0)\E[N^{00}] + \f(0,1)\E[N^{01}] + \f(1,1)\E[N^{11}].
$$
It follows that
$$
\E[\mu_{\f}(M)] = \frac{1}{2}\left(\f(0,0)(1-p)^2 + \f(0,1)2p(1-p) + \f(1,1)p^2\right)n.
$$

In the following we consider the instantaneous regret of random matching for different value functions $\f$.

\paragraph{Maximum value function} In this case, we have 
$$
\E[\mu_{\f}(M)] = \frac{1}{2}(1-(1-p)^2)n.
$$

Any optimal matching $M^*$ has a maximum possible number of $0$-$1$ type pairs and any remaining high-type nodes matched with high-type nodes. Let $N^1$ denote the number of high-type nodes in the given set of nodes. Then, conditional on $\{N_1=n_1\}$, we have
$$
\E[\mu_{\f}(M^*)\mid N_1 = n_1]=\frac{1}{2}\min\left\{2n_1,n\right\}.
$$
This is because if $n_1\leq n/2$, then every high-type node is matched is with a low-type node yielding the total value of $n_1$, and, otherwise, if $n_1 > n/2$, then $n-n_1$ high-type nodes are matched with low-type nodes and the remaining $n_1-(n-n_1)$ high-type nodes are matched with high-type nodes, yielding the total value $(n-n_1) + (1/2)(n_1 -(n-n_1)) = n/2$. 

It follows that the expected weight of an optimal matching is
$$
\E[\mu_{\f}(M^*)] = \frac{1}{2}\E[\min\{2N_1,n\}]% \sim \min\left\{p,\frac{1}{2}\right\}n \hbox{ for large } n.
$$

Hence, we have
$$
\tilde{R}_w = \E[\mu_{\f}(M^*)]-\E[\mu_{\f}(M)] %
= \E\left[\min\left\{\frac{N_1}{n}-p + \frac{1}{2}p^2, \frac{1}{2}(1-p)^2\right\}\right]n.
%\sim \frac{1}{2}\min\{p,1-p\}^2n, \hbox{ for large } n.
$$
Since, for any real number $a$, $x\mapsto \min\{x,a\}$ is a concave function, by Jensen's inequality and $\E[N_1/n]-p = 0$, we have
$$
\tilde{R}_w \leq \frac{1}{2}\min\{p,1-p\}^2 n.
$$

We next show that it holds
$$
\tilde{R}_w \geq \frac{1}{2}\min\{p,1-p\}^2 n (1-o(1))
$$
provided that 
$$
\frac{p}{\log(1/p)^{1/4}} = \omega\left(\frac{1}{(c_n^2 n)^{1/4}}\right),
$$
and
$$
1-p = \omega\left(e^{-\frac{1}{2^6}c_n^2 n}\right)
$$
where $c_n$ is any sequence such that $c_n = o(1)$. Hence, under these conditions, 
$$
\tilde{R}_w \sim \frac{1}{2}\min\{p,1-p\}^2 n, \hbox{ for large } n.
$$

By Hoeffding's inequality,
$$
\Pr\left[\frac{N_1}{n}-p < -c_n \frac{1}{2}p^2\right]\leq e^{-\frac{1}{2}c_n^2 p^4 n}.
$$
It follows
\begin{eqnarray*}
\tilde{R}_w &=& \E\left[\min\left\{\frac{N_1}{n}-p + \frac{1}{2}p^2, \frac{1}{2}(1-p)^2\right\}\right]n\\
&\geq & \min\left\{(1-c_n)\frac{1}{2}p^2, \frac{1}{2}(1-p)^2\right\}n\left(1-e^{-\frac{1}{2}c_n^2 p^4 n}\right) - p e^{-\frac{1}{2}c_n^2 p^4 n}\\
& \geq & \frac{1}{2}\min\{p,1-p\}^2 n \left((1-c_n) \left(1-e^{-\frac{1}{2}c_n^2 p^4 n}\right)-2\max\left\{\frac{1}{p}, \frac{p}{(1-p)^2}\right\}e^{-\frac{1}{2}c_n^2 p^4 n}\right).
\end{eqnarray*}

Under conditions $c_n = o(1)$ and $p/\log(1/p)^{1/4} = \omega(1/(c_n^2 n)^{1/4})$, it clearly holds
$$
(1-c_n) \left(1-e^{-\frac{1}{2}c_n^2 p^4 n}\right) \geq 1-o(1).
$$
We next show that
$$
2\max\left\{\frac{1}{p}, \frac{p}{(1-p)^2}\right\}e^{-\frac{1}{2}c_n^2 p^4 n} 
= o(1).
$$

For the case when $p \leq 1/2$, we have
\begin{eqnarray*}
2\max\left\{\frac{1}{p}, \frac{p}{(1-p)^2}\right\}e^{-\frac{1}{2}c_n^2 p^4 n} & =& 2\frac{1}{p}e^{-\frac{1}{2}c_n^2 p^4 n}\\
&=& 2 e^{-\frac{1}{2}c_n^2 p^4 n (1-\log(1/p)/((1/2)c_n^2 p^4 n))}\\
&=& 2 e^{-\frac{1}{2}c_n^2 p^4 n (1-o(1))}\\
&=& o(1)
\end{eqnarray*}
where the two last relations hold by the condition $p/\log(1/p)^{1/4} = \omega(1/(c_n^2 n)^{1/4})$.

For the case when $p > 1/2$, 
\begin{eqnarray*}
2\max\left\{\frac{1}{p}, \frac{p}{(1-p)^2}\right\}e^{-\frac{1}{2}c_n^2 p^4 n} &=& 
2\frac{p}{(1-p)^2} e^{-\frac{1}{2}c_n^2 p^4 n}\\
&\leq & 2 e^{-\frac{1}{2^5}c_n^2 n + 2\log\left(\frac{1}{1-p}\right)}.
\end{eqnarray*}

The right-hand side in the last inequality is $o(1)$ if, and only if, 
$$
\frac{1}{2^6}c_n^2 n + \log\left(\frac{1}{1-p}\right) \rightarrow \infty \hbox{ as } n \rightarrow \infty, 
$$
i.e.
$$
1-p = \omega\left(e^{-\frac{1}{2^6}c_n^2 n}\right).
$$

\paragraph{Minimum value function} In this case, we have
$$
\E[\mu_{\f}(M)] = \frac{1}{2}p^2 n.
$$

Any optimal matching has maximum number of $1$-$1$ type pairs. Hence, we have
$$
\E[\mu_{\f}(M^*)] = \E\left[\left\lfloor\frac{N_1}{2}\right\rfloor\right]. 
$$

We next show that the regret of random matching satisfies
$$
\tilde{R}_{\f} = \E[\mu_{\f}(M^*)] - \E[\mu_{\f}(M)] = \frac{1}{2}p(1-p)n (1-o(1))
$$
provided that $p = \omega(1/n)$ and $1-p=\omega(1/n)$.

The regret of random matching has the following upper bound,
$$
\tilde{R}_{\f} = \E[\mu_{\f}(M^*)] - \E[\mu_{\f}(M)] \leq \frac{1}{2}pn - \frac{1}{2}p^2 n = \frac{1}{2}p(1-p)n
$$
and the following lower bound,
$$
\tilde{R}_{\f} = \E[\mu_{\f}(M^*)] - \E[\mu_{\f}(M)] \geq \frac{1}{2}pn - 1 - \frac{1}{2}p^2 n = \frac{1}{2}p(1-p)n - 1 = \frac{1}{2}p(1-p)n(1-o(1))
$$
where the last relation holds under conditions $p = \omega(1/n)$ and $1-p=\omega(1/n)$.

\subsubsection{Proof of Theorem~\ref{thm:max}}
\label{app:max}

%\mv{*** Note that appying Wormald's theorem may require some conditions on $p$, e.g. $p$ being a constant ***.}
%\jy{checked. It holds for all $p$ being constant in $(0, 1)$ or $p=1-o(1)$ or $p=o(1)$.}

For each round $t\geq 0$, let $S_t$ be the number of matches known to be of $0$-$0$ type, $U_t$ be the number of matches known to be of either $0$-$1$ or $1$-$1$ type, and $K_t$ be the number of matches known to be of $0$-$1$ type. 

Let $N_1$ be the number of type $1$ items in round $1$. Let $U_t^{01}$ be the number of matches of $0$-$1$ type and $U_t^{11} := U_t - U_t^{01}$. The following equations hold, for every $t\geq 0$,
\begin{eqnarray}
U_t^{01} + U_t^{11} &=& U_t \label{equ:sum0}\\
S_t + U_t + K_t &=& \frac{n}{2}\label{equ:sum1}\\
U^{11}_t + U_t + K_t &=& N_1.\label{equ:sum2}
\end{eqnarray}

By the law of large numbers, the following limits hold almost surely,
\begin{eqnarray}
\lim_{n\rightarrow \infty} \frac{1}{n}N_1 &=& p\label{equ:n1}\\
\lim_{n\rightarrow \infty} \frac{2}{n}U_1^{11} &=& p^2\label{equ:u111}\\
\lim_{n\rightarrow \infty} \frac{2}{n}S_1 &=& (1-p)^2\label{equ:s1}\\
\lim_{n\rightarrow \infty} \frac{2}{n}U_1 &=& 1-(1-p)^2.\label{equ:u1}
\end{eqnarray}

For any $p\in (0,1)$, $S_1 \wedge U_1 > 0$ almost surely.
% Initially, in round $t=1$, the matching is a random matching, for which it holds $\P[\{S_1=0\}\cup \{U_1=0\}] = [1-(1-p)^2]^{n/2} + [(1-p)^2]^{n/2} = o(1)$. \mv{*** This requires $(1-p)^2 n = \omega(1)$ and $pn = \omega(1)$. ***} 
Hence, in what follows we condition on $\{S_1>0\}\cap \{U_1>0\}$. 

The system dynamics is according to the Markov chain $\{X_t\}_{t\geq 0}$ where $X_t = (S_t, U_t^{01}, U_t^{11},K_t)$ with transition probabilities given as follows
\begin{eqnarray}
S_{t+1} &=& S_t - \xi_t \ind_{\{S_t\wedge U_t > 0\}}\label{equ:mc-s}\\
U^{11}_{t+1} &=&   U^{11}_t - \xi_t \ind_{\{S_t\wedge U_t > 0\}}\label{equ:mc-u11}\\
U^{01}_{t+1} &=&  U^{01}_t - (1-\xi_t) \ind_{\{S_t\wedge U_t > 0\}}\label{equ:mc-u01}\\
K_{t+1} &=& K_{t} + (1+\xi_t )\ind_{\{S_t\wedge U_t > 0\}}\label{equ:mc-k}
\end{eqnarray}
where $\xi_t$ is a Bernoulli random variable with mean $U^{11}_t/U_t$ under condition $U_t > 0$.

Conditional on $\{N_1 = n_1\}$, the expected reward of an optimal matching is 
$$
\mu^* = \frac{n_1 + \min\{n_1, n-n_1\}}{2}.
$$
The instantaneous reward at round $t$ is 
$$
\mu_t = U_t + K_t. %+ \xi_t \ind_{\{S_t\wedge U_t > 0\}}.
$$
%\jy{***why $\xi_t \ind_{\{S_t\wedge U_t > 0\}}$ ***}

Hence, the instantaneous regret $R_t$ at round $t$ is equal to
\begin{equation}
R_t = U^{11}_{t} + \min\left\{\frac{n}{2} - N_1, 0\right\}.
\label{equ:rt}
\end{equation}

The algorithm terminates at the stopping time $\tau$ defined as
$$
\tau = \min\{t > 1\mid  S_t = 0 \hbox{ or } U_t = 0\}.
$$
Note that $\tau$ satisfies, with probability $1$,
$$
S_1\wedge U_1 + 1\leq \tau \leq U_1 + 1.
$$
To see this, note that from (\ref{equ:mc-s}), $S_t \geq S_1 - (t-1)$. Hence, $S_t = 0$ implies $t \geq S_1 + 1$. Similarly, from (\ref{equ:mc-u11}) and (\ref{equ:mc-u01}), $U_t \geq U_1 - (t-1)$, hence $U_t = 0$ implies $t \geq U_1 + 1$. From this, it follows that $\tau \geq S_1 \wedge U_1 + 1$. Now, for the upper bound, suppose first that $S_\tau \geq U_\tau = 0$. Since in this case $U_\tau = U_1 - (\tau-1)$, it follows that $\tau = U_1 + 1$. Otherwise, if $0 = S_\tau < U_\tau$, it follows that $\tau \leq U_1 + 1$, which establishes the upper bound.

From (\ref{equ:sum0}), (\ref{equ:mc-u11}) and (\ref{equ:mc-u01}), for all $t\geq 1$, $U_{t+1} = U_t - \ind_{\{S_t \wedge U_t > 0\}}$. Hence, for all $t\geq 1$,
$$
U_{t} = U_1 - (t-1)\ind_{\{S_t\wedge U_t > 0\}} - \tau \ind_{\{S_t\wedge U_t = 0\}}.
$$

From (\ref{equ:mc-s}) and (\ref{equ:mc-u11}), for all $t\geq 1$, $S_t - U^{11}_t = S_1 - U^{11}_1$. Hence, $S_t > 0$ is equivalent to $U^{11}_t > U^{11}_1 - S_1$.

It follows that the system evolves according to a one-dimensional time-inhomogenous Markov chain $\{U^{11}_t\}_{t\geq 1}$ with transition probabilities given as, for $t \leq u_1$,

$$
\P_1[U^{11}_{t+1} = u -1 \mid U^{11}_t = u ] = \frac{u}{u_1-(t-1)}\ind_{\{u > u^{11}_1 - s_1\}}
$$
and
\begin{eqnarray*}
\P_1[U^{11}_{t+1} = u \mid U^{11}_t = u ] 
= 1-\P_1[U^{11}_{t+1} = u -1 \mid U^{11}_t = u ]
\end{eqnarray*}
and, for $ t > u_1$,
$$
\P_1[U^{11}_{t+1} = u \mid U^{11}_t = u] = 1.
$$

For all $1\leq t \leq u_1$, we have
$$
\E_1[U^{11}_{t+1}] = \E_1[U^{11}_t] - \E_1\left[\frac{U^{11}_t}{U_t} \ind_{\{U^{11}_t > u^{11}_1 - s_1\}}\right].
$$
and, for $t > u_1$,
$$
\E_1[U^{11}_{t+1}] = \E_1[U^{11}_t].
$$

% In the following we consider two cases with respect to the sign of $u_1^{11}-s_1$.
In the following we consider two cases with respect to the sign of $p-1/2$.

\noindent {\bf Case 1}: $p\leq 1/2$. From \eqref{equ:n1}, \eqref{equ:u111} and \eqref{equ:n1}, it holds, almost surely 
$$
U_1^{11} \leq S_1, \quad N_1 \leq \frac{n}{2}.
$$
By \eqref{equ:rt}, it holds, almost surely,
$$
R_t = U^{11}_t.
$$
In this case, we have
$$
\E_1\left[\frac{U^{11}_t}{U_t} \ind_{\{U^{11}_t > u^{11}_1 - s_1\}}\right] = \E_1\left[\frac{U^{11}_t}{U_t}\right] = \frac{1}{u_t}\E_1[U^{11}_t]
$$
and $u_t = u_1 -(t-1)$. Hence, for $1\leq t\leq u_1$, $\E_1[U^{11}_{t+1}] = \frac{u_{t+1}}{u_t}\E_1[U^{11}_t]$, and, thus, expanding this recurrence, we obtain
$$
\E_1[U^{11}_{t}] = \frac{u_t}{u_1}u^{11}_1.
$$
It follows that for all $t\geq 1$,
\begin{equation}
\E_1[U^{11}_{t}] = \left(1 - \frac{t-1}{u_1}\right)^+u^{11}_1
\label{equ:e11t}
\end{equation}
where $(\cdot)^+ = \max\{\cdot,0\}$.

We have, for $t\geq 1$,
\begin{eqnarray*}
\E_1[R_t] &=& \E_1[U^{11}_{t}] = \left(1 - \frac{t-1}{U_1}\right)^+U^{11}_1.
\end{eqnarray*}
Hence, we have
\begin{eqnarray*}
\E\left[\sum_{t=1}^{U_1 + 1} R_t \right] &=& \E\left[\sum_{t=1}^{U_1+1} \left(1 - \frac{t-1}{U_1}\right)^+U^{11}_1\right]\\
&=& \frac{1}{2}\E\left[U^{11}_1 \left(U_1 +1\right) \right]\\
&=& \frac{1}{2}\E[U^{11}_1 U_1] + \frac{1}{2}\E[U^{11}_1]\\
&=& \frac{1}{2}\E[\E[U^{11}_1\mid U_1] U_1] + \frac{1}{2}\E[U^{11}_1]\\
&=& \frac{1}{2}\frac{p^2}{1-(1-p)^2}\E[U_1^2] + \frac{1}{4}p^2 n\\
&=& \frac{1}{2}\frac{p^2}{1-(1-p)^2}\left((1-(1-p)^2)^2 \frac{n^2}{4} + \frac{n}{2}(1-(1-p)^2)(1-p)^2\right) + \frac{1}{4}p^2 n\\
&=& \frac{1}{8}(1-(1-p)^2)p^2 n^2 + \frac{1}{4}p^2 (1-p)^2 n + \frac{1}{4}p^2 n\\
&=& \frac{1}{8}(1-(1-p)^2)p^2 n^2 + \frac{1}{4}p^2(1+(1-p)^2) n\\
&=& \frac{1}{8}(1-(1-p)^2)p^2 n^2 (1+o(1))
%&=& \frac{1}{8} (1-(1-p)^2)p^2 n^2 + \jy{\frac{1}{4}p^2n}. 
\end{eqnarray*}
where the last relation holds under condition $p = \omega(1/n)$.

%When
%$$
%p > 1- \sqrt{1-\frac{2}{n}}
%$$
%the first term is the dominant term.

% \jy{---------------------Remove----------------------}

% \jy{By \eqref{equ:rt},
% \begin{eqnarray*}
% \E_1[R_t] &=& \E_1[U^{11}_{t+1}] + \min\{\frac{n}{2}-N_1,0\} = \left(1 - \frac{t}{u_1}\right)^+u^{11}_1+ \min\{\frac{n}{2}-N_1,0\} .
% \end{eqnarray*}
% }
% \jy{
% Hence we have
% \begin{eqnarray*}
% \E\left[\sum_{t=1}^{U_1 + 1} R_t \right] 
% &=& \E\left[\sum_{t=1}^{U_1 + 1} \E_1[R_t] \right]  \\
% &=& \E\left[\sum_{t=1}^{U_1+1} \left(1 - \frac{t}{U_1}\right)^+U^{11}_1\right]\\
% && + \E\left[(U_1+1)\min\left\{\frac{n}{2}-N_1,0\right\}\right]\\
% &\leq& \frac{1}{2}\E\left[U_1 U^{11}_1\right] + o(n^2)\\
% &\leq& \frac{1}{8} (1-(1-p)^2)p^2 n^2 + o(n^2). 
% \end{eqnarray*}
% }

\noindent {\bf Case 2}: $p>1/2$ From \eqref{equ:n1}, \eqref{equ:u111} and \eqref{equ:n1}, we have, almost surely, 
$$
U_1^{11} > S_1, \quad N_1 > \frac{n}{2}.
$$
By \eqref{equ:rt}
$$
R_t = U^{11}_t + \frac{n}{2} - N_1
$$

We next establish a limit value of $U^{11}_t$ as $n$ goes to infinity, for $p > 1/2$ %\mv{*** Why is this condition here?***}\jy{solved.}
%% \jy{***Do not know where $p>1/2$ comes from ***} 
by applying Theorem~\ref{thm:wormald}. 

Condition (C1) holds because for all $t\geq 1$, it always holds,
$$
|U^{11}_{t+1} - U^{11}_t| \leq 1.
$$
Condition (C2) holds as follows 
% \jy{***Why this holds as Condition (2)***}
$$
\E[U^{11}_{t+1}-U^{11}_t\mid {\mathcal F}_t] = f(t/n,U^{11}_t/n)\ind_{\{U_t^{1,1}/n > \tilde{z}\}}
$$
where $U^{11}_1-S_1 = \tilde{z} n + o(n)$, and 
$$f(t,x) = -\frac{x}{\tilde{u}_1-t},$$ 
for $x > \tilde{z}$, where $u_1 = \tilde{u}_1 n + o(n)$ and $\tilde{u}_1 = \frac{1}{2}p(2-p)$.

Hence, when $p\in (1/2, 1)$ is fixed,
$$
\E[U^{11}_{t+1}-U^{11}_t\mid {\mathcal F}_t] = -\frac{U^{11}_t/n}{\frac{1}{2}p(2-p)-t/n} + o(1),
$$
and, when $p=1-o(1)$,
$$
\E[U^{11}_{t+1}-U^{11}_t\mid {\mathcal F}_t] = -\frac{U^{11}_t/n}{\frac{1}{2}-t/n} + o(1).
$$
For condition (C3), we need to show that $f$ satisfies a Lipschitz condition on $D = \{(t,x): 0 \leq t < t^\star, 0 \leq x \leq 1\}$. Note that
\begin{eqnarray*}
|f(t,x)-f(t',x')|  &= & |f(t,x)-f(t',x)+f(t',x)-f(t',x')|\\
&\leq & \frac{|f(t,x)-f(t',x)|}{|t-t'|}|t-t'|+ \frac{|f(t',x)-f(t',x')|}{|x-x'|}|x-x'|\\
&\leq & L (|t-t'| + |x-x'|) 
\end{eqnarray*}
where
$$
L = \max_{(t,x)\in D} \left|\frac{\partial}{\partial t } f(t,x)\right| \vee \max_{(t,x)\in D}\left|\frac{\partial}{\partial x} f(t,x)\right|.
$$
Since $\partial f(x,t)/\partial t = x/(\tilde{u}_1 -t)^2$ and $\partial f(x,t)\partial x = -1/(\tilde{u}_1 - t)$, we have
$$
L = \frac{\tilde{u}^{11}_1}{(\tilde{u}_1-t^\star)^2} \vee \frac{1}{\tilde{u}_1-t^\star}
$$
where $t^\star$ is the unique positive value such that $u^{11}(t^\star) = \tilde{z}$. Note that
$$
t^\star = \left(1-\frac{\tilde{z}}{\tilde{u}^{11}_1}\right)\tilde{u}_1 = \frac{(1-(1-p)^2)(1-p)^2}{2p^2}
$$
and
$$
\tilde{u}_1 - t^\star = \frac{1-(1-p)^2}{p^2}\left(p-\frac{1}{2}\right).
$$

Next, consider the differential equation, 
$$
\frac{d}{dt}u^{11}(t) = f(t,u^{11}(t)), \hbox{ for } 0\leq t < t^\star.
$$
This differential equation has a unique solution which is given by
$$
u^{11}(t) = \left(1-\frac{t}{\tilde{u}_1}\right)\tilde{u}^{11}_1
$$
where $u^{11}_1 = \tilde{u}^{11}_1 n + o(n)$.

By Theorem~\ref{thm:wormald}, we have, almost surely, for every $1\leq t < t^\star n$,
$$
U^{11}_t = \left(1-\frac{t}{u_1}\right) u^{11}_1 + o(n).
$$
and $o(n) \leq 2n^{4/15}\log(n).$

For the regret, we have %\mv{***Check scalings with $n$ - any conditions on $p$?***}\jy{checked. holds for all $p\in (1/2, 1)$.}
\begin{eqnarray*}
\sum_{t=1}^{t^*n} R_t &=& \sum_{t=1}^{t^\star n} \left(1-\frac{t}{U_1}\right)U^{11}_1 + t^\star n \left(\frac{n}{2}-N_1\right) + t^\ast o(n^2)\\
&=& t^*n \left(U^{11}_1 - N_1 + \frac{n}{2}\right) - \frac{1}{2}(t^\star n)^2 \frac{U^{11}_1}{U_1} + t^\ast o(n^2)\\
&=& \frac{1}{2} (1-p)^2t^\star n^2 - \frac{1}{2}\frac{p^2}{1-(1-p)^2}(t^\star)^2 n^2 + t^\ast o(n^2)\\
&=& \frac{(1-(1-p)^2)(1-p)^4}{8p^2}n^2 + \frac{(2-p)(1-p)^2}{2p} o(n^2).
\end{eqnarray*}

\subsubsection{Proof of Lemma~\ref{lm:pa}}
\label{sec:pa}

We need to show that for any set $A$, the posterior probability that $A$ contains a type $1$ node is equal to
\begin{equation}
\pp(|A|) = \frac{|A|p}{1-p + |A|p}.
\label{equ:pa1}
\end{equation}

We will show this by induction over the cardinality of sets. We first consider the base case for sets of cardinality $2$. It suffices to consider  $(X_1, X_2)$ that are two independent Bernoulli ($p$) random variables, conditional on the event $(X_1,X_2)\neq (1,1)$. Then,
\begin{eqnarray*}
\P[(X_1, X_2)\neq (0,0) \mid (X_1, X_2) \neq (1,1)]
&=& \frac{\P[(X_1,X_2)\in \{(0,1),(1,0)\}]}{\P[(X_1,X_2)\neq (1,1)]}\\
&=& \frac{2p(1-p)}{1-p^2}\\
&=& \frac{2p}{1+p}.  
\end{eqnarray*}
This shows that (\ref{equ:pa1}) holds for the base case.

For the induction step, suppose that (\ref{equ:pa1}) holds for sets of cardinality $\leq k$. We then show that it holds for sets of cardinality $\leq k+1$. To this end, let $A\cup B$ be a set of cardinality $k+1$, with $A$ and $B$ being two disjoint sets of cardinality $\leq k$. Let $\tau(S)$ denote the event that set $S$ has at least one type $1$ item. Then, we have
\begin{eqnarray*}
\P[\tau(A\cup B) \mid \tau(A)^c \vee \tau(B)^c]
&=& \P[\tau(A) \vee \tau(B) \mid \tau(A)^c \vee \tau(B)^c ] \\
&=& \frac{\P[(\tau(A)^c \wedge \tau(B)) \vee (\tau(A)\wedge \tau(B)^c)]}{\P[\tau(A)^c \vee \tau(B)^c ]}\\
&=& \frac{(1-\P[\tau(A)])\P[\tau(B)] + \P[\tau(A)](1-\P[\tau(B)])}{1-\P[\tau(A)]\P[\tau(B)]}.
\end{eqnarray*}

Now, note
\begin{eqnarray*}
(1-\P[\tau(A)])\P[\tau(B)] &=& \frac{1}{C} p(1-p)|B|\\
\P[\tau(A)](1-\P[\tau(B)]) &=& \frac{1}{C}p(1-p)|A|
\end{eqnarray*}
and
\begin{eqnarray*}
1-\P[\tau(A)]\P[\tau(B)]
&=& \frac{1}{C}(1-p)(1-p - (|A|+|B|)p)
\end{eqnarray*}
where $C : = (1-p + |A|p)(1-p+|B|p)$.

Hence,
\begin{eqnarray*}
\P[\tau(A\cup B) \mid \tau(A)^c \vee \tau(B)^c]
= \frac{p(|A|+|B|)}{1-p+(|A|+|B|)p}
\end{eqnarray*}
which shows that (\ref{equ:pa1}) holds for $A\cup B$.

%xxx
%
%For any given integers $R$ and $T$ such that $R\leq T$, consider the Markov chain $\{(X_t, A_t)\}_{t=0}^T$, with initial value $(X_0,A_0) = (R,0)$, and transitions defined by, for $0\leq t < T$,
%\begin{eqnarray*}
%X_{t+1} &=& X_t - \xi_t\\
%A_{t+1} & = & A_t + X_t(\alpha \xi_t + \beta (1-\xi_t))
%\end{eqnarray*}
%where $\xi_t$ are independent and identically distributed Bernoulli random variables with mean $\gamma$, and $\alpha,\beta$ are positive constants.

%\begin{lemma} For the Markov chain $\{(X_t,A_t)\}_{t=0}^T$, we have
%$$
%\E[A_T] = T\left(R-\frac{1}{2}\gamma(T-1)\right)((1-\gamma)\alpha + \gamma \beta).
%$$
%\end{lemma}

%Note that
%$$
%A_T = \sum_{t=0}^{T-1}X_t(\alpha (1-\xi_t) + \beta \xi_t). 
%$$
%Hence, 
%$$
%\E[A_T] = \alpha \sum_{t=0}^{T-1}\E[(1-\xi_t)X_t] + \beta\sum_{t=0}^{T-1}\E[X_t\xi_t].
%$$

%Now, note, for $0\leq t \leq T$,
%$$
%X_t = R - \sum_{s=0}^{t-1} \xi_s.
%$$

%We have
%\begin{eqnarray*}
%\E[X_t\xi_t] &=& R\gamma - \sum_{s=0}^{t-1}\E[\xi_s \xi_t]
%= \gamma(R - t\gamma)
%\end{eqnarray*}
%and
%\begin{eqnarray*}
%\E[X_t(1-\xi_t)] &=& \E[X_t] - \E[X_t \xi_t]\\
%&=& (R-t\gamma) -(R\gamma-t\gamma^2)\\
%&=& (1-\gamma)R - t \gamma(1-\gamma)\\
%&=& (1-\gamma)(R-t\gamma). 
%\end{eqnarray*}

%Hence,
%$$
%\sum_{t=0}^{T-1}\E[\xi_t X_t] = \gamma\left( RT - \frac{1}{2}\gamma T(T-1)\right) 
%$$
%and
%$$
%\sum_{t=0}^{T-1}\E[(1-\xi_t)X_t] = (1-\gamma)\left(RT - \frac{1}{2}\gamma T(T-1)\right).
%$$
%
%It follows
%$$
%\E[A_T] = T\left(R-\frac{1}{2}\gamma(T-1)\right)((1-\gamma)\alpha + \gamma \beta).
%$$

\subsubsection{Proof of Proposition \ref{pro:epochs}}
\label{sec::greedy-bayes:min:theorem-1}
We break $\tau$ into two parts: (a) $\tau_1$ denoting the number of steps spent examining pairs of regular matching sets, and (b) $\tau_2$ denoting the number of steps spent for examining a pair of matching sets involving a special set. 

We first upper bound the expected value of $\tau_1$. Conditional on $X_s$, $X_{s+1}$ is a binomial random variable with parameters $\lfloor X_s/2\rfloor$ and $1-\pp(2^s)^2$. Hence, for $s \geq 1$,
\begin{equation}
\E[X_{s+1}] = \E\left[\left\lfloor\frac{X_s}{2}\right\rfloor \right] (1-\pp(2^{s})^2)
\label{equ:exs}
\end{equation}
and
\begin{equation}
\E[X_1] = \frac{n}{2}(1-\pp(1)^2).
\label{equ:iexs}
\end{equation}

From (\ref{equ:exs}), we have
$$
\E[X_{s+1}] \leq \E[X_s] \frac{1-\pp(2^s)^2}{2}, \hbox{ for } s\geq 1
$$
which together with (\ref{equ:iexs}) yields
\begin{equation}
\E[X_s] \leq n\frac{1}{2^s}\prod_{i=0}^{s-1}(1-\pp(2^i)^2).
\label{equ:exs-1}
\end{equation}

Since in each super-epoch $s$ exploration of pairs from any two regular matching sets takes at most $2^{s}2^{s} = 4^s$ steps and there are at most $X_s/2$ pairs of matching sets, and there are at most $\log_2(n)$ super-epochs, we have 
\begin{equation}
\E[\tau_1] \leq \frac{1}{2}\sum_{s=1}^{\log_2(n)} 4^s\E[X_s].
\label{equ:etau1}
\end{equation}

From (\ref{equ:exs}), $\E[X_s]\leq n 2^{-s}$, hence it follows
$$
\E[\tau_1] \leq n^2.
$$

Let us define $s^* = \left\lfloor\log_2\left(c/p\right) \right\rfloor$, where $c$ is some constant such that $c > 2/(\sqrt{2}-1)$. 

From (\ref{equ:exs}) and (\ref{equ:etau1}), we have
$$
\E[\tau_1] \leq n(I(1,s^*) + I(s^*,\infty))
$$
where 
$$
I(x,y) := \sum_{s=x}^{y-1} 2^s\prod_{i=0}^{s-1}(1-\pp(2^i)^2).
$$

We first bound the term $I(1,s^*)$ as follows
$$
I(1,s^*) \leq  \sum_{s=0}^{s^*-1} 2^s \leq 2^{s^*} \leq \frac{c}{p}.
$$

We next bound the term $I(s^*,\infty)$. By definition of $s^*$, note that $s^* \geq \log_2(c/p) -1$, hence $p2^{s^*} \geq c/2$. It is readily observed that this with $c > 2/(\sqrt{2}-1)$ implies $2(1-\pp(2^{s^*})^2) < 1$.

We can write
\begin{eqnarray*}
I(s^*,\infty) &\leq & \sum_{s=s^*}^\infty (2(1-\pp(2^{s^*})^2))^s\\
&=& \frac{1}{2\pp(2^{s^*})^2-1}\left(2(1-\pp(2^{s^*})^2)\right)^{s^*}\\
&\leq & \frac{1}{2\pp(2^{s^*})^2-1}\\
& \leq & \frac{1}{2 (c/(c+2))^2 - 1} \\
& = & O(1)
\end{eqnarray*}
where in the last inequality we use the fact that $\pi(a)$ is increasing in $a$, and $2^{s^*}\geq c/(2p)$, hence $\pi(2^{s^*})\geq (c/2)/(1-p+c/2) \geq c / (c+2)$.

Therefore, we have
\begin{equation}
    \label{MIN-tau1-bound}
    \frac{1}{2}\sum_{s=1}^{\log_2(n)} 4^s\E[X_s] \leq \left(\frac{1}{2 (c/(c+2))^2 - 1} + \frac{c}{p}\right) n.
\end{equation}

Hence, it follows
$$
\E[\tau_1] = O\left(\min\left\{\frac{n}{p},n^2\right\}\right).
$$

We next consider the number of steps $\tau_2$ spent on exploring special matching sets. We have already noted that in each super-epoch $s$, the cardinality of the special set is at most $2^{s+1}$. The special matching set can be paired with a set of cardinality at most $2^{s+1}$. Hence, in super-epoch $s$, the expected number of steps spent on examining matching sets involving the special matching set is at most $2^{s+1} 2^{s+1} = 4^{s+1}$. Now, note
\begin{eqnarray*}
\E[\tau_2] &\leq & 4\sum_{s=1}^{\log_2(n)} 4^{s}\Pr[X_s\geq 1]\\
&\leq & 4\sum_{s=1}^{\log_2(n)} 4^{s}\E[X_s]
\end{eqnarray*}
where the second inequality is by Markov's inequality. We have already shown that 
\begin{equation}
    \label{greedy-bayes-tau-1}
    \sum_{s=1}^{\log_2(n)} 4^{s}\E[X_s] = O(\min\{n/p,n^2\})
\end{equation}
, hence, the result follows.

\subsubsection{Proof of Theorem~\ref{thm:min-reg}}
\label{proof-min-reg}

Let $X_s$ be the number of regular matching sets and let $Y_s$ be the size of the special matching set at the beginning of super-epoch $s$. Let $Z_r$ be a random variable taking value 1 if both of the two matching sets examined in epoch $r$ contain a type 1 node, and $Z_r = 0$, otherwise.

Note that conditional on that in epoch $r$ of super-epoch $s$ two regular matching sets are examined, we have $Z_r \sim \operatorname{Ber}(\pp(2^s)^2)$, and if the examination involves the special matching set, we have $Z_r \sim \operatorname{Ber}(\pp(Y_s)\pp(\xi_s))$, where $\xi_s=2^s$ if $X_s$ is odd, and $\xi_s = 2^{s+1}$ if $X_s$ is even and $X_{s+1}>0$.

\begin{figure}
    \centering
    \includegraphics[width=.8\textwidth]{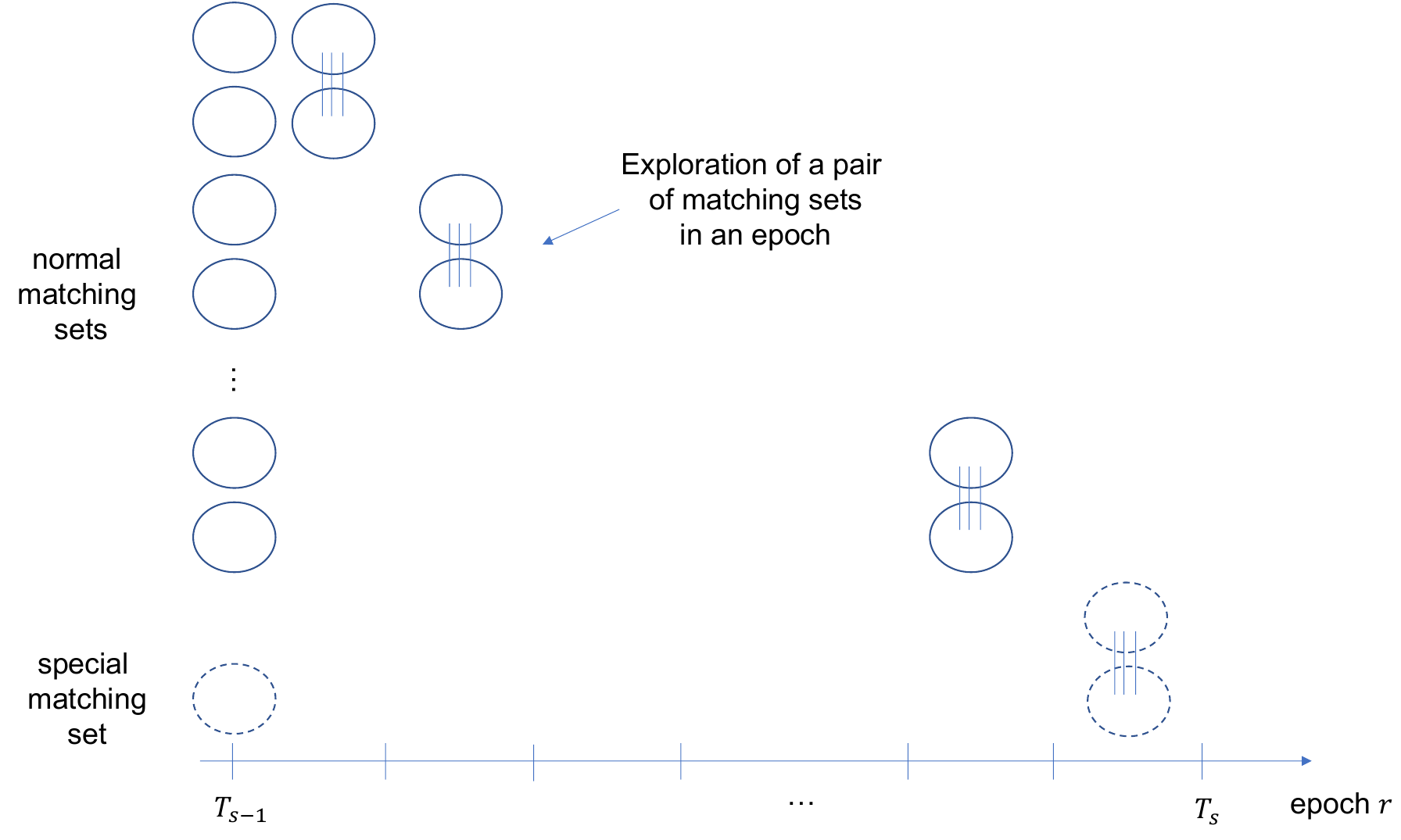}
    \caption{Matchings created in super-epoch $s$.}
    \label{fig:lms-algo}
\end{figure}

Let $S_r$ be the number of steps in epoch $r$, $T_{s-1}$ be the first epoch of super-epoch $s$ with $T_0=1$. 
% \mv{*** Double check this -- at epoch $1$ matching is according to random matching and only after observing outcomes under random matching, we can define matching sets. So, it should be $T_0 = 1$ instead? ***}
Let $\tilde{R}_s$ denote the regret accumulated in super-epoch $s$.
To simplify notation, we define $K_s := \lfloor X_s/2\rfloor$, which corresponds to the number of epochs in which two regular matching sets are explored in super-epoch $s$. We also use the following notation $p_s := \pp(2^s)$ and $q_s := 1 - p^2_s$. We show a graphic illustration of various definitions in Figure~\ref{fig:lms-algo}.

Note that there can be at most $\log_2(n)$ super-epochs.

Let $X^1_s$ be the number of regular matching sets at the beginning of super-epoch $s$ that contain a type 1 node.
Let $Y^1_s$ be a random variable taking value 1 if the special matching set has a type 1 node, and otherwise $Y^1_s = 0$.
If $Y_s=0$ (indicating that at the beginning of super-epoch $s$ there is no special matching set), we define $Y^1_s = 0$.
Note that $X^1_s \mid X_s \sim \operatorname{Bin}(X_s, p_s)$ and $Y^1_s \mid Y_s \sim \operatorname{Ber}(\pp(Y_s))$.

Let $\alpha_s$ (resp. $\beta_s$) be the conditional expected number of steps in epoch $r$, conditional on $Z_r = 0$ (resp. $Z_r=1$), where $s$ is the super-epoch of epoch $r$, i.e.
$\alpha_s = \E\left[S_r | Z_r=0\right]$ and $\beta_s = \E\left[S_r | Z_r=1\right]$. Note that $\beta_s = \alpha_s / 2$.

In each step, the instantaneous regret is equal to the number of pairs of type 1 nodes that can be formed from the number of unidentified type 1 nodes. Since each regular or special matching set contains at most one type 1 node, we can express the instantaneous regret at the beginning of super-epoch $s$ as follows:
\begin{equation}
    \label{eq:R_super-epoch}
    R_{T_{s-1}} = \Bigl\lfloor \frac{X^1_s}{2} \Bigr\rfloor + Y^1_s \1_{\{X^1_s \text{ is odd}\}}.
\end{equation}
The instantaneous regret in epoch $r$, such that $T_{s-1}<r\leq T_s$, satisfies the following relation
\begin{equation}
    \label{eq:R_r}
    R_r = R_{T_{s-1}} - \sum_{i=T_{s-1}+1}^r Z_i.
\end{equation}
We will evaluate the cumulative regret of the algorithm by summing up the regret accumulated in each super-epoch. 
The regret accumulated in super-epoch $s$ can be expressed as follows:
$$
\tilde{R}_s = A_s + B_s
$$
where 
\begin{equation}
    A_s := \sum_{r=T_{s-1}+1}^{T_{s-1}+K_s} R_{r-1} S_r \label{eq:A_s}
\end{equation}
and
\begin{equation}
B_s := R_{T_{s-1}+K_s} S_{T_{s-1}+K_s+1}\left(\1_{\{X_s \text{ is odd}\}} + \1_{\{Y_s > 0\}} \1_{\{X_s \text{ is even}\}} + \1_{\{X_{s+1} > 0\}}  \right)\1_{\{X_s>0\}}. \label{eq:B_s}
\end{equation}
Here $A_s$ is the cumulative regret incurred in epochs of super-epoch $s$ in  which regular matching sets are examined, and $B_s$ is the regret incurred in the last epoch of super-epoch $s$ that involves examination of the special matching set, or a regular matching set of size $2^{s+1}$ and a remaining regular matching set of size $2^s$ if there is odd number of regular matching sets of size $2^s$ at the initial epoch of the super-epoch $s$.

We next note some basic properties of binomial distribution, which will be used in the rest of the proof.

\begin{lemma} Assume that $X$ is a binomial random variable with parameters $n$ and $p$. Then,
\begin{enumerate}
    \item $\E\left[X^{2}\right]=n(n-1) p^{2}+n p$
    \item $\E\left[X^{3}\right]=n(n-1)(n-2) p^{3}+3 n(n-1) p^{2}+n p$.
\end{enumerate}
\label{lm:bino}
\end{lemma}

In the following we evaluate the expected values of $A_s$ and $B_s$ defined in (\ref{eq:A_s}) and (\ref{eq:B_s}), respectively.

\begin{lemma}
For every super-epoch $1\leq s\leq \log_2(n)$, we have
$$
\E[A_s] = \frac{1}{4} \left(1- \frac{1}{2}p_s\right) \left(1-\frac{1}{2}p_s^2\right) p_s\alpha_s\E[X_s^2] + O(n).
$$
\end{lemma}

\begin{proof}
From (\ref{eq:R_r}) and (\ref{eq:A_s}), $A_s = A^{(1)}_s - A^{(2)}_s$ where
\begin{equation}
A^{(1)}_s = R_{T_{s-1}} \sum_{r=T_{s-1}+1}^{T_{s-1}+K_s} S_r
\end{equation}
and
\begin{equation}
A^{(2)}_s := \sum_{r=T_{s-1}+1}^{T_{s-1}+K_s} S_r \sum_{i=T_{s-1}+1}^r Z_i.
\end{equation}

We separately upper bound the expected values of $A^{(1)}_s$ and $A^{(2)}_s$. First, we upper bound the expected value of $A_s^{(1)}$. Note that
\begin{eqnarray*}
\E\left[A_{s}^{(1)} \mid T_{s-1}, X_{s}, X_{s}^{1}, Y_{s}^{1}\right] &=& \E\left[R_{T_{s-1}} \mid X_{s}, X_{s}^{1}, Y_{s}^{1}\right]
N_s
\end{eqnarray*}
where 
$$
N_s := \E\left[\sum_{r=T_{s-1}+1}^{T_{s-1}+K_{s}} S_{r} \mid T_{s-1}, X_{s}, X_{s}^{1}\right].
$$

Combining with (\ref{eq:R_super-epoch}), we have 
\begin{equation}
    \E\left[A^{(1)}_s \mid T_{s-1} X_{s}, X_{s}^{1}, Y_{s}^{1}\right]=\left( \left\lfloor\frac{X_{s}^{1}}{2}\right\rfloor+Y_{s}^{1} \1_{\{X^1_s \text{ is odd}\}} \right) N_s.
\label{equ:As1}
\end{equation}

We next evaluate $N_{s}$. Let us define $K_{s}^{11}$ to be the number of epochs in super-epoch $s$ in which two regular matching sets are examined such that both have a value 1 node, i.e. $K_{s}^{11}:=$ $\sum_{r=T_{s-1}+1}^{T_{s-1}+K_{s}} Z_{r} .$ Then, note
$$
N_{s} = \beta_{s} \E\left[K_{s}^{11} \mid X_{s}, X_{s}^{1}\right]+\alpha_{s} \E\left[K_{s}-K_{s}^{11} \mid X_{s}, X_{s}^{1}\right] 
= \alpha_s \left(K_s - \frac{1}{2} \E[K^{11}_s \mid X_s, X^1_s]\right).
$$

%This requires evaluating the expected number of pairs of 1-1 matching sets, given that there are $X_s$ matching sets in which $X^1_s$ matching sets contain a type 1 node.

Now, note $\E\left[K^{11}_s \mid X_s, X^1_s\right] = K_s (X^1_s/X_s)^2$ and, hence, 
$$
N_s = \alpha_s K_s \left(1 - \frac{1}{2}  \left(\frac{X^1_s}{X_s}\right)^2\right) = \alpha_s \left\lfloor \frac{X_s}{2} \right\rfloor \left(1 - \frac{1}{2}  \left(\frac{X^1_s}{X_s}\right)^2\right).
$$
Note that 
\begin{equation*}
\begin{aligned}
\E\left[ \left\lfloor\frac{X_{s}^{1}}{2}\right\rfloor \left\lfloor \frac{X_s}{2} \right\rfloor \left(1 - \frac{1}{2}  \left(\frac{X^1_s}{X_s}\right)^2\right) \mid X_{s}\right] 
&= \frac{X_s}{4} \E\left[ X^1_s - \frac{(X^1_s)^3}{2X_s^2} \mid X_s \right]\\
&= \frac{X_s \E[X^1_s \mid X_s]}{4} - \frac{\E[(X^1_s)^3 \mid X_s]}{8X_s} \\
&= \frac{1}{4}p_s X_s^2 - \frac{1}{8}p_s^3 X_s^2 + O(n)\\
&= \frac{1}{4}p_s\left(1-\frac{1}{2}p_s^2\right)X_s^2 + O(n)
\end{aligned}
\end{equation*}
where we used Lemma~\ref{lm:bino}.

Combining with (\ref{equ:As1}) we have
$$
\E\left[A_{s}^{(1)} \right]
= \frac{1}{4} p_s \left(1- \frac{1}{2}p_s^2\right) \alpha_s\E[X_s^2] + O(n).
$$

Next, we upper bound the expected value of $A_s^{(2)}$. Note that
\begin{equation*}
\begin{aligned}
\E\left[A_{s}^{(2)}\right] &=\E\left[\sum_{r=T_{s-1}+1}^{T_{s-1}+K_{s}} S_{r} \sum_{i=T_{s-1}+1}^{r} Z_{i}\right] \\
&=\E\left[\E\left[\sum_{r=T_{s-1}+1}^{T_{s-1}+K_{s}} S_{r} \sum_{i=T_{s-1}+1}^{r} Z_{i} \mid T_{s-1}, K_{s}\right]\right] \\
&= \E\left[\sum_{r=T_{s-1}+1}^{T_{s-1}+K_{s}} \E\left[\left(\alpha_{s}\left(1-Z_{r}\right)+\beta_{s} Z_{r}\right) \sum_{i=T_{s-1}+1}^{r} Z_{i} \mid T_{s-1}, K_{s}\right]\right] \\
& = \E\left[\sum_{r=T_{s-1}+1}^{T_{s-1}+K_{s}} \E\left[ \alpha_s \sum_{i=T_{s-1}+1}^{r} Z_i - \frac{1}{2}\alpha_s Z_r \sum_{i=T_{s-1}+1}^{r} Z_i \mid T_{s-1}, K_{s}\right]\right]
\end{aligned}
\end{equation*}
where the last equality holds because $\beta_s = \alpha_s / 2$.

Now, note
\begin{equation*}
\begin{aligned}
\E\left[ \sum_{i=T_{s-1}+1}^{r} Z_i \mid T_{s-1}, K_{s}\right]
&= \sum_{i=T_{s-1}+1}^{r} \P\left[ Z_i = 1 \right] = (r - T_{s-1}) p_s^2
\end{aligned}
\end{equation*}
and
\begin{equation*}
\begin{aligned}
\E\left[ Z_r \sum_{i=T_{s-1}+1}^{r} Z_i \mid T_{s-1}, K_{s}\right]
&= \E\left[ Z_r^2 \mid T_{s-1}, K_{s}\right] + \sum_{i=T_{s-1}+1}^{r-1} \E\left[ Z_r Z_i \mid T_{s-1}, K_{s}\right] \\
&= \P\left[ Z_i = 1 \right] + \sum_{i=T_{s-1}+1}^{r-1} \P\left[ Z_r = 1 \right] \P\left[ Z_i = 1 \right] \\
&= p_s^2 + (r - T_{s-1} - 1) p_s^4.
\end{aligned}
\end{equation*}
Putting the pieces together, we obtain
\begin{equation*}
\begin{aligned}
\E\left[A_{s}^{(2)}\right] &=\E\left[\sum_{r=T_{s-1}+1}^{T_{s-1}+K_{s}} \alpha_s (r - T_{s-1}) p_s^2 - \frac{1}{2}\alpha_s \left[ p_s^2 + (r - T_{s-1} - 1) p_s^4 \right] \right] \\
&= \E\left[\sum_{r=T_{s-1}+1}^{T_{s-1}+K_{s}} \alpha_s p_s^2(1-\frac{1}{2}p_s^2)(r-T_{s-1}) - \frac{1}{2}\alpha_s p_s^2(1-p_s^2) \right] \\
&= \E\left[ \alpha_s p_s^2(1-\frac{1}{2}p_s^2) \sum_{i=1}^{K_s} i -  \frac{1}{2}\alpha_s p_s^2(1-p_s^2) K_s \right] \\
&= \E\left[ \frac{1}{2}\alpha_s p_s^2(1-p_s^2) K_s^2 +  \frac{1}{4}\alpha_s p_s^4 K_s(K_s + 1) \right] \\
&= \frac{1}{8} p_s^2 \left(1-\frac{1}{2}p_s^2\right) \alpha_s \E[X_s^2] + O(n)
\end{aligned}
\end{equation*}
where the last equality holds because $K_s = \lfloor X_s / 2 \rfloor$.

Using the upper bounds for $\E[A^{(1)}_s]$ and $\E[A^{(2)}_s]$, we have 
$$
\E[A_s] = \frac{1}{4} \left(1- \frac{1}{2}p_s\right) \left(1-\frac{1}{2}p_s^2\right) p_s\alpha_s\E[X_s^2] + O(n).
$$
\end{proof}

We next show an upper bound for the expected value of $B_s$.

\begin{lemma}
For every super-epoch $1\leq s\leq \log_2(n)$, we have
$$
\E[B_{s}] \leq n p_{s} 2^{s+1}  \prod_{i=0}^{s-1} q_{i}+4^{s+1} \P\left[X_s > 0\right].
$$
\end{lemma}

\begin{proof} 
In the last epoch of super-epoch $s$, the two matching sets are of cardinality at least $2^s$ and at most $2^{s+1}$, hence, from (\ref{eq:B_s}), we have
\begin{equation}
B_{s} \leq R_{T_{s-1}+K_{s}} \1_{\{X_{s} > 0\}} 4^{s+1}.
\label{equ:BR}
\end{equation}

From (\ref{eq:R_super-epoch}), it follows that $R_{T_{s-1}+K_s} \leq R_{T_{s-1}} \leq \lfloor X_s^1 / 2 \rfloor + 1$ where recall $X^1_s\mid X_s \sim \operatorname{Bin}(X_s, p_s)$, and hence
$$
\E[R_{T_{s-1}+K_s}\mid X_s = x] \leq \frac{1}{2}p_s x + 1.
$$
Now, note
\begin{eqnarray*}
\E[R_{T_{s-1}+K_s}\mid X_s > 0] &=& \frac{\E[R_{T_{s-1}+K_s} \1_{\{X_s > 0\}}]}{\P[X_s > 0]}\\
&=& \frac{\sum_{x=1}^\infty \E[R_{T_{s-1}+K_s}\mid X_s = x]\P[X_s = x] }{\P[X_s > 0]}\\
&\leq & \frac{1}{2}p_s \frac{\sum_{x=1}^\infty x \P[X_s = x]}{\P[X_s > 0]} + 1\\
&= & \frac{1}{2}p_s \frac{\E[X_s \1_{\{X_s > 0\}}]}{\P[X_s > 0]} + 1\\
&=& \frac{1}{2}p_s \frac{\E[X_s]}{\P[X_s > 0]} + 1.
\end{eqnarray*}

Hence, we have
$$
\E[R_{T_{s-1}+K_s}\1_{\{X_s > 0\}}] \leq \frac{1}{2}p_s \E[X_s] + \P[X_s > 0]
$$
which combined with $\E[X_s]\leq n(1/2^s)\prod_{i=0}^{s-1}q_i$ that follows from (\ref{equ:X}), yields
$$
\E[R_{T_{s-1}+K_s}\1_{\{X_s > 0\}}] \leq n p_s\frac{1}{2^{s+1}} \prod_{i=0}^{s-1}q_i + \P[X_s > 0].
$$
Combining with (\ref{equ:BR}), we have
$$
\E\left[B_{s} \right] \leq n p_s 2^{s+1} \prod_{i=0}^{s-1}q_i + 4^{s+1}\P[X_s > 0].
$$

% It follows that
% \begin{align*}
%     \E\left[R_{T_{s-1}+K_{s}}\1_{\{X_{s} > 0\}}\right] 
%     &=  \E\left[\E\left[R_{T_{s-1}+K_{s}} \mid X_s > 0\right]\1_{\{X_{s} > 0\}}\right] \\
%     &\leq n p_{\mathrm{s}} \frac{1}{2^{s+1}} \prod_{i=0}^{s-1} q_{i} \P\left[X_s > 0\right] + \P\left[X_s > 0\right] \\
%     &\leq n p_{\mathrm{s}} \frac{1}{2^{s+1}} \prod_{i=0}^{s-1} q_{i}+\P\left[X_s > 0\right]
% \end{align*}
% % Note that for any super-epoch $s$, $X_s \geq 1$ holds. Otherwise, there exists at most one matching set and the algorithm stops.
% and
% $$
% \E\left[B_{s} \1_{\{X_{s} > 0\}} \right] \leq n 2^{s+1}  p_{s}\left(1-p_{s}\right) \prod_{i=0}^{s-1} q_{i}+4^{s+1} \P\left[X_s > 0\right].
% $$
\end{proof}
%\endproof\Halmos

Now, note by Markov's inequality and (\ref{MIN-tau1-bound}), for a constant $c > 2/(\sqrt{2}-1)$,
$$
\sum_{s=1}^{\log_2(n)} 4^{s+1} \P[X_s > 0]\leq \sum_{s=1}^{\log_2(n)} 4^{s+1} \E[X_s] \leq 8\left(\frac{1}{2(c/(c+2))^2-1} + \frac{c}{p}\right)n
$$
and, obviously, we have
$$
\sum_{s=1}^{\log_2(n)} 4^{s+1} \P[X_s > 0] \leq \sum_{s=1}^{\log_2(n)} 4^{s+1} \leq \frac{4}{3}n^2.
$$

Putting the pieces together, we have 
$$
\E[R] \leq \sum_{s=1}^{\log _2(n)} \E[A_s + B_s] \leq a_{n, p} n^2 + b_{n,p} n^2 + c_{n,p} n^2 + O(n \log(n))
$$
where
$$
a_{n, p}=\frac{1}{4} \sum_{s=1}^{\log _{2}(n)} \left(1- \frac{1}{2}p_s\right) \left(1-\frac{1}{2}p_s^2\right) p_s\left(\prod_{i=0}^{s-1} q_{i}^{2}+\frac{1}{n} \sum_{i=0}^{s-1} 2^{i+1}\left(\prod_{j=0}^{i} q_{j}\right)\left(\prod_{j=i+1}^{s-1} q_{j}^{2}\right)\right),
$$
$$
b_{n,p}= \frac{1}{n} \sum_{s=1}^{\log _{2}(n)}  2^{s+1}  p_{s} \prod_{i=0}^{s-1} q_{i},
$$
and
$$
c_{n,p} = \min\left\{\frac{c_1}{np},\frac{4}{3}\right\}
%c_{n,p} = \frac{c_1}{np} + \frac{8}{2(c_1/(c_1+16))^2-1}\frac{1}{n} 
%c_{n,p} = 8 \min\left\{ \frac{\lambda}{np}, 1 \right\}
$$
where $c_1$ is a constant such that $c_1 > 16 / (\sqrt{2}-1)$.

\subsubsection{Properties of $a_{n,p}$, $b_{n,p}$ and $c_{n,p}$} 
\label{greedy-Bayes-anp-bnp}

We consider properties of $a_{n,p}$, $b_{n,p}$ and $c_{n,p}$ that appear in the regret upper bound in Theorem~\ref{thm:min-reg}. For $c_{n,p}$ we note that $c_{n,p} = o(1)$ whenever $p = \omega(1/n)$. In other words, $c_{n,p}$ diminishes with $n$ whenever $p$ is a constant or it goes to $0$ as $\omega(1/n)$. In the following, we focus on showing some properties for $a_{n,p}$ and $b_{n,p}$. 

Let $a_{n,p}'$ and $a_{n,p}''$ be the two terms in $a_{n,p}$, i.e.
\begin{equation}
a_{n,p}' := \frac{1}{4} \sum_{s=1}^{\log _{2}(n)} \left(1- \frac{1}{2}p_s\right) \left(1-\frac{1}{2}p_s^2\right) p_s \prod_{i=0}^{s-1} q_{i}^{2}
\label{equ:a1}
\end{equation}
and
\begin{equation}
a_{n,p}'' := \frac{1}{4}\sum_{s=1}^{\log_2(n)} \left(1-\frac{1}{2}p_s\right)\left(1-\frac{1}{2}p_s^2\right)p_s \frac{1}{n}\sum_{i=0}^{s-1} 2^{i+1}\left(\prod_{j=0}^{i}q_j\right)\left(\prod_{j=i+1}^{s-1}q_j^2\right).
\label{equ:a2}
\end{equation}
%and, recall that
%$$
%\jy{b_{n,p} = \frac{1}{n}\sum_{s=1}^{\log_2(n)} 2^{s+1} p_s\prod_{i=0}^{s-1}q_i.}
%$$

We will use the following basic facts. Note that 
$$
p_i = \frac{p2^i}{1-p+p 2^i} \geq 1- \frac{1-p}{p}2^{-i}.
$$ 
The right-had side in the last inequality is positive if and only if $i\geq \log_2((1-p)/p))$. The latter condition holds for all $i\geq 0$ provided that $1/2\leq p\leq 1$.

It follows that for every $i\geq \log_2((1-p)/p)$, we have
$$
q_i = 1-p_i^2 \leq 2 \frac{1-p}{p}2^{-i}.
$$ 

We next consider properties of $a_{n,p}$ and $b_{n,p}$, for three different cases depending on the value of parameter $p$.

\paragraph{Case $1/2 < p < 1$} From (\ref{equ:a1}) we have 
$$
a_{n,p}' \leq \frac{1}{4}\sum_{s=1}^{\log_2(n)} \prod_{i=0}^{s-1}q_i^2
$$
and, note that
$$
\prod_{i=0}^{s-1} q_i^2 
\leq \prod_{i=0}^{s-1} \left(2\frac{1-p}{p}\right)^2 4^{-i} = \left(8\left(\frac{1-p}{p}\right)^2\right)^{s} 2^{-s^2}.
$$

Hence, it follows 
\begin{equation}
a_{n,p}' \leq \frac{1}{4}\sum_{s=1}^{\log_2(n)} \left(8\left(\frac{1-p}{p}\right)^2\right)^{s} 2^{-s^2}.\label{equ:cnp1}
\end{equation}

%The following holds
%\begin{eqnarray*}
%\sum_{s=1}^{\log_2(n)} \left(8\left(\frac{1-p}{p}\right)^2\right)^{s} 2^{-s^2} &\leq & \sum_{s=1}^{\log_2(n)} \left(\frac{1}{2^{s-3}}\left(\frac{1-p}{p}\right)^2\right)^{s}\\
%& \leq & 4 \left(\frac{1-p}{p}\right)^2 + 4 \left(\frac{1-p}{p}\right)^4 + \sum_{s=3}^{\log_2(n)} \left(\frac{1}{2^{s-3}}\left(\frac{1-p}{p}\right)^2\right)^{s}\\
%&\leq & 4 \left(\frac{1-p}{p}\right)^2 + 4 \left(\frac{1-p}{p}\right)^4 + \min\left\{\frac{\left(\frac{1-p}{p}\right)^2}{1-\left(\frac{1-p}{p}\right)^2} , 2\right \}.
%\end{eqnarray*}
%
%Thus, we have
%$$
%a_{n,p}' \leq  \left(\frac{1-p}{p}\right)^2 + \left(\frac{1-p}{p}\right)^4 + \min\left\{\frac{1}{4}\frac{\left(\frac{1-p}{p}\right)^2}{1-\left(\frac{1-p}{p}\right)^2}, \frac{1}{2}\right\}.
%$$
%Note that the right hand side is equal to $(4/3)(1-p)^2$ asymptotically for small $1-p$. 

%We can obtain a tighter upper bound as follows.

We use the following facts to bound the sum in (\ref{equ:cnp1}). By straightforward calculus, we have 
\begin{equation}
\int_a^b e^{cx}e^{-dx^2} dx = \frac{\sqrt{\pi}}{\sqrt{d}}e^{\frac{c^2}{4d}}\left[\Phi\left(\sqrt{2d}\left(b-\frac{c}{2d}\right)\right) -\Phi\left(\sqrt{2d}\left(a-\frac{c}{2d}\right)\right)\right]
\label{equ:int}
\end{equation} 
where $\Phi(x)$ is the cumulative distribution function of a normal random variable.  

It follows
$$
\int_a^b e^{cx}e^{-dx^2} dx \leq \frac{\sqrt{\pi}}{\sqrt{d}}e^{\frac{c^2}{4d}}\Phi^c\left(\sqrt{2d}\left(a-\frac{c}{2d}\right)\right)
$$
where $\Phi^c(x) = 1-\Phi(x)$.

The following is a well known bound:
$$
\Phi^c(x) \leq \frac{1}{\sqrt{2\pi}}\frac{1}{x}e^{-\frac{x^2}{2}} \hbox{ for } x > 0.
$$

If $a - c/(2d) > 0$, we have
\begin{equation}
\int_a^b e^{cx}e^{-dx^2} dx \leq \frac{1}{2d}\frac{1}{a-\frac{c}{2d}}e^{-(da^2 - ca)}.
\label{equ:intb}
\end{equation}

Note that
\begin{eqnarray*}
\sum_{s=1}^{\log_2(n)} \left(8\left(\frac{1-p}{p}\right)^2\right)^{s} 2^{-s^2} &\leq & 4 \left(\frac{1-p}{p}\right)^2 + 4 \left(\frac{1-p}{p}\right)^4 + \left(\frac{1-p}{p}\right)^{6}\\
&& + \sum_{s=4}^{\log_2(n)} \left(\left(\frac{1-p}{p}\right)^2\right)^{s} 2^{-\frac{1}{4}s^2}.
\end{eqnarray*}

Now, by applying (\ref{equ:intb}) with $c = \log((1-p)/p)$ and $d = (1/4)\log(2)$, we have

\begin{eqnarray*} 
\sum_{s=4}^{\log_2(n)} \left(\left(\frac{1-p}{p}\right)^2\right)^{s} 2^{-\frac{1}{4}s^2}
&\leq & \int_{3}^{\log_2(n)-1} \left(\left(\frac{1-p}{p}\right)^2\right)^{x} 2^{-\frac{1}{2}x^2}dx\\
& \leq & \frac{2}{\log(2)}\frac{1}{3-4\log_2\left(\frac{1-p}{p}\right)} \\
& & \exp\left(-\left(\frac{9}{4}\log(2)-6\log(2)\log_2\left(\frac{1-p}{p}\right)\right)\right)\\
&=& \frac{1}{2^{5/4}\log(2)}\frac{1}{3+4\log_2\left(\frac{p}{1-p}\right)}\left(\frac{1-p}{p}\right)^6\\
 &\leq & \frac{1}{3\cdot 2^{5/4}\log(2)}\left(\frac{1-p}{p}\right)^6.
\end{eqnarray*} 

We have shown that
\begin{eqnarray*}
\sum_{s=1}^{\log_2(n)} \left(8\left(\frac{1-p}{p}\right)^2\right)^{s} 2^{-s^2} \leq 4 \left(\frac{1-p}{p}\right)^2 + 4 \left(\frac{1-p}{p}\right)^4 + \left(1+\frac{1}{3\cdot 2^{5/4}\log(2)}\right)\left(\frac{1-p}{p}\right)^{6}.
\end{eqnarray*}

Thus, we have
$$
a_{n,p}' \leq  \left(\frac{1-p}{p}\right)^2 + \left(\frac{1-p}{p}\right)^4 + \frac{1}{4}\left(1+\frac{1}{3\cdot 2^{5/4}\log(2)}\right)\left(\frac{1-p}{p}\right)^{6}.
$$

From this, observe that $a_{n,p}' = O((1-p)^2)$.

%Observe that the right-hand side is $(1-p)^2$ asymptotically for small $1-p$.

Next, we consider $a_{n,p}''$ and $b_{n,p}$. Using (\ref{equ:a2}), we have 
\begin{eqnarray*}
a_{n,p}'' &\leq & \frac{1}{4}\sum_{s=1}^{\log_2(n)} \frac{1}{n}\sum_{i=0}^{s-1} 2^{i+1}\left(\prod_{j=0}^{i}q_j\right)\left(\prod_{j=i+1}^{s-1}q_j^2\right)\\
&\leq & \frac{1}{4}\sum_{s=1}^{\log_2(n)} \frac{1}{n}\sum_{i=0}^{s-1} 2^{i+1}q_{i+1}\left(\prod_{j=0}^{s-1}q_j\right)\\
&\leq & \frac{1}{4}\sum_{s=1}^{\log_2(n)} \frac{1}{n}\sum_{i=0}^{s-1} 2^{i+1}\left(2\frac{1-p}{p} 2^{-(i+1)}\right) \left(\prod_{j=0}^{s-1}q_j\right)\\
&=& \frac{1}{4}\sum_{s=1}^{\log_2(n)} \frac{s}{n}\left(2\frac{1-p}{p}\right) \left(\prod_{j=0}^{s-1}q_j\right)\\
&\leq & \frac{1}{2}\frac{\log_2(n)}{n}\frac{1-p}{p}\sum_{s=1}^{\log_2(n)} \prod_{j=0}^{s-1}q_j.
\end{eqnarray*}

From definition of $b_{n,p}$, we have
\begin{eqnarray*}
b_{n,p} & \leq & 2 \frac{1}{n}\sum_{s=1}^{\log_2(n)} 2^{s}\prod_{i=0}^{s-1} q_i.
\end{eqnarray*}

Now, note that
\begin{eqnarray*}
\prod_{i=0}^{s-1} q_i & \leq & \prod_{i=0}^{s-1} \left(2\frac{1-p}{p}\right)2^{-i}\\
&=& \left(2\frac{1-p}{p}\right)^s 2^{-\frac{s(s-1)}{2}}\\
&=& \left((\sqrt{2})^3\frac{1-p}{p}\right)^s (\sqrt{2})^{-s^2}.
\end{eqnarray*}

Summing up, we have
\begin{eqnarray*}
\sum_{s=1}^{\log_2(n)}\prod_{i=0}^{s-1} q_i 
&\leq & 2 \left(\frac{1-p}{p}\right) + 2 \left(\frac{1-p}{p}\right)^2 + \sum_{s=3}^{\log_2(n)} \left(\frac{1}{(\sqrt{2})^{s-3}}\frac{1-p}{p}\right)^s\\
&\leq & 2\left(\frac{1-p}{p}\right) + 2 \left(\frac{1-p}{p}\right)^2 + \min\left\{ \frac{\frac{1-p}{p}}{1-\frac{1-p}{p}},\frac{1}{1-1/\sqrt{2}}\right\}.
\end{eqnarray*}

\begin{eqnarray*}
\sum_{s=1}^{\log_2(n)} 2^s\prod_{i=0}^{s-1} q_i 
&\leq & \sum_{s=1}^{\log_2(n)} \left( 2^{\frac{5-s}{2}} \frac{1-p}{p}\right)^s\\
&\leq & \sum_{s=1}^{5} \left( 2^{\frac{5-s}{2}} \frac{1-p}{p}\right)^s  + \sum_{s=5}^{\log_2(n)} \left(\frac{1-p}{p}\right)^s\\
&\leq & \sum_{s=1}^{5} \left( 2^{\frac{5-s}{2}} \frac{1-p}{p}\right)^s + \frac{(1-p)^5}{p^4(2p-1)}\\
&=& O(1-p).%\frac{1-p}{2-p}.
\end{eqnarray*}

It follows that 
$$
a_{n,p}'' = O\left((1-p)^2\frac{\log(n)}{n}\right)
$$
and 
$$
b_{n,p} = O\left((1-p)\frac{1}{n}\right).
$$
Summing $a_{n,p}'$ and $a_{n,p}''$, we have
$$
a_{n,p} \leq \left(\left(\frac{1-p}{p}\right)^2 + \left(\frac{1-p}{p}\right)^4 + \frac{1}{4}\left(1+\frac{1}{3\cdot 2^{5/4}\log(2)}\right)\left(\frac{1-p}{p}\right)^{6}\right)\left(1+O\left(\frac{\log(n)}{n}\right)\right).
$$

%From the above results, we conclude $a_{n,p} + b_{n,p} \leq c_{n,p}$ for some $c_{n,p}$ such that for $\gamma_n = o(1)$, we have
%$$
%c_{n,1-\gamma_n} \sim \gamma_n \hbox{ for large } n.
%$$

\paragraph{Case $p = O(1/n)$} From (\ref{equ:a1}) and (\ref{equ:a2}), note that
$$
a_{n,p}' \leq \frac{1}{4}\sum_{s=1}^{\log_2(n)} p_s \hbox{ and } a_{n,p}'' \leq \frac{1}{2}\sum_{s=1}^{\log_2(n)} p_s 
$$
and, from the definition of $b_{n,p}$, we have
$$
b_{n,p} \leq 2\sum_{s=1}^{\log_2(n)} p_s.
$$

We will use the following basic fact
\begin{eqnarray}
\int_a^b \frac{p 2^s}{1-p+p2^s}ds &=& 
%\frac{1}{\log(2)}\int_{2^a}^{2^b} \frac{p}{1-p+pz}dz\\
%&=& \frac{1}{\log(2)}\int_{1-p+p2^a}^{1-p+p2^b}\frac{dy}{y}\\
%&=& 
\log_2\left(\frac{1-p+p2^b}{1-p+p2^a}\right). \label{equ:basicint}
\end{eqnarray}

Using this basic fact, we have

\begin{eqnarray*}
\sum_{s=1}^{\log_2(n)} p_s &=& \sum_{s=1}^{\log_2(n)} \frac{p2^s}{1-p+p2^s}\\
&\leq & \int_1^{\log_2(n)+1} \frac{p 2^s}{1-p+p2^s}ds\\
&=& \log_2\left(\frac{1-p+2np}{1+p}\right)\\
&\leq & \log_2(1+2np)\\
&\leq & \log_2(1+2c).
\end{eqnarray*}

Hence, it follows that 
$$
a_{n,p}' = O(1), a_{n,p}'' = O(1), \hbox{ and }b_{n,p} = O(1).
$$

\paragraph{Case $\omega(1/n) \leq p < 1/2$} We first upper bound $a_{n,p}'$. Let $c_2$ be a fixed constant in $(0,1)$ and $s^* = (1/2)\log_2(1/c_2) + \log_2((1-p)/p)$. From (\ref{equ:a1}), we have
\begin{eqnarray*}
a_{n,p}' &\leq & 
\frac{1}{4}\sum_{s=1}^{\log_2(n)} p_s \prod_{i=0}^{s-1}q_i^2
\leq \frac{1}{4}\sum_{s=1}^{s^*} p_s + \frac{1}{4}\sum_{s=s^*+1}^{\log_2(n)} \prod_{i=s^*}^{s-1}q_i^2.
\end{eqnarray*}

Note that
\begin{eqnarray*}
\prod_{i=s^*}^{s-1} q_i^2 & \leq &  \prod_{i=s^*}^{s-1}\left(2\frac{1-p}{p}\right)^2 4^{-i}\\
& = & \left(4\left(\frac{1-p}{p}\right)^2\right)^{s-s^*} 2^{-((s-1)s-(s^*-1)s^*)}\\
& = & \left(8\left(\frac{1-p}{p}\right)^2\right)^{s-s^*} 2^{-(s^2-s^*)}\\
&=& \left(\frac{1}{2^{s+s^*-3}}\left(\frac{1-p}{p}\right)^2\right)^{s-s^*}.
\end{eqnarray*}

For every $s\geq s^*+1$, we have
\begin{eqnarray*}
\frac{1}{2^{s+s^*-3}}\left(\frac{1-p}{p}\right)^2 & \leq & \frac{1}{2^{2(s^*-1)}}\left(\frac{1-p}{p}\right)^2 = c_2.
\end{eqnarray*}

Hence, it follows
$$
\sum_{s=s^*+1}^{\log_2(n)} \prod_{i=s^*}^{s-1}q_i^2 \leq \sum_{s=s^*+1}^{\log_2(n)} c_2^s \leq \frac{1}{1-c_2}.
$$

Now, note also
\begin{eqnarray*}
\sum_{s=1}^{s^*} p_s & = & \sum_{s=1}^{s^*} \frac{p2^s}{1-p+p2^s}\\
&\leq & \int_1^{s^*+1} \frac{p2^s}{1-p+p2^s}ds\\
&=& \log_2\left(\frac{1-p+p 2^{s^*+1}}{1+p}\right)\\
&\leq & \log_2(1+ p 2^{s^*+1})\\
&\leq & \log_2\left(1+2/\sqrt{c_2} \right)
\end{eqnarray*}
where we used (\ref{equ:basicint}).

We have thus shown that
$$
a_{n,p}' \leq \frac{1}{4}\left(\log_2\left(1+\frac{1}{\sqrt{c_2}}\right)+\frac{1}{1-c_2}\right).
$$

We next consider $a_{n,p}''$ defined in (\ref{equ:a2}). Fix a constant $c$ in $(0,1)$ and let $s^* = \log_2(1/c_2) + \log_2((1-p)/p) + 1$. We break the sum in the definition of $a_{n,p}''$ in two parts, $A$ and $B$, that can be bounded as follows: 

\begin{eqnarray*}
A &:=& \sum_{s=1}^{s^*} \left(1-\frac{1}{2}p_s\right)\left(1-\frac{1}{2}p_s^2\right)p_s \frac{1}{n}\sum_{i=0}^{s-1} 2^{i+1}\left(\prod_{j=0}^{i}q_j\right)\left(\prod_{j=i+1}^{s-1}q_j^2\right)\\
&\leq & \sum_{s=1}^{s^*} p_s \frac{1}{n}\sum_{i=0}^{s-1} 2^{i+1}\\
&\leq & \sum_{s=1}^{s^*} p_s \frac{1}{n} 2(2^s-1)\\
&\leq & 2\frac{2^{s^*}-1}{n}\sum_{s=1}^{s^*} p_s\\
&\leq & 2\frac{2^{s^*}-1}{n} \log_2\left(\frac{1-p+p2^{s^*+1}}{1+p}\right)\\
&\leq & 2\frac{2^{s^*}-1}{n} \log_2\left(1+p2^{s^*+1}\right)
\end{eqnarray*}

\begin{eqnarray*}
B &:=& \sum_{s=s^*+1}^{\log_2(n)} \left(1-\frac{1}{2}p_s\right)\left(1-\frac{1}{2}p_s^2\right)p_s \frac{1}{n}\sum_{i=0}^{s-1} 2^{i+1}\left(\prod_{j=0}^{i}q_j\right)\left(\prod_{j=i+1}^{s-1}q_j^2\right)\\
&\leq & \sum_{s=s^*+1}^{\log_2(n)} \frac{1}{n}\sum_{i=0}^{s-1} 2^{i+1}\left(\prod_{j=s^*}^{s-1}q_j\right)\\
&\leq & 2 \sum_{s=s^*+1}^{\log_2(n)} \left(\prod_{j=s^*}^{s-1}q_j\right).
\end{eqnarray*}

Note that
\begin{eqnarray*}
\prod_{i=s^*}^{s-1} q_i & \leq &  \prod_{i=s^*}^{s-1}\left(2\frac{1-p}{p}\right) 2^{-i}\\
& = & \left(2\left(\frac{1-p}{p}\right)\right)^{s-s^*} 2^{-\frac{1}{2}((s-1)s-(s^*-1)s^*)}\\
& = & \left((\sqrt{2})^3\left(\frac{1-p}{p}\right)\right)^{s-s^*} (\sqrt{2})^{-(s^2-s^*)}\\
&=& \left(\frac{1}{(\sqrt{2})^{s+s^*-3}}\left(\frac{1-p}{p}\right)\right)^{s-s^*}.
\end{eqnarray*}

For every $s\geq s^*+1$, we have
$$
\frac{1}{(\sqrt{2})^{s+s^*-3}}\left(\frac{1-p}{p}\right) \leq \frac{1}{(\sqrt{2})^{2(s^*-1)}}\left(\frac{1-p}{p}\right) = c_2.
$$

It follows that 
$$
B \leq 2 \frac{1}{1-c_2}
$$
and 
\begin{eqnarray*}
A &\leq & 2 \frac{2\frac{1}{c_2}\frac{1-p}{p}-1}{n}\log_2\left(1+4(1-p)\frac{1}{c_2}\right)\\
&\leq & \frac{4}{c_2}\frac{1}{pn}\log_2\left(1+\frac{4}{c_2}\right).
\end{eqnarray*}

We have thus shown that
$$
a_{n,p}'' \leq \frac{1}{c_2}\frac{1}{pn}\log_2\left(1+\frac{4}{c_2}\right) + \frac{1}{2}\frac{1}{1-c_2} = \frac{1}{2} \frac{1}{1-c_2} + o(1).
$$

We next consider $b_{n,p}$. This is similar to bounding $a_{n,p}''$ by noting that
\begin{eqnarray*}
b_{n,p} &\leq & \frac{1}{n}\sum_{s=1}^{s^*} 2^{s+1}p_s + \frac{1}{n}\sum_{s=s^*+1}^{\log_2(n)} 2^{s+1} \prod_{i=s^*}^{s-1}q_i\\
&\leq & \frac{2(2^{s^*}-1)}{n}\sum_{s=1}^{s^*} p_s + 2 \sum_{s=s^*+1}^{\log_2(n)} \prod_{i=s^*}^{s-1}q_i.
\end{eqnarray*}

By the bounds established for $a_{n,p}''$, we have 
$$
b_{n,p} \leq \frac{1}{c_2}\frac{1}{pn}\log_2\left(1+\frac{4}{c_2}\right) + \frac{1}{2}\frac{1}{1-c_2} = \frac{1}{2} \frac{1}{1-c_2} + o(1).
$$

\newpage
\section*{\textbf{\Large{Chapter~7}}}
\addcontentsline{toc}{section}{Chapter~7.  Conclusions}

\bigskip
\bigskip

\textbf{\Large{Conclusions}}

\bigskip
\bigskip
\bigskip

We conclude with a summary of the work in this thesis and discussions on future research.

\paragraph{Cooperative non-stochastic multi-armed bandits}

We presented new results for the collaborative multi-agent non-stochastic multi-armed bandit with communication delays.
We showed a lower bound on the regret of each individual agent and proposed two algorithms (CFTRL and DFTRL) together with their regret upper bounds. CFTRL provides an optimal regret of each individual agent with respect to the scaling with the number of arms. 
DFTRL has an optimal average regret with respect to the scaling with the edge-delay. Our numerical results validate our theoretical bounds and demonstrate that significant performance gains can be achieved by our two algorithms compared to state-of-the-art algorithms. 

There are several open research questions for future research. The first question is to consider the existence of a decentralized algorithm which can provide  $O((1/\sqrt{|\mathcal{N}(v)|})\sqrt{KT})$ individual regret for each agent $v$. It is unclear whether a center-based communication protocol is necessary to achieve this regret. The second question is to consider whether an algorithm exists with an optimal scaling with the number of arms and the edge-delay parameter. The third question is to understand the effect of edge-delay heterogeneity on individual regrets of agents. 

\paragraph{Doubly adversarial federated bandits}

We studied doubly adversarial federated bandits, a new adversarial (non-stochastic) setting for federated bandits, which complement prior studies on stochastic federated bandits. Firstly, we derived regret lower bounds for any federated bandit algorithm when the agents have access to full-information or bandit feedback.
These regret lower bounds relate the hardness of the problem to the algebraic connectivity of the network through which the agents communicate. Then we proposed the {\fedexp} algorithm which is a federated version of the Exp3 algorithm. We showed that there is only a small polynomial gap between the regret upper bound of {\fedexp} and the lower bound. Numerical experiments performed by using both synthetic and real-word datasets demonstrated that {\fedexp} can outperform the state-of-the-art stochastic federated bandit algorithm by a significant margin in non-stationary environments.

We point out some interesting avenues for future research on doubly adversarial federated bandits. The first is to close the gap between the regret upper bound of {\fedexp} algorithm and the lower bounds shown in this chapter. The second is to extend the doubly adversarial assumption to federated linear bandit problems, where the doubly adversarial assumption could replace the stochastic assumption on the noise in the linear model.

\paragraph{Communication-regret trade-off in federated online learning}
In this study, we examine the online convex optimization in a federated learning setting, where multiple agents collaborate to minimize a convex objective function composed of convex functions distributed over a $T$ time horizon and $N$ agents.
To this end, we propose a novel algorithm called {\pfedexp}, which offers a communication-efficient solution with a cumulative regret upper bound of $o(T)$ and communication complexity of $o(\sqrt{T})$ with high probability.

As future research avenues, we aim to determine the lower bound of communication complexity for algorithms that ensure sub-linear regret upper bounds in federated online convex optimization problems. Additionally, we plan to extend our stochastic communication protocol to more challenging non-stochastic multi-agent multi-armed bandit problems.

\paragraph{Greedy Bayes incremental matching strategies}
We discuss the sequential learning problem of a maximum value matching on a set fixed of nodes with different value functions.
We have identified greedy Bayes incremental matching strategies for Boolean value functions AND and OR, and analysed their regret. Our analysis reveals how regret depends on the population size and its profile with respect to node types. For AND function, the dynamics of posterior beliefs can be represented by a stochastic process of matching sets which are merged or dissolved under greedy matching strategy. 

Future work may consider the problem under different relaxations of assumptions, for different domains of latent node variables, noisy observations of match values, and other definitions of incremental matchings that allow for a larger number of re-matches per round. 

%%%%%%%%%%%%%%%%%%%%%%%%%%%%%%%%%%%%%%

\newpage\onehalfspacing

\bibliographystyle{ACM-Reference-Format}
\bibliography{references}
\addcontentsline{toc}{section}{Bibliography}

%%%%%%%%%%%%%%%%%%%%%%%%%%%%%%%%%%%%%%
\end{document}